\documentclass[]{fairmeta}
\usepackage{longtable}

\title{GIM: Evaluating models via tasks that integrate multiple cognitive domains}

\author[1]{Rohit Patel}
\author[1]{Alexandre Rezende}
\author[1]{Steven McClain}

\affiliation[1]{Meta Superintelligence Labs}

\abstract{
As LLM benchmarks saturate, the evaluation community has pursued two strategies to increase difficulty: escalating knowledge demands (GPQA, HLE) or removing knowledge entirely in favor of abstract reasoning (ARC-AGI). The first conflates memorization with capability; the second divorces reasoning from the practical contexts in which it matters. We take a different approach. The \textbf{Grounded Integration Measure (GIM)} is a benchmark of 820 original problems (615 public, 205 private) where difficulty comes from \emph{integration}; individual problems require coordinating multiple cognitive operations (constraint satisfaction, state tracking, epistemic vigilance, audience calibration) over broadly accessible knowledge, so that reasoning stays grounded in realistic tasks without being gated on specialized expertise. Each problem is an original expert-authored composition, majority with rubric-decomposed scoring (median 6 independently judged criteria). A balanced public--private split provides a built-in contamination diagnostic. We calibrate a judge-aware continuous response 2-parameter logistic (2PL) IRT model across 53 test-configurations (unique model × thinking-level pairs) and five calibrated judges, using 203{,}800 epoch-averaged prompt--judge cells derived from >1M raw judge-scored observations, producing robust ability estimates that correctly order test-configurations even when raw accuracy is distorted by errors, missing data, or judge leniency differences. Using this framework, we present a comprehensive leaderboard spanning 22 models and 47 reporting test-configurations, and conduct what is to our knowledge the most extensive published study of how test-time compute trades off against model capability on a fixed benchmark: 11 models swept across 35 test-configurations. We observe that within-family configuration choices, such as thinking budget and quantization, matter as much as model selection, and increasing thinking tokens has diminishing marginal returns. We release the evaluation framework, calibrated IRT parameters, and all public problems.
}

\date{\today}
\correspondence{\email{gimpaper@rpeml.com}}

\metadata[Code]{\url{https://github.com/facebookresearch/gim}}
\metadata[Dataset]{\url{https://huggingface.co/datasets/facebook/gim}}

\begin{document}
\maketitle

\section{Introduction}
\label{section:intro}

LLM benchmarks saturate quickly \citep{akhtar2026saturation, dehghani2021benchmarklottery}, and the field has responded along two divergent axes (Section~\ref{section:related-work}; Appendix~\ref{appendix:benchmarking-gap}). One axis pushes the knowledge bar into ever more obscure expert domains---GPQA \citep{rein2023gpqa}, HLE \citep{phan2025hle}---where gains conflate coverage with capability. The other strips real-world grounding away in favor of abstract synthetic reasoning---ARC-AGI \citep{chollet2019measure, arcprize2026arcagi3}---where strong performance has not been shown to predict practical utility. We pursue a third route: difficulty arising from the need to integrate many kinds of reasoning at once---coordinating interacting constraints, tracking state, exercising epistemic vigilance, calibrating to an audience, reasoning under temporal ambiguity, and following procedure (Appendix~\ref{appendix:examples}). Consider two GIM examples: \textbf{(1)} A variant of the classic wolf--goat--cabbage river-crossing puzzle with added weight constraints (60\,kg passenger, 70\,kg raft, 5\,kg lift from a dove when held) that invalidate the textbook solution and require coordinating five interacting constraints at once. \textbf{(2)} A historian presents a letter dated October 4, 1955 whose letterhead bears a ZIP code, asking which building hosted the meeting it describes. The right answer: the meeting likely never happened---ZIP codes weren't introduced until 1963. The difficulty is epistemic vigilance, not multi-step reasoning---catching one anachronism and resisting the impulse to answer as asked.

These examples span a broad spectrum of cognitive demands, yielding a benchmark grounded in everyday knowledge whose difficulty stems from cognitive demand rather than specialized expertise.

\subsection{Contributions}

\textbf{Dataset:} 820 original, expert-authored problems (615 public \textbf{(GIM-615)}, 205 private \textbf{(GIM-205)}; collectively \textbf{GIM-820}) spanning 7 categories and 229 multimodal items (Appendix~\ref{appendix:design-principles}). 528 problems (64\%) carry bespoke rubrics (median 6 criteria per problem) that decompose each ideal response into independently judged components. Problems target realistic tasks (analysis, planning, professional communication) and are validated through independent human review (Section~\ref{section:dataset}).

\textbf{IRT Model and Item Bank:} A well-calibrated judge-aware 2PL Item Response Theory model over GIM-820 across 53 \emph{test-configurations}---52 unique (model, thinking-level) pairs and a pilot centaur study---and five calibrated judges, using 203{,}800 epoch-averaged prompt--judge cells derived from >1M raw judge-scored observations. This produces ability estimates comparable across runs with different item coverage and judge labels---essential given the infrastructure instability inherent in running frontier models at high reasoning on complex problems. The item bank is publicly released for future model scoring.

\textbf{Comprehensive leaderboard:} A leaderboard spanning more than $3$ logits of ability with tight ($\pm 0.05$) IRT confidence intervals, covering 47 test-configurations\footnote{Six of the 53 calibrated test-configurations are withheld from reporting for confidentiality.} across 22 models---including a Gemma~4 quantization sweep (bf16/fp8/fp4)---and a human--LLM centaur cohort, providing the breadth needed to make robust cross-family, cross-configuration comparisons (Section~\ref{section:results}).

\textbf{Test-time compute study:} To our knowledge the most extensive published characterization of how test-time compute trades off against model capability on a fixed benchmark: 11 models swept across multiple thinking levels (35 (model, thinking-level) configurations) within an overall set of 53 calibrated test-configurations (Section~\ref{subsection:thinking}).

\section{Related Work}
\label{section:related-work}

Table~\ref{table:comparison} positions GIM relative to existing benchmarks across key design dimensions. GIM does not claim to replace any of these benchmarks---each serves a legitimate purpose---but fills a specific gap left by them. Three methodological strands inform GIM's evaluation framework. First, \emph{rubric-decomposed open-ended grading}. \citet{arora2025healthbench} demonstrated at scale that decomposing responses into expert-authored rubric criteria yields fine-grained, audit-friendly scoring; \citep{wei2024simpleqa, patel2025cws} introduced confidence-weighted scoring where judge certainty scales each judgment's contribution. GIM extends rubric-decomposed grading from medicine to general reasoning: 528 of 820 problems carry bespoke rubrics under confidence-weighted aggregation. Decomposing judgments into atomic, verifiable checks is also a structural defense against heuristic exploitation \citep{mccoy2019right, geirhos2020shortcut}.

\begin{table}[!htbp]
\centering
\caption{Positioning of GIM relative to existing LLM benchmarks. Each column represents a benchmark or benchmark class; rows capture key design dimensions.}
\label{table:comparison}
\resizebox{\textwidth}{!}{%
\begin{tabular}{l c c c}
\toprule
\textbf{Dimension} & \textbf{HLE / GPQA} & \textbf{GIM (Ours)} & \textbf{ARC-AGI / AGI-3} \\
\midrule
Difficulty source & Knowledge obscurity & \textbf{Integration density} & Novel rule inference \\
Knowledge required & Deep, specialized & Broadly accessible & None (self-contained) \\
Format & MC + short answer & Single-turn Q\&A & Grid / agentic environment \\
Real-world relevance & Low (niche expertise) & \textbf{High (practical tasks)} & Low (synthetic) \\
Scoring granularity & Binary & \textbf{Rubric-based (multi-component)} & Binary \\
Multimodal & No & \textbf{Text + images + PDFs} & Visual (grids) \\
Ability estimation & Mean accuracy & \textbf{2PL IRT (missing-data robust)} & Mean accuracy \\
\bottomrule
\end{tabular}%
}
\end{table}

Second, \emph{Item Response Theory for ability estimation}. \citet{lalor2016building} introduced IRT to NLP evaluation; \citet{polo2024tinybenchmarks} showed that 100 IRT-selected items from MMLU's 14{,}000 reproduce full-leaderboard rankings, establishing IRT as practical for sample-efficient LLM benchmarking. GIM calibrates a 2-parameter logistic model over GIM-820 and reports ability $\theta$ with standard errors, motivated by the fact that no two frontier model runs cover identical item sets (timeouts, broken multimodal pipelines, and partial sweeps are routine).

Third, \emph{test-time compute as a measurement dimension}. As thinking-budget controls have become primary capability levers \citep{openai2024o1systemcard, snell2024scaling}, single-point evaluation per model is increasingly misleading. Yet most published results report each model at a single inference configuration.

\section{Dataset Construction \& Quality Control}
\label{section:dataset}

Of the 820 problems, 591 are text-only, 150 include image attachments, and 79 include PDF attachments. Each prompt carries free-text descriptive labels (e.g., ``math'', ``reasoning'', ``verifiable'') assigned by annotators at authoring time and subsequently rationalized by hand (consolidating near-duplicates and correcting spelling), yielding 117 distinct labels across the bank. In addition to these author-assigned labels, we developed an independent taxonomy in which an LLM reads each prompt and classifies it into 7 primary categories and 18 sub-categories, providing a structured view of the bank that is decoupled from the original labeling process. Additional details, including the full taxonomy, the hierarchical visualization, and the per-prompt category-overlap distribution, can be found in Appendices~\ref{appendix:taxonomy}, \ref{appendix:taxonomy-sunburst}, and~\ref{appendix:co-occurrence}.

\phantomsection\label{subsection:leakage-prevention}\textbf{Leakage prevention:}
GIM applies multiple safeguards against data contamination: prior to the leaderboard reported in this paper, 100\% of the 820 prompts were held privately, so no model evaluated here could have been trained on any GIM problem. Looking forward, the public-private split will be the main safeguard (Section~\ref{subsection:contamination-diagnostic}). Evaluation and judging is done using APIs with agreements preventing use of data for training. Additional details in Appendix~\ref{appendix:leakage-prevention}.

\textbf{Domain Expert Sourcing:} Subject-matter experts---mathematicians, software engineers, and specialists in law, medicine, and finance---collaboratively brainstorm each problem, following structured authoring guidelines that enforce originality, determinism, appropriate difficulty, and timelessness. For rubric-graded problems, the same experts decompose each problem into criteria that are atomic, self-contained (judgable without the reference answer, e.g., ``Calculates the length of side AB as 12 meters'' rather than ``Calculates the length of side AB''), mutually exclusive, and collectively exhaustive. See Appendix~\ref{appendix:prompt-guidelines} and Appendix~\ref{appendix:rubric-guidelines} for full guidelines.

\textbf{Rubric Review:} Rubrics were stress-tested under multiple judge models; criteria that consistently elicited low judge confidence across runs were flagged and rewritten.

\textbf{Two-Round Correctness Review:} All problems pass through two review rounds with the same objectives: clarity, unambiguous ground-truth answers, sound reasoning paths, and no alternative valid interpretations. Reviewers also check formatting consistency, difficulty calibration, and taxonomy alignment.

\textbf{Authoring Effort:} Each of the 820 prompts in GIM required, on average, approximately 11 person-hours from initial drafting through reference-answer derivation, rubric decomposition, peer review, and quality-assurance---roughly 9{,}000 person-hours, or about four person-years of expert author time, in total across the benchmark.

\section{Evaluation Methodology}
\label{section:methodology}

GIM's evaluation infrastructure targets three requirements: (i)~full reproducibility from a single command, (ii)~standardized prompting that isolates model reasoning from prompt engineering, and (iii)~scoring that captures partial credit rather than collapsing each response to a binary verdict. Implementation details are in Appendix~\ref{appendix:eval-details}.

\phantomsection\label{subsection:cot-temperature}\textbf{Pipeline:} The evaluator is implemented as an Inspect AI~\citep{inspect2024} task plugin with \texttt{uv}-pinned dependencies~\citep{astral2024uv}; the dataset is distributed in HuggingFace \texttt{datasets} format with \texttt{public} and \texttt{private} splits, modality-filtered so models are only evaluated on problems they can process. Each sample is presented for $k=5$ epochs as a single-turn prompt under each provider's default tool configuration; transient API failures are handled by in-run retries plus a manually launched second-pass run. We do not inject chain-of-thought, and models with built-in reasoning are evaluated with thinking enabled~\citep{snell2024scaling} at multiple budgets where supported. Generation temperature is left unset (reasoning-enabled models universally lock it~\citep{openai2024o1systemcard}), with multi-epoch averaging stabilizing scores.

\phantomsection\label{subsection:hybrid-scoring}\textbf{Hybrid scoring:} Two complementary strategies, deterministically routed by the presence of rubrics~\citep{kim2024prometheus, ye2024flask, liu2023geval}. \emph{Rubric-graded} prompts decompose into $n$ independently judged criteria~\citep{min2023factscore}, each yielding a score $s_i \in [0,1]$ that rewards partial credit~\citep{kadavath2022language}. \emph{Exact-answer} prompts compare the model's output to a golden target, accounting for representational equivalence (e.g., $0.5 = \tfrac{1}{2}$), format variations, and null equivalences, yielding a single score $s_1 \in [0,1]$. In both strategies the judge additionally emits a confidence $c_i \in [0,1]$ alongside each score, and the per-sample score is the confidence-weighted average $\frac{1}{n}\sum_{i=1}^{n} s_i \cdot c_i$. GIM supports a calibrated panel of five LLM judges under structured-output constraints~\citep{zheng2023judging} that mitigate verbosity and positional biases~\citep{wang2023llmfairevaluators}: Gemini~3~Flash, Gemini~3.5~Flash, GPT-5.4~Mini, Claude~4.5~Haiku, and Gemma~4~31B.

\phantomsection\label{subsection:judge-consistency}\textbf{Judge consistency:} We re-scored the full bank with five judges---Gemini~3~Flash, Gemini~3.5~Flash, GPT-5.4~Mini, Claude~4.5~Haiku, and Gemma~4~31B---and re-estimated the calibration across all 53 configurations. Pairwise agreement is high at the score level ($r=0.919$ on average) and for IRT difficulty ($r=0.911$), discrimination ($r=0.910$), ability ($r=0.996$), and rank order (Spearman $\rho=0.996$). Judges differ in raw leniency by up to 8.2 percentage points, but the centered judge fixed effects $\gamma_k$ absorb this: scoring a model with any single calibrated judge against the released item bank recovers the joint ability to within $\approx0.13$ logits (MAE) and at most three rank places, so a previously unranked model can be placed on the leaderboard from a single judge run. Appendix~\ref{appendix:judge-multi} reports the full five-judge analysis.

\phantomsection\label{subsection:centaur-evaluation}\textbf{Human Baseline:} We adopt a \emph{centaur} baseline---humans with unrestricted tool access, including LLMs---as the realistic point of comparison for frontier models. This framing suits GIM, whose problems rarely hinge on obscure facts but on knowledge anyone can grasp once retrieved, isolating reasoning from recall. Concretely, 246 participants (via a third-party staffing partner) each receive 20 random prompts within a five-hour budget; we report "Top" and "Average" groups. See Appendix~\ref{appendix:centaur-details}.

\textbf{Compute footprint:} Each reported test-configuration consumes $\sim$56M inference tokens at $k=5$ epochs plus judge-scoring tokens. The legacy single-judge cost estimate is $\sim$20M judge tokens per configuration, or $\sim$76M total tokens; five-judge panel scoring reuses model responses but adds judge calls. The 47 reporting configurations therefore cost roughly $10^{21}$ inference FLOPs at frontier active-parameter counts, with judging cost scaling by the selected judge panel. Additional details can be found in Appendix~\ref{appendix:compute-footprint}.

\phantomsection\label{subsection:irt-model}\textbf{Model:} We summarize each model's performance on GIM with a single ability parameter $\theta$ estimated under a three-facet (configuration $\times$ item $\times$ judge) continuous-response 2PL \citep{samejima1973continuous,bock1970fitting,linacre1994many}, hereafter the judge-aware 2PL: two parameters per prompt on the item facet, and a single centered severity term per judge. Let $p_{ijk} \in [0,1]$ be the rubric-weighted score for model-thinking configuration $i$ on prompt $j$ from calibrated judge $k$, after averaging repeated epochs within each observed \mbox{(prompt, judge)} cell. We apply the edge-corrected logit transform~\citep{smithson2006better}
\begin{equation}
p'_{ijk} = 0.998p_{ijk}+0.001, \qquad y_{ijk} = \log\!\left(\frac{p'_{ijk}}{1-p'_{ijk}}\right),
\end{equation}
and assume
\begin{equation}
y_{ijk} \;=\; a_j (\theta_i - b_j) + \gamma_k + \varepsilon_{ijk}, \qquad \varepsilon_{ijk} \sim \mathcal{N}(0, \sigma^2),
\end{equation}
where $a_j > 0$ is the prompt's \emph{discrimination}, $b_j$ is its \emph{difficulty} (the ability at which a centered judge expects $p = 0.5$), and $\gamma_k$ is a centered judge fixed effect with $\sum_k \gamma_k = 0$. Positive $\gamma_k$ means judge $k$ is more lenient in logit-score units.\footnote{The judge term enters additively rather than on the latent scale as in the standard many-facet form $a_j(\theta_i - b_j - \gamma_k)$~\citep{linacre1994many}, since a lenient judge reads responses more generously against a fixed rubric rather than re-estimating item difficulty.} The exact transform and squeeze-magnitude sensitivity check are in Appendix~\ref{appendix:calibration-trust-detail}; we defer further motivation to \S\ref{subsection:irt}.

\textbf{Calibration:} We jointly estimate $\{a_j, b_j, \theta_i, \gamma_k\}$ by minimizing the masked logit-space MSE across all 53 configurations, 820 prompts, and five calibrated judges, with ridge penalties fixing the 2PL scale/location and the judge effects centered for identifiability. Calibrated parameters for the 615 public prompts are released; calibration trust is quantified in Section~\ref{subsection:calibration-trust}.

\section{Results}
\label{section:results}

We score models on GIM using an Item Response Theory (IRT) approach; the underlying response model and joint calibration procedure are specified in Section~\ref{subsection:irt-model}. The reported GIM score for each model is its IRT-derived ability estimate $\theta$. IRT is essential at this scale: at higher thinking budgets on harder prompts a non-trivial fraction of attempts return truncated or empty responses for purely infrastructural reasons, and a raw mean would read these as low ability. The IRT scorer treats such cells as missing rather than zero (e.g., GPT 5.4 X-High's $2.3\%$ failure rate nudges its raw mean below High, while $\theta$ correctly orders X-High above High; see Appendix~\ref{appendix:irt-affordances}). All results in this section report on GIM-820. Benchmark-side diagnostics, including the public/private contamination check, are deferred to Section~\ref{section:benchmark-characteristics}.

\subsection{IRT Scoring}
\label{subsection:irt}

Given the calibrated item bank $\{a_j, b_j, \gamma_k\}$ from the joint fit (Section~\ref{subsection:irt-model}), the maximum-likelihood ability for any model with observed scores on a subset of prompt--judge cells $S$ admits a closed-form weighted least squares solution:
\begin{equation}
\theta \;=\; \frac{\sum_{(j,k) \in S} a_j\bigl(y_{jk} - \gamma_k + a_j b_j\bigr)}{\sum_{(j,k) \in S} a_j^2}, \qquad \mathrm{SE}(\theta) \;=\; \frac{\hat\sigma}{\sqrt{\sum_{(j,k) \in S} a_j^2}}.
\end{equation}
No optimization is required, missing prompt--judge cells contribute nothing, and high-discrimination items dominate. The leaderboard is scored on the joint five-judge-calibrated scale: each observed prompt--judge cell is adjusted by that judge's calibrated fixed effect before contributing to $\theta$, and multi-judge analyses retain judge-labeled observations after aggregation within each \mbox{(prompt, judge)} cell. The \textbf{GIM score} reported throughout this paper is this $\theta$ (with 95\% CI given by $\theta \pm 1.96\,\mathrm{SE}$); the same scorer applies to any subset of the bank, which we exploit for category-conditional $\theta^{(c)}$ (Section~\ref{subsection:categories}) and for the public/private split (Section~\ref{subsection:contamination-diagnostic}). The IRT layer does not invent a different ranking (see raw-mean correlation in Section~\ref{subsection:saturation}) but absorbs missing data coherently, adjusts for calibrated judge leniency/severity, weights items by Fisher information, and rectifies inference-failure noise (Appendix~\ref{appendix:irt-affordances}).

\subsection{Main Leaderboard}
\label{subsection:leaderboard}

\begin{figure}[htbp]
\centering
\includegraphics[width=\textwidth,height=0.9\textheight,keepaspectratio]{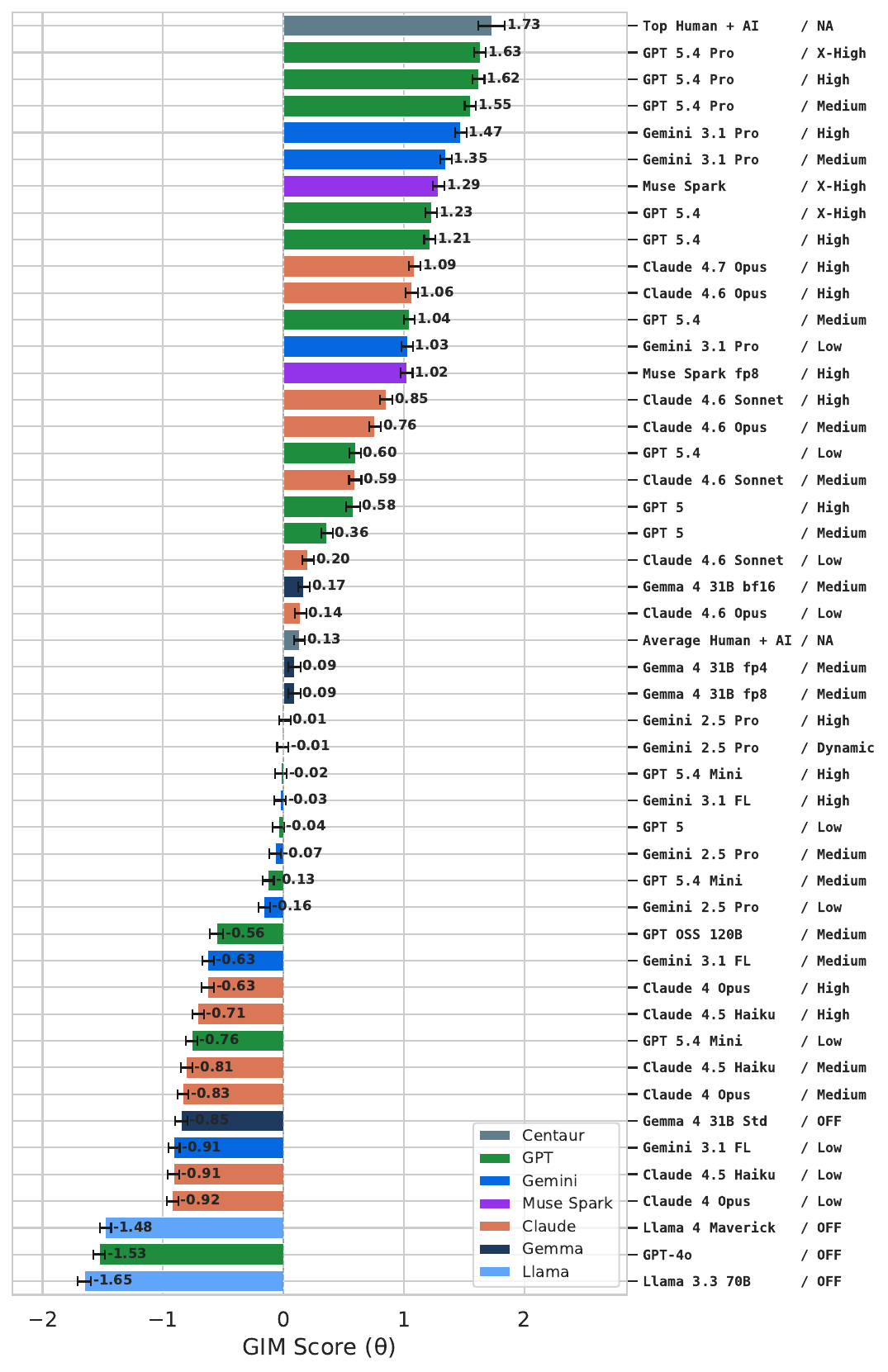}
\caption{GIM-820 leaderboard: IRT ability $\theta$ with 95\% CIs by model and thinking level. All models have 5 epochs except Claude 4.7 Opus (2) and the centaur entries (pilot study, partial sample: \emph{Top Human + AI} aggregates 15 operators / 231 prompt--response pairs; \emph{Average Human + AI} aggregates 195 retained participants). Both centaur entries are scored across all five judges, like the autonomous models. Centaur CIs are correspondingly wider and the entries should be read as pilot estimates rather than head-to-head comparisons.}
\label{fig:overall-model}
\end{figure}

Figure~\ref{fig:overall-model} shows the GIM-820 leaderboard: IRT ability $\theta$ with 95\% CIs for selected test-configurations, colored by family. The board spans roughly $3$ logits, and the CIs are tight for fully covered model entries ($\approx\!\pm 0.05$ for most 95\% intervals); configurations within $\sim 0.1$ logits overlap and should not be over-interpreted.

The headline is that within-family configuration choices matter at least as much as the choice of family. Several families span roughly $0.6$--$0.9$ logits between their lowest and highest thinking levels (Claude 4.6 Opus, Gemini 3.1 FL, GPT 5, GPT 5.4, GPT 5.4 Mini)---comparable to or larger than the gap between adjacent frontier families---and the Gemma 4 31B bf16/fp8/fp4 quantization sweep makes the same point on a different axis. ``Pick the right model'' is incomplete; the configuration it is run in (thinking budget, quantization, runtime parameters) is comparably consequential.

The lower tail is dominated by configurations run without an internal reasoning step---a clean break from thinking-enabled models. This is not merely an age effect: Gemma 4 31B with thinking disabled drops to $\theta = -0.85$, roughly one logit below the same base model with thinking enabled at Medium ($+0.17$, bf16).

The human--LLM centaur results bracket the model leaderboard at both ends, but should be read as a pilot rather than as a head-to-head comparison with the autonomous models. The \emph{Top Human + AI} aggregate---the top 15 operators by individual judge-aware IRT ability, covering 189 prompts after missing-response filtering---comes in at $\theta=1.73$, above the autonomous frontier. We read this as a suggestive existence proof that with the right operators a human-in-the-loop can extract additional signal from frontier models, not as evidence that humans systematically beat them. The \emph{Average Human + AI} aggregate (195 retained participants covering all 820 prompts) lands at $\theta=0.13$, near the middle of the broader reporting set. Several caveats temper any strong reading: the per-prompt time budget ($\approx 15$ minutes) was too short for typical participants to consistently consult, validate, and edit model outputs; the pilot does not separate ``human picks among LLM candidates'' from ``human supplies reasoning the LLM lacked''; and the small $N$ behind the top slice yields wider CIs than the model entries. An AI-choice-only ablation that isolates the picking-vs-supplying confound is the natural follow-up; see Section~\ref{subsection:centaur-evaluation} and Section~\ref{section:limitations}.

\subsection{Performance and Difficulty by Labels and Category}
\label{subsection:categories}

We score abilities on slices of the bank---by primary category and by free-text label---using the same closed-form WLS scorer, and use the calibrated $b_j$ to rank items by difficulty.

Per-slice ability is remarkably stable: $\theta$ is typically within a few tenths of a logit of a configuration's Overall, and the Overall ranking is preserved on most slices. Specialization peaks do appear---Muse Spark on \emph{spatial} and \emph{intuitive}, GPT 5.4 Pro on \emph{puzzles}, Claude 4.6 Opus on \emph{temporal reasoning}, Gemini 3.1 Pro on \emph{lateral thinking}---and knowledge-heavy labels (\emph{web search}, \emph{biology}, \emph{chemistry}) compress inter-model spread sharply at the frontier. We read the flatness as a property of the prompts: a typical GIM item taxes several reasoning dimensions at once, so per-slice $\theta$ is a differential lens on a shared ability, not a score on a separable sub-skill. Given the centaur pilot's small per-operator $N$ and the unaudited picking-vs-supplying confound (Section~\ref{section:limitations}), centaur per-category gaps should be read as suggestive rather than precise.

Looking at difficulty, reasoning-flavored slices (ordering, spatial, multi hop; QR, WK, SI, LR primary categories) sit at the hard end of the $b_j$ distribution, and instruction-following-flavored slices (formatting, language, constraints; CT) sit at the easy end. Crucially, that the two independently constructed views (labels and categories) of the bank converge on the same reasoning-hard / instruction-easy split is a strong cross-validation of the taxonomy itself. The CT placement is partly a labeling artifact---hard multi-faceted prompts that mix constraints with quantitative or logical demands get coded under their dominant non-CT axis---but the underlying split holds in both views. Heatmaps, the top-configurations radar, and the full per-label difficulty rankings are in Appendix~\ref{appendix:category-breakdown} (Figures~\ref{fig:ability-by-category}, \ref{fig:ability-by-label}, \ref{fig:radar-top-models}) and Appendix~\ref{appendix:difficulty-breakdown}.

\subsection{Effect of Test-Time Compute}
\label{subsection:thinking}

Modern frontier models expose a configurable ``thinking budget'' (Low / Medium / High / X-High in our taxonomy) which trades inference cost for additional internal reasoning before producing a final answer. Because we treat each (model, thinking-level) combination as an independent IRT examinee, we can read the effect of test-time compute directly off the calibrated $\theta$ scale, with proper standard errors and without confounding by the model's overall ability. We focus on the marginal gain $\Delta\theta = \theta(\text{level}) - \theta(\text{previous level})$ between consecutive thinking budgets.

\begin{figure}[htbp]
\centering
\begin{minipage}[c]{0.55\linewidth}
\centering
\includegraphics[width=\linewidth,keepaspectratio]{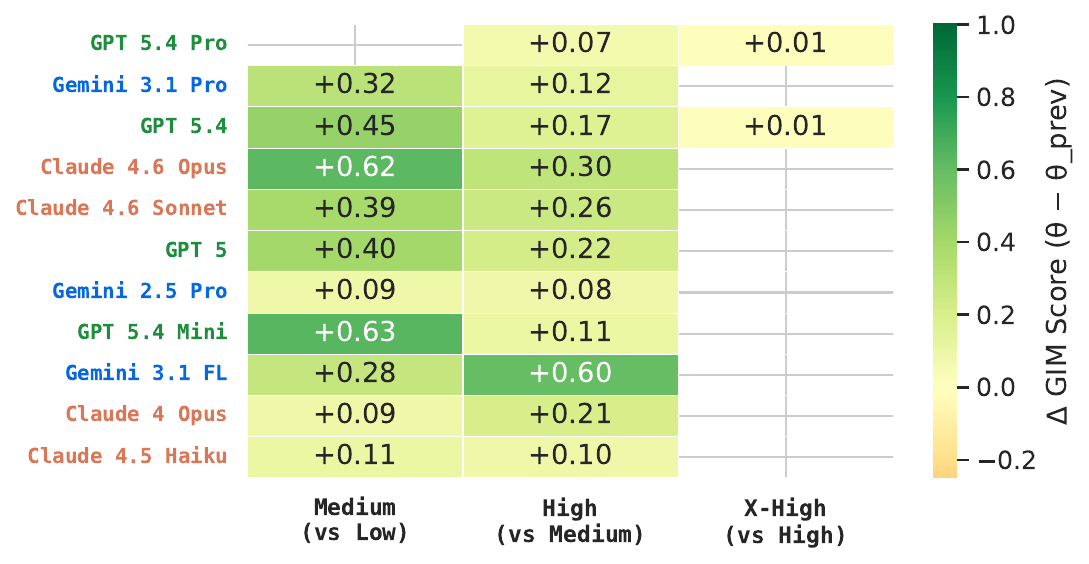}
\end{minipage}\hfill
\begin{minipage}[c]{0.43\linewidth}
\centering
\includegraphics[width=\linewidth,keepaspectratio]{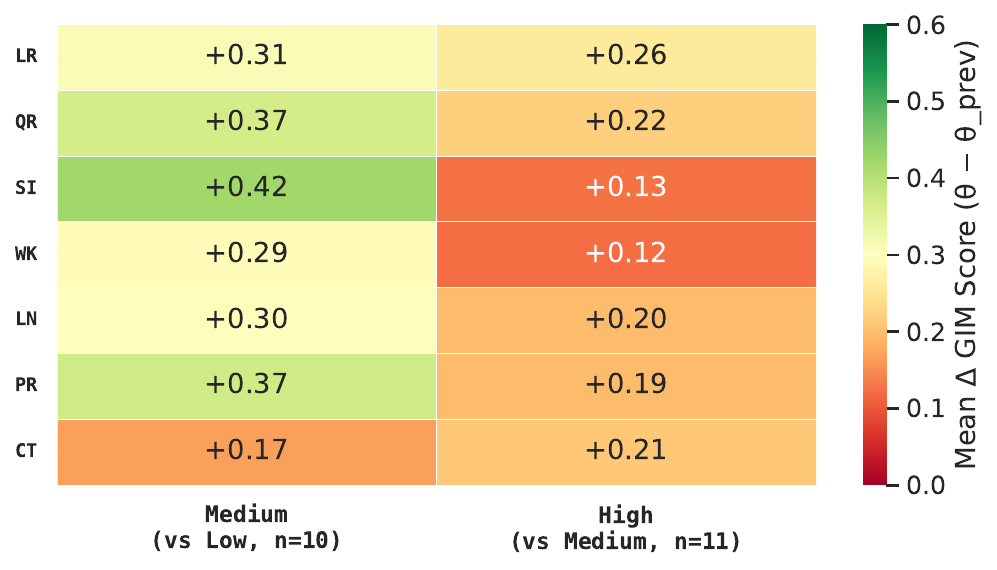}
\end{minipage}
\caption{Marginal gain in GIM score $\Delta\theta$ when stepping up one thinking level. \emph{Left:} per-model heatmap, $\theta(\text{column}) - \theta(\text{previous level})$. \emph{Right:} mean per-model gain by primary cognitive category at each step. Per-label and per-modality breakdowns are in Appendix~\ref{appendix:thinking-detail}.}
\label{fig:thinking-gain}
\end{figure}

Figure~\ref{fig:thinking-gain} (left) shows $\Delta\theta$ for every model with at least two consecutive thinking levels. Two regimes are visible. \emph{Strong gainers}---Claude 4.6 Opus ($+0.92$ Low~$\to$~High), Gemini 3.1 FL ($+0.88$), GPT 5.4 Mini ($+0.74$), Claude 4.6 Sonnet ($+0.65$), GPT 5 ($+0.62$), GPT 5.4 ($+0.62$)---move roughly $0.6$--$0.9$ logits from Low to High, comparable to the spread between two distinct frontier models. \emph{Diminishing returns} appear at the top of the leaderboard: GPT 5.4 Pro's X-High gain is only $+0.012$ logits over High, well within its standard error.

Aggregating the same marginal $\Delta\theta$ across models and slicing by primary cognitive category (Figure~\ref{fig:thinking-gain}, right) shows where extra thinking pays off: at the Medium-vs-Low step, Spatial \& Intuitive ($+0.42$), Procedural and Quantitative Reasoning (both $+0.37$), and Logical Reasoning ($+0.31$) lead, with Constraints ($+0.17$) at the bottom. At the High-vs-Medium step, Logical Reasoning ($+0.26$), Quantitative Reasoning ($+0.22$), and Constraints ($+0.21$) continue to gain, while World Knowledge and Spatial \& Intuitive are lower ($+0.12$--$+0.13$). The same pattern reproduces at the finer free-text-label resolution and along input modality (Appendix~\ref{appendix:thinking-detail}, Figures~\ref{fig:thinking-gain-by-labels}, \ref{fig:thinking-gain-by-modality}): reasoning-flavored slices and text-only inputs gain the most; knowledge- and retrieval-flavored slices the least.

\section{Benchmark Characteristics}
\label{section:benchmark-characteristics}

Section~\ref{section:results} ranked the models. We now turn the lens back on the bank itself with diagnostics that progressively license the leaderboard. Additional details can be found in Appendix~\ref{appendix:bank-diagnostics}.

\begin{figure}[htbp]
\centering
\begin{minipage}[c]{0.8\textwidth}
\centering
\includegraphics[width=\linewidth,keepaspectratio]{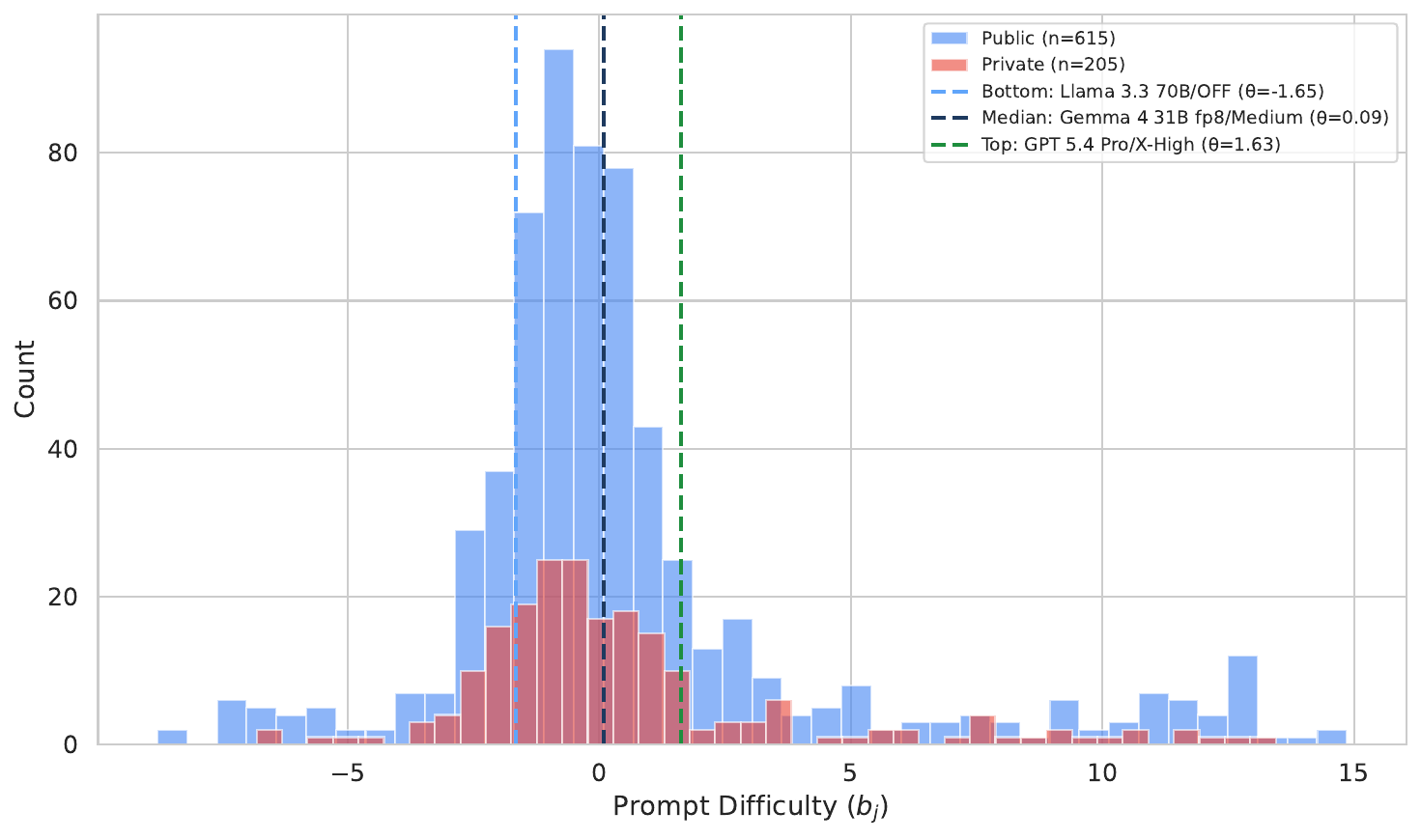}
\end{minipage}\hfill
\begin{minipage}[c]{0.18\textwidth}
\caption{Distribution of fitted item difficulty $b_j$ on the public (blue) and private (red) splits.}
\label{fig:difficulty-distribution}
\end{minipage}
\end{figure}

\phantomsection\label{subsection:item-bank}\textbf{Item difficulty:}
Figure~\ref{fig:difficulty-distribution} shows the distribution of fitted item difficulties $b_j$ for the public (615) and private (205) splits, with the bottom, median, and top configurations overlaid as reference markers. The bulk of the bank lies in the central $\approx\!14$ logits (5th--95th percentile $\approx -3.3$ to $+10.3$), with heavier tails extending from about $-8.8$ to $+14.9$; the public and private histograms are visually indistinguishable in shape and range---a first hint of public--private equivalence. Crucially, even the strongest configuration's $\theta$ ($\approx 1.63$) sits well to the left of the upper tail of the difficulty distribution: many items have $b_j$ above frontier ability, so the benchmark is far from saturated. Discrimination structure ($a_j$) is detailed in Appendix~\ref{appendix:item-bank-detail}.

\phantomsection\label{subsection:saturation}\textbf{Model-side saturation coverage:}
Two quick checks. (1) The fraction of prompts a configuration scores exactly $1.0$ on (epoch-mean) tops out at $\approx 39\%$ for the strongest model, with most configurations well below---no model is anywhere near solving the bank (Appendix~\ref{appendix:saturation-detail}, Figure~\ref{fig:saturation-rates}). (2) Across all 47 reporting configurations, raw mean and $\theta$ correlate at Pearson $r \approx 0.99$ (Figure~\ref{fig:raw-vs-theta} in the same appendix), confirming that the IRT layer is not inventing a different ranking---it adds the affordances of Section~\ref{subsection:irt} on top of an order the raw mean already agrees with.

\phantomsection\label{subsection:tokens-vs-difficulty}\textbf{Token cost vs.\ item difficulty:}
Per-prompt mean total tokens (input + output, averaged across all configurations) grow monotonically and roughly exponentially in $b_j$: harder prompts elicit longer reasoning traces from essentially every model, regardless of family or thinking tier (Appendix~\ref{appendix:tokens-detail}, Figure~\ref{fig:difficulty-vs-tokens}). The difficulty axis thus tracks a directly observable behavioral cost, complementing the model-side cost-of-thinking analysis in Section~\ref{subsection:thinking}.

\phantomsection\label{subsection:calibration-trust}\textbf{Calibration trust:}
A leave-one-model-out (LOMO) refit removes each of the 47 reporting configurations in turn, refits the entire bank on the remaining 46, and re-scores the held-out configuration against those frozen item parameters. The held-out $\theta_i^{(-i)}$ recovers the joint $\theta_i$ to within $0.063$ logits across all configurations (median $0.024$, mean $0.024$), close to the typical 95\% CI half-width of $\approx 0.05$, and no refit produced a deviation in the ``model-dependent'' range we would consider a calibration failure. We therefore think the released item bank is well-calibrated and valuable in calculating GIM Score for future model runs without retraining (Appendix~\ref{appendix:calibration-trust-detail}, Figure~\ref{fig:lomo-calibration}).

Re-scoring each model's 5 epochs independently against the frozen bank quantifies single-epoch noise: deviations from the across-epoch mean mostly fall within $\pm 1.96 \cdot \widetilde{\mathrm{SE}}$ (median $|\Delta\theta| \approx 0.05$, 95th pct $\approx 0.16$, max $\approx 0.30$), with no drift across ability. The broad leaderboard ordering is stable at a single epoch and the IRT uncertainty absorbs most epoch-to-epoch variability, but since the largest swings rival adjacent leaderboard gaps, we recommend multiple epochs when fine-grained ranking matters (Appendix~\ref{appendix:calibration-trust-detail}, Figure~\ref{fig:epoch-variance}).

\phantomsection\label{subsection:contamination-diagnostic}\textbf{Contamination diagnostic using public--private split:}
Using the frozen item parameters from the joint IRT calibration, we re-scored every model separately on the public 615 and private 205 subsets via the same closed-form WLS---no re-fitting. Across all 47 configurations, $\theta_{\text{public}}$ and $\theta_{\text{private}}$ track the identity line (Pearson $r \approx 0.97$); the median $|\theta_{\text{public}} - \theta_{\text{private}}|$ is $\approx 0.13$ logits and the 95th percentile is $\approx 0.45$. No model shows the systematic public-side advantage that contamination would produce. Going forward, we will flag any model whose $|\theta_{\text{public}} - \theta_{\text{private}}|$ exceeds $2 \times \mathrm{SE}_{\Delta}$ for review (Appendix~\ref{appendix:contamination-detail}, Figures~\ref{fig:public-vs-private} and~\ref{fig:public-vs-private-absdiff}; response procedure and item-retirement policy in Appendix~\ref{appendix:maintenance}).

\section{Limitations}
\label{section:limitations}

\textbf{LLM-judge dependency and judging cost:} GIM uses a calibrated five-judge panel with centered fixed effects, and Appendix~\ref{appendix:judge-multi} shows that score correlations, item parameters, and rankings are stable across the panel. The residual limitation is that all scorers are still LLM judges rather than human ground truth, and issuing separate judge calls per rubric criterion exceeds 4{,}000 calls per epoch per test-configuration. Our top priority is therefore to make panel scoring cheaper by supporting calibrated open-weight judging and by running a \emph{batched-rubric} study---scoring multiple criteria (and prompts) per call to find the largest batch that preserves per-criterion scores.

\textbf{Single-turn, model-only evaluation:} Every GIM problem is presented as a single user turn answered in a single assistant turn (Section~\ref{subsection:cot-temperature}). While nothing in GIM's design is opinionated about the internals of the system under test, the results in this paper evaluate only models. The rise of agentic systems introduces a new class of test subject in which the orchestration layer, tool choice, and multi-step execution strategy can matter as much as the underlying model weights.

\textbf{Rubric and ranking coverage:} Of the 820 problems, 292 (36\%) use exact-answer grading. In retrospect, we would author rubrics for every problem---rubric-graded items capture partial credit that a binary verdict discards. A related gap is ranking problems ($\sim$10\% of the bank), currently scored with the exact-answer judge; a rank-correlation metric such as Kendall's $\tau$ would extract finer-grained signal.

\textbf{Modality coverage:} The 229 multimodal problems (28\%) cover images and documents but not audio or video, which are now widely supported by frontier systems.

\textbf{Centaur design constraints:} The centaur study (Section~\ref{subsection:centaur-evaluation}) is a pilot. It conflates ``picking among LLM candidates'' with ``supplying reasoning the LLM lacked,'' the 15-min per-prompt budget is tight for consulting and editing model outputs, and the \emph{Top Human + AI} entry rests on a small slice (15 operators, 231 pairs) with correspondingly wide CIs. We thus treat centaur numbers as suggestive context, not head-to-head wins. A pick-only ablation (no free-form reasoning) would isolate the confound; a longer budget and larger top-operator pool would address the rest.

\textbf{English-only:} GIM-820 is monolingual English. Reasoning processes that interact with linguistic structure may not transfer to languages with substantially different syntactic or pragmatic affordances.

\section{Conclusion}
\label{section:conclusion}

We release the GIM-615 public split, the calibrated item bank, and the evaluation framework, so that future models can be placed on the same ability scale at marginal cost. We see the most immediate extensions as broadening modality (audio, video), evaluating agentic systems end-to-end rather than single-turn models, and reducing per-criterion judging cost via batched-rubric studies (Section~\ref{section:limitations}) so the five-judge panel analysis in Appendix~\ref{appendix:judge-multi} can scale to broader judge sets and cheaper open-weight scoring.

\textbf{Broader impact.} GIM offers researchers and policymakers a stable reference point for tracking frontier model capability and for grounding discussions of capability trajectories and governance.

\textbf{AI usage disclosure:} A stage-by-stage breakdown can be found in Appendix~\ref{appendix:ai-usage}.

\section*{Acknowledgments}
Special thanks to \textbf{Di Lin} for invaluable support throughout the project. We are also grateful to the \textbf{CRAG-MM}~\citep{cragmm2025} \textbf{and WearVQA}~\citep{wearvqa2025} \textbf{teams}, who, in collaboration with the GIM authors, each contributed 50 prompts produced through their respective data-construction pipelines as a dedicated contribution to this benchmark. We also thank \textbf{Manohar Paluri} for sponsorship and support, and \textbf{Labelbox} for providing annotation support.

\clearpage
\newpage
\nocite{*}
\bibliographystyle{assets/plainnat}
\bibliography{paper}

\clearpage
\newpage
\beginappendix

\section{The Benchmarking Gap}
\label{appendix:benchmarking-gap}

Large language model benchmarks have followed a pattern of rapid saturation: evaluations that once discriminated sharply between model generations---GLUE \citep{wang2018glue}, SuperGLUE \citep{wang2019superglue}, HellaSwag \citep{zellers2019hellaswag}---now yield near-ceiling scores. MMLU \citep{hendrycks2021mmlu}, which in 2021 exposed substantial gaps between human experts and frontier models, saw GPT-4 exceed 86\% accuracy within two years \citep{openai2023gpt4} and subsequent systems approach or surpass 90\% \citep{team2023gemini, anthropic2024claude}. A systematic study of 60 benchmarks finds that nearly half exhibit saturation that intensifies with age \citep{akhtar2026saturation, dehghani2021benchmarklottery}.

The community has responded along two divergent axes. One family escalates difficulty by demanding deeper or more obscure knowledge---GPQA \citep{rein2023gpqa}, HLE \citep{phan2025hle}---so that gains conflate knowledge coverage with cognitive capability. The other removes domain knowledge entirely, testing reasoning in synthetic, self-contained environments---ARC-AGI \citep{chollet2019measure}, ARC-AGI-3 \citep{arcprize2026arcagi3}---but strong performance on grid puzzles has not been shown to predict a model's ability to analyze a contract, plan a logistics operation, or evaluate the reliability of a technical claim. Between these extremes lies a large, underserved region: problems that require genuine multi-step reasoning, grounded in the kinds of tasks humans actually perform, yet whose difficulty is not gated on specialized expertise. GAIA \citep{mialon2023gaia} targets this band, reporting humans at 92\% versus GPT-4-with-plugins at 15\% on questions that are conceptually simple yet require multi-step execution. GIM shares GAIA's anchoring of difficulty in human-accessible reasoning but differs in two key respects: GIM uses rubric-decomposed grading (median 6 criteria per problem) that produces gradient signal where binary verdicts are uninformative, and GIM spans a broader cognitive spectrum---from spatial and temporal reasoning to audience-calibrated communication---rather than concentrating on multi-step tool-assisted execution.

\textbf{The knowledge axis:}
The dominant strategy for keeping benchmarks discriminating has been to escalate the depth or obscurity of the knowledge each question demands. GLUE \citep{wang2018glue} and SuperGLUE \citep{wang2019superglue} aggregated diverse NLU tasks into composite scores; both saturated rapidly as BERT-era models exceeded human baselines. MMLU \citep{hendrycks2021mmlu} scaled to 57 academic subjects with $\sim$16{,}000 multiple-choice questions and remained the de facto standard until frontier models pushed accuracy past 86--90\% \citep{openai2023gpt4, anthropic2024claude, team2023gemini}; MMLU-Pro \citep{wang2024mmlupro} restored some discrimination by expanding answer sets but kept the knowledge-recall multiple-choice format. GPQA \citep{rein2023gpqa} pushed further by sourcing 448 graduate-level ``Google-proof'' physics, chemistry, and biology questions; Humanity's Last Exam \citep[HLE;][]{phan2025hle} extends the construction logic to its limit, soliciting expert-authored questions explicitly engineered to defeat current models. HELM \citep{liang2023helm} retained the same orientation but argued methodologically that single-metric leaderboards obscure capability differences, while AGIEval \citep{zhong2023agieval} reframed evaluation around human-centric standardized exams (SAT, LSAT)---inheriting their contamination risk and difficulty profile. Domain-specific benchmarks---MATH \citep{hendrycks2021math}, GSM8K \citep{cobbe2021gsm8k}, HumanEval \citep{chen2021humaneval}, SWE-bench \citep{jimenez2024swebench}---deepen evaluation within a single capability but are by design narrow. The visual-modality counterpart MMMU \citep{yue2024mmmu} mirrors GPQA's profile: difficulty derives from specialized domain knowledge embedded in images rather than from visual reasoning itself.

These benchmarks differ in scale, format, and modality, but share a common difficulty mechanism: a question is hard because it presupposes knowledge or training that few possess, so a model with broader pretraining coverage scores higher without necessarily reasoning better. This conflation has structural costs. Independent evidence indicates that optimizing against single-axis benchmarks selects for shortcut and pattern-matching strategies over robust reasoning, both in vision \citep{geirhos2020shortcut} and in NLP \citep{mccoy2019right, du2024shortcut, ye2024cleverhans}. \citet{dziri2023faith} show that transformer LLMs solve compositional tasks via linearized subgraph matching, and \citet{press2023measuring} document a compositionality gap that does not narrow with scale. Benchmark decay over time is itself empirically documented: \citet{akhtar2026saturation} systematically study 60 LLM benchmarks and find nearly half exhibit saturation, with private test sets offering no protection but expert-curated benchmarks remaining discriminative substantially longer than crowdsourced ones; \citet{dehghani2021benchmarklottery} additionally show that algorithm rankings are sensitive to which benchmarks are selected. Contamination is a parallel concern: \citet{jacovi2023stop} demonstrate publication-driven leakage risks, \citet{oren2024proving} provide statistical detection tests, \citet{deng2024investigating} document inflated scores, and LiveBench \citep{white2024livebench} addresses contamination via continuous question refresh at substantial operational cost. GIM is designed against this background: rather than escalate along the knowledge axis, it gates difficulty on \emph{integration density}, embeds visual and document context where the visual element is reasoning-load-bearing rather than descriptive (229 of 820 problems), and uses a held-out 205-problem private split as a persistent contamination diagnostic (Section~\ref{section:dataset}).

\textbf{The abstraction axis:}
At the opposite end of the spectrum, abstraction-axis benchmarks remove domain knowledge entirely. \citet{chollet2019measure}'s Abstraction and Reasoning Corpus (ARC-AGI) presents grid-transformation tasks requiring novel-rule inference from few examples, deliberately operationalizing fluid intelligence \citep{cattell1987intelligence} by excluding prior knowledge. The recently released ARC-AGI-3 \citep{arcprize2026arcagi3} extends this lineage to interactive, agentic, language-free environments, reporting humans at 100\% and frontier AI below 1\% as of March 2026 and demonstrating that the abstraction axis is far from saturated. BIG-Bench \citep{srivastava2022bigbench} and BIG-Bench Hard \citep{suzgun2022bigbenchhard} compile over 200 community tasks under no unified cognitive framework, and the AI2 Reasoning Challenge \citep[ARC;][]{clark2018arc} provides multi-step science questions in a single domain.

The abstraction axis is rigorous in isolating $G_f$ from $G_c$ but at a cost. \citet{carroll1993human}'s factor analysis indicates that real-world intellectual performance depends on the interaction of fluid reasoning with acquired knowledge, not on fluid reasoning in isolation. ARC-AGI-3 and GIM cleanly partition the design space rather than overlapping: the former measures agentic exploration in abstract environments, the latter measures single-turn integration of multi-step reasoning in knowledge-grounded contexts.

\section{Dataset Details}
\label{appendix:dataset-details}

\subsection{Design Principles}
\label{appendix:design-principles}

GIM is guided by four design principles:

\begin{enumerate}
    \item \textbf{Difficulty through cognitive demand.} GIM problems require solvers to coordinate multiple cognitive operations---parsing ambiguous specifications, satisfying interacting constraints, tracking state across sequential steps, evaluating the reliability of presented information, and calibrating responses to context. Some problems concentrate difficulty in the number of interacting constraints (integration density); others in the subtlety of a single critical judgment, such as recognizing that a question's premise is flawed. In all cases, difficulty is anchored in what the solver must do with the information, not in how hard the information is to obtain. Authoring guidelines (Appendix~\ref{appendix:prompt-guidelines}) explicitly prohibit reliance on obscure facts, nitpicks, or deep domain expertise gaps; a GIM failure therefore predominantly reflects the reasoning process rather than a gap in domain-specific expertise.

    \item \textbf{Practical grounding.} Problems are drawn from realistic scenarios---analysis, decision-making, synthesis, planning, professional communication---that reflect the tasks users actually apply LLMs to. Some problems test practical interaction directly: calibrating a technical explanation for an expert audience, or evaluating the internal consistency of a professional document. This distinguishes GIM from abstract benchmarks where strong performance does not straightforwardly predict practical utility.

    \item \textbf{Quality through originality and multi-round review.} Every one of the 820 problems is an original composition authored specifically for GIM by a domain expert, with no prompts copied from prior deliverables, competitions, or external corpora. An additional 100 of the 820 problems were contributed by the CRAG-MM~\citep{cragmm2025} and WearVQA~\citep{wearvqa2025} teams (50 each), produced through their respective data-construction pipelines as a collaborative, GIM-specific contribution. These items were authored for GIM and have not been released elsewhere, preserving GIM's contamination-safety guarantee for them as well (Appendix~\ref{appendix:prompt-guidelines}). Each prompt and---where applicable---its rubric pass through multiple rounds of independent human review (Section~\ref{section:dataset}) examining clarity, unambiguous ground truth, sound reasoning paths, and the absence of alternative valid interpretations, alongside state-of-the-art-model validation that the problem is appropriately difficult. Problems that cannot be brought to standard are discarded.

    \item \textbf{Rubric-first, partial-credit grading.} The majority of GIM problems (528 of 820, 64\%) are graded by decomposing the ideal response into independently judged criteria (median 6, mean 7, range 2--80). Each criterion is scored separately by an LLM judge under confidence-weighted aggregation \citep{wei2024simpleqa, patel2025cws}, extending the rubric-decomposed grading paradigm of \citet{arora2025healthbench} from medicine to general reasoning. The remaining 292 problems use exact-answer grading under the same judge framework. This produces gradient signal on multi-constraint problems where a binary correct/incorrect verdict is uninformative.
\end{enumerate}

\subsection{Detailed Taxonomy}
\label{appendix:taxonomy}

Table~\ref{table:prompt-distribution} lists the seven primary categories and the eighteen sub-categories nested within them. Categories code the broad cognitive dimension (e.g., quantitative reasoning, world knowledge); sub-categories pin down the specific operation tested (e.g., calculation vs.\ word-problem framing). The taxonomy partitions only the \emph{primary} category assignment of each prompt; multi-category overlap is captured by the secondary tags discussed in Appendix~\ref{appendix:co-occurrence}.

\begin{table}[!htbp]
\centering
\caption{GIM Taxonomy: Categories and Sub-Categories}
\label{table:prompt-distribution}
\begin{tabular}{|l|l|l|}
\hline
\textbf{Code} & \textbf{Category} & \textbf{Sub-Category} \\ \hline
\multirow{2}{*}{LR} & \multirow{2}{*}{Logical Reasoning} & LR-DD: Deduction \\
 & & LR-IF: Inference \\ \hline
\multirow{3}{*}{QR} & \multirow{3}{*}{Quantitative Reasoning} & QR-CL: Calculation \\
 & & QR-WP: Word Problems \\
 & & QR-DA: Data Analysis \\ \hline
\multirow{2}{*}{SI} & \multirow{2}{*}{Spatial \& Intuitive} & SI-PS: Physical \& Spatial \\
 & & SI-TR: Temporal Reasoning \\ \hline
\multirow{2}{*}{WK} & \multirow{2}{*}{World Knowledge} & WK-GK: General Knowledge \\
 & & WK-DM: Domain Knowledge \\ \hline
\multirow{3}{*}{LN} & \multirow{3}{*}{Language \& Intent} & LN-IT: Interpretation \\
 & & LN-TR: Transformation \\
 & & LN-EP: Expression \\ \hline
\multirow{2}{*}{PR} & \multirow{2}{*}{Procedural} & PR-PL: Planning \\
 & & PR-EX: Execution \\ \hline
\multirow{4}{*}{CT} & \multirow{4}{*}{Constraints} & CT-AV: Avoidance \\
 & & CT-FT: Format \\
 & & CT-BD: Boundaries \\
 & & CT-CO: Optimization \\ \hline
\end{tabular}
\end{table}

\subsection{Bank Composition: Taxonomy, Labels, and Modality}
\label{appendix:taxonomy-sunburst}
\label{appendix:co-occurrence}
\label{appendix:modality}

\begin{figure}[!htbp]
\centering
\includegraphics[width=0.75\textwidth]{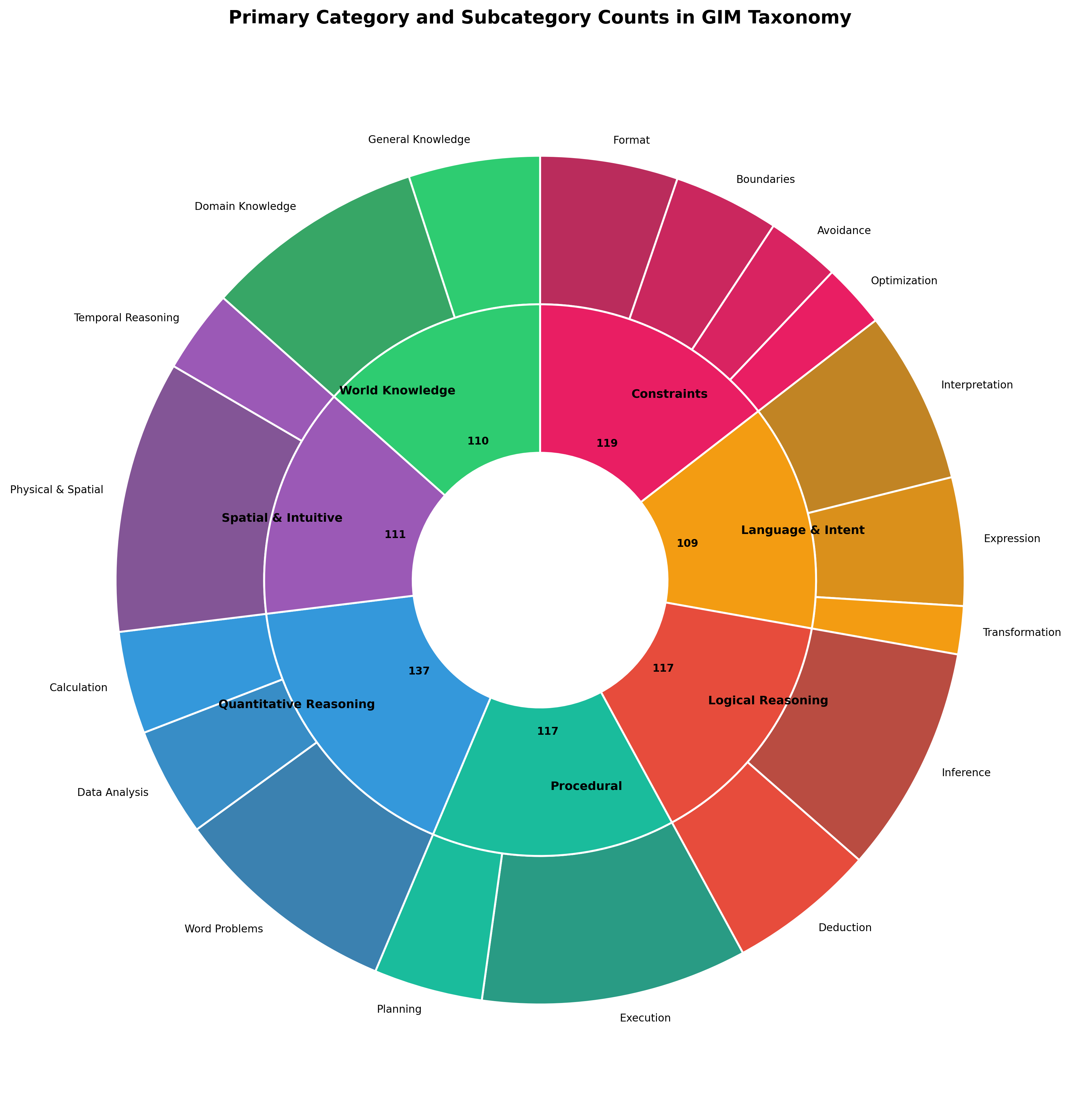}
\caption{Hierarchical visualization of the GIM taxonomy showing the distribution of problems across primary categories and sub-categories. Each prompt is counted once, under its primary assignment.}
\label{fig:sunplot-taxonomy}
\end{figure}

Figure~\ref{fig:sunplot-taxonomy} visualizes the distribution of each problem's \emph{primary} category and sub-category as a hierarchical sunburst. Inner-ring slices correspond to the seven primary categories; outer slices show the sub-category distribution within each. The plot makes the relative weight of each cognitive dimension immediately readable and confirms that the bank covers all 18 sub-categories with no extreme over- or under-representation.

\begin{figure}[!htbp]
\centering
\includegraphics[width=\textwidth]{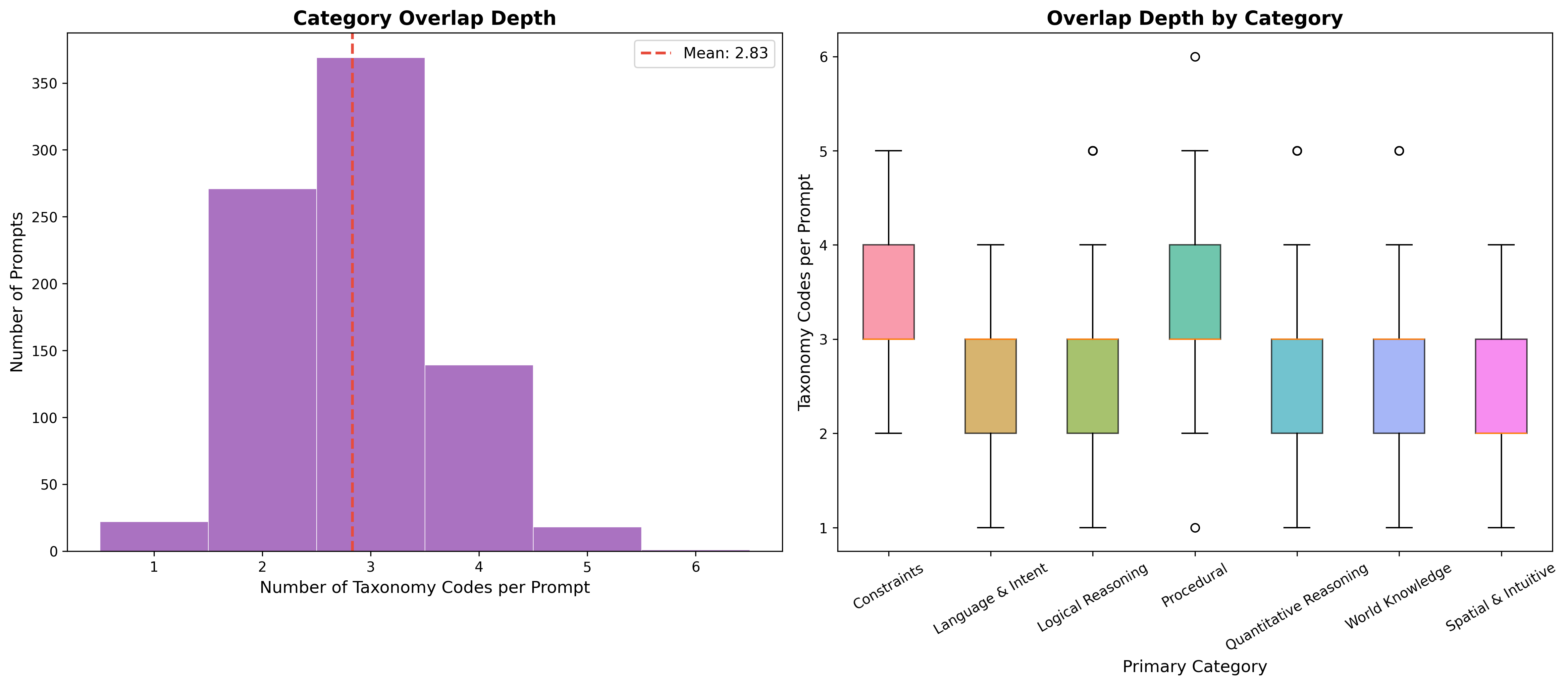}
\caption{Distribution of the number of taxonomy categories assigned per prompt. The mean of 2.83 categories per prompt confirms GIM's design goal of \emph{integrated} reasoning---most problems exercise multiple cognitive dimensions simultaneously rather than testing a single skill in isolation.}
\label{fig:overlap-depth}
\end{figure}

\begin{figure}[!htbp]
\centering
\includegraphics[width=0.75\textwidth]{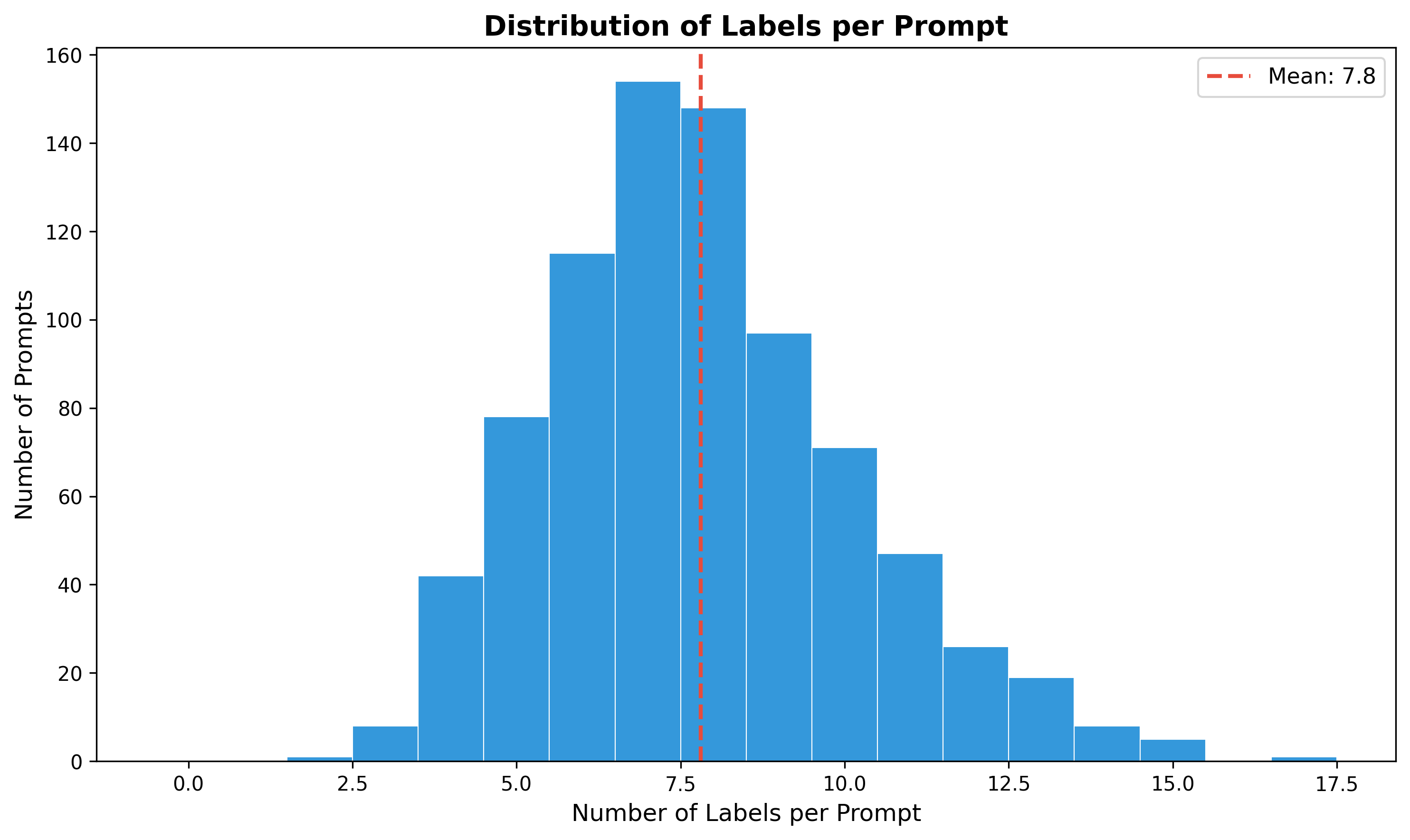}
\caption{Distribution of the number of descriptive labels per prompt. The mean of 7.8 labels per prompt (in addition to the taxonomy categories) illustrates the rich, multi-faceted annotation that supports fine-grained slicing of evaluation results.}
\label{fig:labels-per-prompt}
\end{figure}

GIM is multi-label on two axes: each prompt has one \emph{primary} taxonomy category plus secondary categories (Figure~\ref{fig:overlap-depth}) and a set of free-text descriptive labels orthogonal to the taxonomy (e.g.,\ \texttt{rubrics}, \texttt{verifiable}, \texttt{math}; mean 7.8 per prompt; Figure~\ref{fig:labels-per-prompt}). Together they support fine-grained slicing of evaluation results along axes the seven primary categories do not separate.

\begin{figure}[!htbp]
\centering
\includegraphics[width=0.55\textwidth]{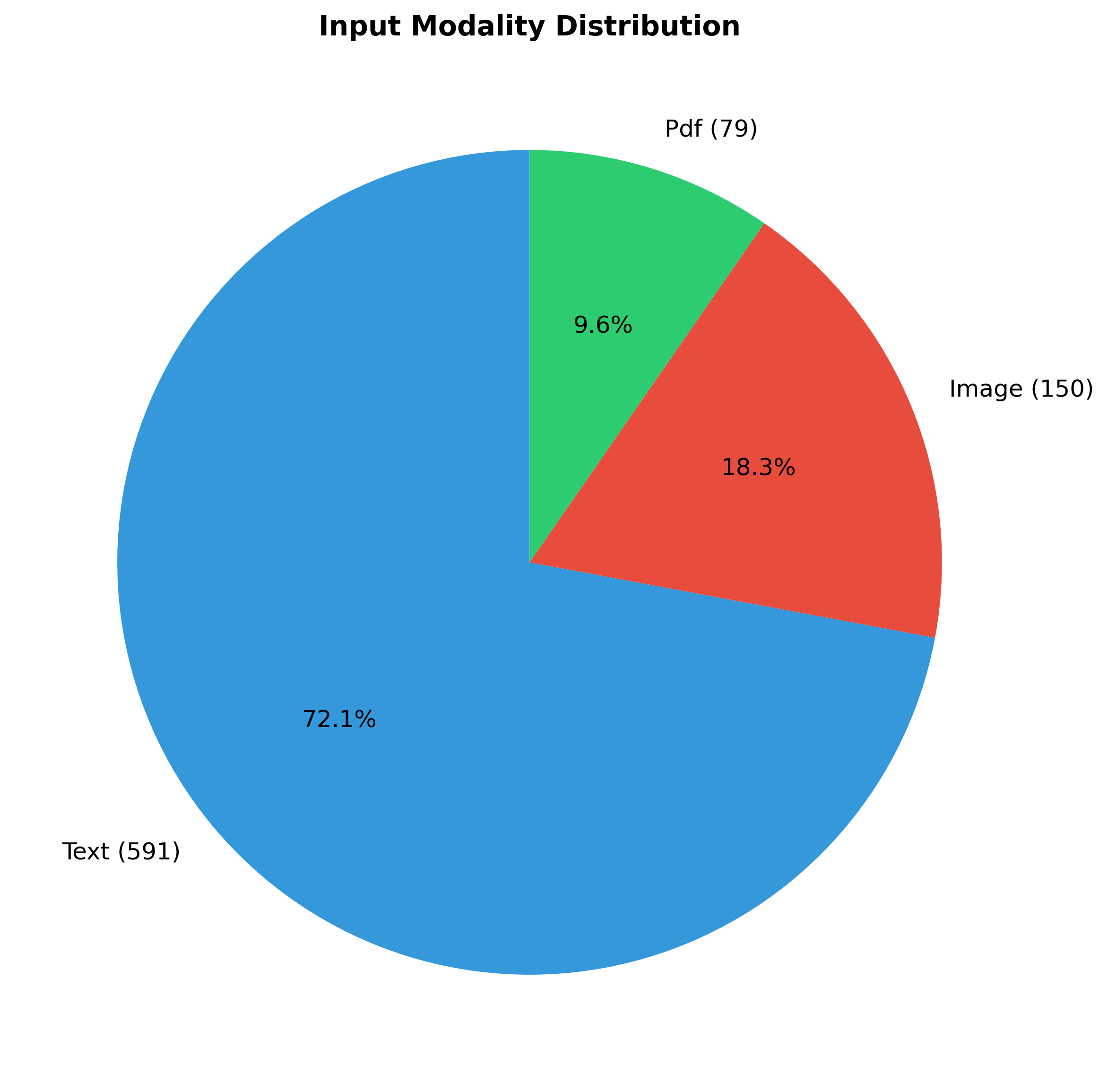}
\caption{Input modality distribution across the 820 GIM problems. The benchmark is text-dominant (72.1\%) but contains a substantial multimodal subset (18.3\% image, 9.6\% PDF) that meaningfully exercises vision-language capabilities.}
\label{fig:modality-distribution}
\end{figure}

\begin{figure}[!htbp]
\centering
\includegraphics[width=\textwidth]{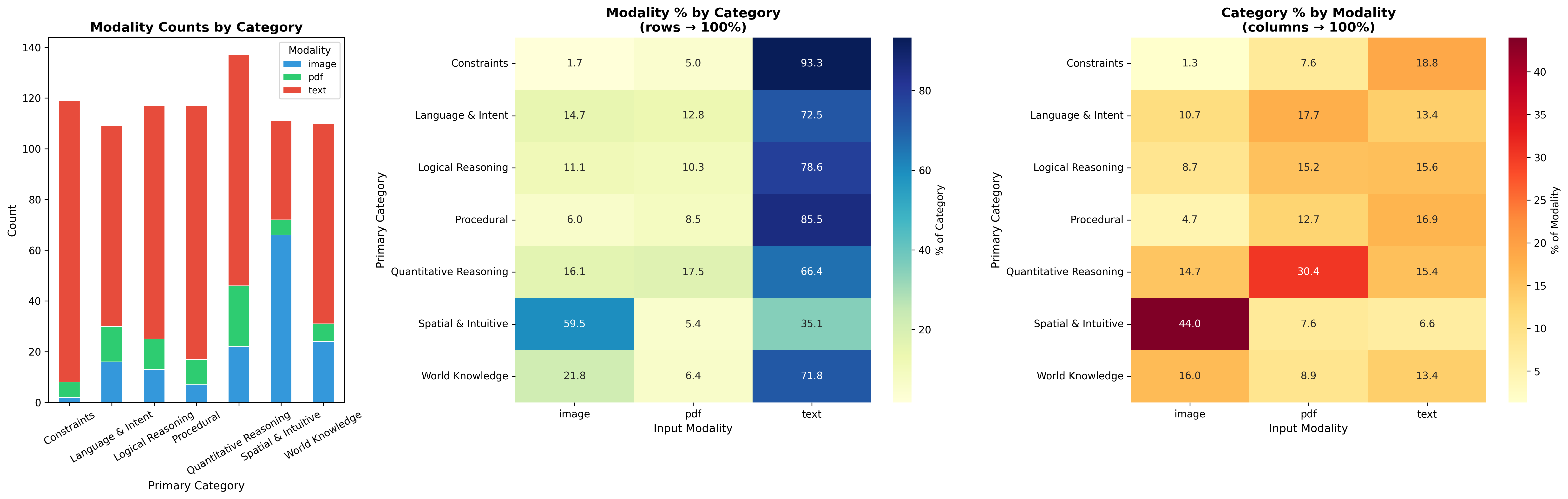}
\caption{Input modality broken down by primary category. Image-heavy prompts concentrate in Spatial \& Intuitive (where 59.5\% of prompts include an image attachment), while text-only prompts dominate Constraints and Procedural categories.}
\label{fig:modality-category-heatmap}
\end{figure}

GIM is also a multimodal benchmark: prompts may contain text, attached images, or attached PDFs. Figure~\ref{fig:modality-distribution} reports the overall split---text-dominant at $72.1\%$, with $18.3\%$ of prompts including an image attachment and $9.6\%$ a PDF---and Figure~\ref{fig:modality-category-heatmap} breaks the same distribution down by primary category, showing that image-heavy prompts concentrate in Spatial \& Intuitive while text-only prompts dominate Constraints and Procedural.

\subsection{Leakage Prevention}
\label{appendix:leakage-prevention}

This appendix expands on the leakage-prevention summary in Section~\ref{subsection:leakage-prevention} and provides statistical evidence that the public/private split is balanced enough to support its use as a contamination diagnostic.

\textbf{Held-Out at Time of Evaluation:} Prior to public release, 100\% of the 820 prompts were held privately and never disclosed outside the core evaluation team. The leaderboard reported in this paper was therefore produced on a fully held-out set: no model evaluated here could have been trained on any GIM problem. The protections below govern how the dataset behaves \emph{after} the public release of the 615-problem split.

\textbf{Public--Private Split:} Of the 820 problems, 615 (75\%) are released publicly to enable community reproducibility and transparent evaluation. The remaining 205 (25\%) are held privately and never released; access is restricted to the core evaluation team. This split serves as a built-in contamination diagnostic: if a model has been trained on leaked public problems, its GIM-615 scores will inflate relative to GIM-205, producing a detectable divergence (Section~\ref{subsection:contamination-diagnostic}).

\textbf{Access-Controlled Private Set:} The 205 private problems are stored in a secure database with strict access controls. Only core evaluation team members may view or modify private prompts and ground-truth answers.

\textbf{LLM-Free Human Sourcing:} Contributors are prohibited from using large language models during the sourcing process, ensuring that prompts are not derived from existing LLM outputs.

\textbf{Aggregate-Only Reporting for Private Problems:} Evaluation results on the private set are published exclusively at an aggregate level---per-category scores, overall accuracy, and statistical summaries. Individual prompt-level results for private problems are never disclosed, preventing reverse-engineering of held-out questions.

\subsection{Maintenance \& Release Plan}
\label{appendix:maintenance}

We intend to maintain GIM as a living artifact, though we do not bind ourselves to a fixed release cadence. The notes below describe how we currently plan to handle updates; specifics may evolve.

\textbf{Hosting.} The public split, the calibrated item parameters $\{a_j, b_j\}$, the centered judge fixed effects $\gamma_k$ for the calibrated judge set, the evaluation harness, and the closed-form scorer are distributed via HuggingFace, with the source mirrored on GitHub. Each release carries a version tag so downstream results can be pinned to a specific item bank and judge calibration.

\textbf{Corrections.} The GitHub issue tracker on the project repo is the canonical channel for reporting errata, ambiguous prompts, suspected mis-scores, or candidate leakage evidence; the dataset card lists a maintainer contact for reports that should not be disclosed publicly. Substantiated issues are folded into subsequent releases.

\textbf{Item updates and recalibration.} We anticipate periodic item-bank refreshes---candidate triggers include credible evidence that a specific public item has been memorized (in which case the item would be retired from the public split and the bank recalibrated on the remaining items), and expanding or revising the calibrated judge set, including the open-weight Gemma~4~31B judge analyzed in Appendix~\ref{appendix:judge-multi}. Because the IRT scorer is well-defined on any subset of the bank, prior $\theta$ values remain interpretable after such updates and we plan to re-score the published leaderboard against the new parameters.

\subsection{Prompt Authoring Guidelines}
\label{appendix:prompt-guidelines}

The following guidelines govern how contributors design problems for the GIM benchmark. These constraints ensure that each prompt tests genuine reasoning ability rather than memorization, obscure knowledge, or ambiguous interpretation.

\begin{itemize}
    \item \textbf{Text-Based Question.} The question itself must be written in text. Prompts may include image or PDF attachments as supporting context, but multimodal capabilities should not be required to read the question itself.

    \item \textbf{Avoid Obscurity.} Problems should not rely on obscure facts, nitpicks, or extremely deep domain expertise gaps. GIM targets reasoning ability, not encyclopedic recall of niche information.

    \item \textbf{Avoid Heavy Math.} Unless otherwise specified, contributors should avoid purely mathematical problems. GIM already maintains sufficient coverage of quantitative reasoning; new prompts should emphasize other cognitive dimensions.

    \item \textbf{Singular Question.} Each prompt must be phrased as a single, standalone question. Multiple sub-questions must not be rolled into one prompt.

    \item \textbf{Originality.} Prompts must not be copied from previous deliverables, competitions, or external sources. All problems must be original compositions. The 100 prompts contributed by the CRAG-MM~\citep{cragmm2025} and WearVQA~\citep{wearvqa2025} teams (50 each) were produced through their internal pipelines as collaborative contributions authored specifically for GIM, and have not been released as part of any other benchmark.

    \item \textbf{Timelessness.} The answer must remain factually correct regardless of when the problem is evaluated. Dynamic references (e.g., ``the current president'') must be anchored to a specific date or ordinal (e.g., ``the 47th US president'').

    \item \textbf{Target Difficulty.} Each prompt must be designed to stump at least two of the three target frontier models used during validation. Problems that all target models solve trivially are revised or discarded.

    \item \textbf{Deterministic.} The prompt must be fully objective and yield a single, verifiable truth. Questions that rely on opinion, interpretation, or subjective metrics are not permitted.
\end{itemize}

\subsection{Rubric Authoring Guidelines}
\label{appendix:rubric-guidelines}

The following guidelines govern the construction of rubrics used in GIM's rubric-graded evaluation. Contributors follow these principles when decomposing an ideal response into individually assessable criteria.

\begin{itemize}
    \item \textbf{Self-Contained.} Include the specific answer value within the criterion text.
    \begin{itemize}
        \item \emph{Bad:} Calculates the length of side AB.
        \item \emph{Good:} Calculates the length of side AB as 12 meters.
    \end{itemize}

    \item \textbf{Clear and Specific.} Specify the exact output format or content required.
    \begin{itemize}
        \item \emph{Bad:} The response is clearly formatted to be easy to read.
        \item \emph{Good:} Uses a bullet-point list.
    \end{itemize}

    \item \textbf{Atomic.} Test only one distinct fact or requirement per criterion.
    \begin{itemize}
        \item \emph{Bad:} Identifies the God of War as Ares and specifies that he is the son of Zeus and Hera.
        \item \emph{Good:} (1) Identifies the God of War as Ares. (2) Specifies that Ares is the son of Zeus and Hera.
    \end{itemize}

    \item \textbf{Mutually Exclusive.} Ensure no two criteria grade the exact same piece of information.
    \begin{itemize}
        \item \emph{Bad:} (1) Bob is the victim. (2) Larry killed Bob.
        \item \emph{Good:} (1) Bob is the victim. (2) Larry was the killer.
    \end{itemize}

    \item \textbf{Timeless.} Use static facts that remain true regardless of the current date.
    \begin{itemize}
        \item \emph{Bad:} States the current US president is Donald Trump.
        \item \emph{Good:} States that the 47th US president is Donald Trump.
    \end{itemize}

    \item \textbf{Ease of Verification.} Criteria should be relatively easy to check. Ensure that the criterion can be verified quickly without requiring excessive calculation or ambiguous interpretation.

    \item \textbf{Collectively Exhaustive.} Ensure all parts of the ideal response are covered so the answer can be reverse-engineered entirely from the rubric.

    \item \textbf{Factually Verifiable.} Criteria must be solvable based on the prompt and supported by official sources, not opinions.

    \item \textbf{Target Quantity.} While quality is prioritized over quantity, aim for a comprehensive breakdown. Each rubric set should ideally contain at least 5 criteria to ensure granularity, up to a maximum of 20.
\end{itemize}

\section{Example Problems}
\label{appendix:examples}

This appendix presents eleven rubric-graded problems selected to illustrate the breadth of cognitive demands GIM tests. The first two reprise the examples from Section~\ref{section:intro}, with full rubrics and scores; the remaining nine are drawn from across the seven primary categories. For each problem we show the prompt, the rubric criteria used for partial-credit grading, and the mean score achieved by four frontier models---Claude~4.6 Opus (High), GPT~5.4 (X-High), Gemini~3.1 Pro (High), and Muse Spark (X-High)---each at its highest available thinking level. Problems that reference attached files or images are noted.

\subsection{Anti-Memorization Through Constraint Modification}
\label{appendix:ex-river-crossing}

\textbf{Category:} Logical Reasoning / Inference (LR-IF) \hfill \texttt{p65920956} \quad \textbf{Frontier avg: 0.796}

\begin{quote}\small
You weigh 60\,kg and need to cross a river on a raft with a 70\,kg limit. You are carrying a 10\,kg wolf, a 10\,kg goat, a 1\,kg cabbage, and a 1\,kg dove. You cannot leave the goat alone with the cabbage or it will eat it, nor can you leave the wolf alone with the goat or it will eat it. However, holding the dove provides 5\,kg of lift. What is the fewest number of river crossings required to get everyone across?
\end{quote}

\textbf{Rubric criteria:}
\begin{enumerate}\small\setlength{\itemsep}{0pt}\setlength{\parskip}{0pt}
    \item Rules out crossing the river in a single turn with the wolf, goat, cabbage, and dove due to weight restrictions.
    \item Rules out crossing the river in two turns, since you would end up on the wrong side of the river.
    \item Demonstrates crossing the river in three turns by transporting either the goat alone or the wolf, cabbage, and dove together on the first turn, returning alone on the second turn, and taking the remaining group across on the third turn.
    \item Concludes that the minimum number of river crossings required to get everyone across is three.
\end{enumerate}

\textit{Commentary:} The classic wolf-goat-cabbage puzzle is among the most heavily represented reasoning problems in training data. This variant adds weight constraints and a dove that provides lift, fundamentally changing the solution structure: the traditional incompatibility constraints are no longer the binding bottleneck.

\subsection{Detecting Anachronism in a Period Document}
\label{appendix:ex-anachronistic-letter}

\textbf{Category:} World Knowledge / General Knowledge (WK-GK) \hfill \texttt{p69379021} \quad \textbf{Frontier avg: 0.482}

\begin{quote}\small
I am a historian and came across the following original letter. I would like to investigate the meeting: which building and on what day was it held?

\medskip
\begin{tabular}{@{}l@{}}
October 4, 1955 \\[4pt]
Mr.\ Julian V.\ Thorne \\
Thorne Industrial Imports, Ltd.\ \\
422 Madison Avenue \\
New York, N.Y.\ \\[4pt]
RE: Acquisition of the Miller Foundry Properties \\
STERLING, WHITTAKER \& ASSOCIATES \\
Attorneys at Law \\
120 Broadway, New York, NY 10271
\end{tabular}
\medskip

Dear Mr.\ Thorne,

Following our luncheon at the Bankers Club yesterday, I have reviewed the preliminary titles for the properties in Pennsylvania.

While the deeds appear to be in order, my associate, Mr.\ Crane, has noted a discrepancy regarding the mineral rights lease dated 1922. It seems the previous owners may have retained subsurface rights that could complicate your proposed expansion.

We have scheduled a meeting with the executors of the Miller estate for next Tuesday at 10:00 A.M.\ here at our offices. I suggest you bring the original articles of incorporation for the new holding company, as we intend to finalize the transfer by the end of the fiscal quarter.

In the meantime, please find the enclosed invoice for the retainer as discussed.

\medskip
Very truly yours, \\
Arthur J.\ Sterling, Esq.\ \\
Senior Partner \\
AJS:ers \\
Encl.
\end{quote}

\textbf{Rubric criteria:}
\begin{enumerate}\small\setlength{\itemsep}{0pt}\setlength{\parskip}{0pt}
    \item Notes that ZIP codes did not exist in 1955.
    \item Notes that the letter is likely a forgery or otherwise not authentic.
    \item Notes that a historian cannot rely on this meeting having taken place.
\end{enumerate}

\textit{Commentary:} The letterhead includes a five-digit ZIP code (10271), but the U.S.\ ZIP code system was not introduced until 1963. A letter dated 1955 containing a ZIP code is anachronistic, indicating the document is not genuine. The correct response is to flag the inconsistency and decline to investigate the meeting, rather than answer the question as posed. Models overwhelmingly accept the letter at face value and attempt to identify the building and date, failing to exercise the epistemic vigilance that the problem demands.

\subsection{Visual Spatial Reasoning}
\label{appendix:ex-maze}

\textbf{Category:} Spatial \& Intuitive / Physical \& Spatial (SI-PS) \hfill \texttt{p35802646} \quad \textbf{Frontier avg: 0.342}

\begin{figure}[h]
\centering
\includegraphics[width=0.35\textwidth]{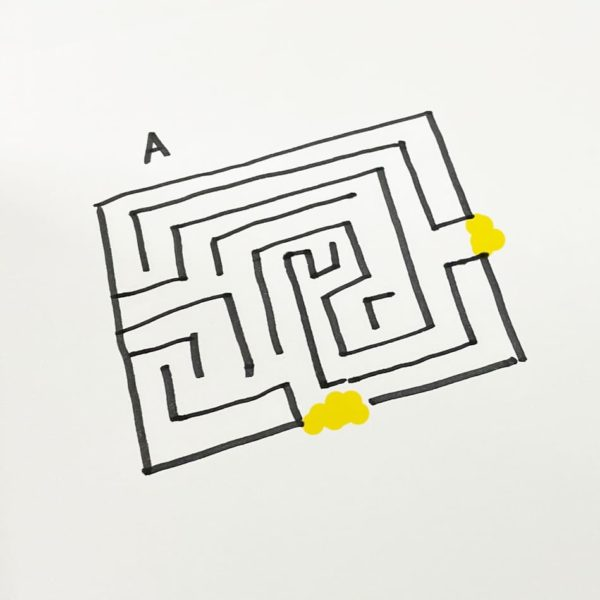}
\caption{Maze image for the visual spatial reasoning problem. The character enters from the arrow on the right side and must reach the green circle.}
\label{fig:maze-example}
\end{figure}

\textit{Note: this problem references the attached maze image (Figure~\ref{fig:maze-example}).}

\begin{quote}\small
In a PC maze game, the character Bradley can only walk forward, and if he reaches a dead end, the game ends. Bradley can only make a choice when he hits a wall (T-junctions), at which he must make a decision to turn either left or right. When he encounters a corner in the maze, he does not make any decisions and simply continues along the path. If Bradley enters the maze in the direction of the arrow on the right side of the image, what decisions must he make to reach the green circle? Format the decisions in a numbered list using the following options: TURN RIGHT, TURN LEFT.
\end{quote}

\textbf{Rubric criteria:}
\begin{enumerate}\small\setlength{\itemsep}{0pt}\setlength{\parskip}{0pt}
    \item Identifies the first decision as TURN RIGHT.
    \item Identifies the second decision as TURN LEFT.
    \item Identifies the third decision as TURN RIGHT.
    \item Identifies the fourth decision as TURN RIGHT.
    \item Identifies the fifth decision as TURN RIGHT.
    \item Formats the decisions in a numbered list.
\end{enumerate}

\textit{Commentary:} The solver must parse the maze image, distinguish T-junctions (where decisions occur) from corners (where the path simply bends), mentally simulate the character's orientation as it changes with each turn, and maintain a consistent left/right frame of reference throughout. Partial credit from the rubric reveals that models often get the first few decisions right but lose track of orientation as turns accumulate.
\subsection{Recognizing Impossibility Across Reference Frames}
\label{appendix:ex-date-line}

\textbf{Category:} Spatial \& Intuitive / Temporal Reasoning (SI-TR) \hfill \texttt{p39075438} \quad \textbf{Frontier avg: 0.675}

\begin{quote}\small
I just celebrated New Years. It's 11:37am on the 1st. I get on a plane, fly for a half hour, then get off. It's now 11:37am on Dec 31. Where am I?
\end{quote}

\textbf{Rubric criteria:}
\begin{enumerate}\small\setlength{\itemsep}{0pt}\setlength{\parskip}{0pt}
    \item Recognizes that crossing the International Date Line eastward gains a calendar day.
    \item Recognizes that even on the shortest possible crossing (e.g., Apia, Samoa to Pago Pago, American Samoa), the flight takes about 30 minutes and you gain 24 hours on the calendar.
    \item Recognizes that the actual clock time would be approximately 12:07pm on Dec~31, not 11:37am---you cannot arrive at an earlier clock time than you departed on a 30-minute flight.
    \item Concludes that the scenario is therefore impossible.
\end{enumerate}

\textit{Commentary:} The problem is designed to look like a lateral-thinking puzzle with a clever geographic answer. Models readily identify the International Date Line mechanism and propose specific routes, but fail to notice that the clock times are inconsistent: a 30-minute flight must advance local time by approximately 30 minutes regardless of date-line effects. The correct answer is that no such trip exists.

\subsection{Audience Calibration}
\label{appendix:ex-chemistry-friend}

\textbf{Category:} Language \& Intent / Expression (LN-EP) \hfill \texttt{p17739818} \quad \textbf{Frontier avg: 0.100}

\begin{quote}\small
My friend decided to become a coal miner after we were offered PhDs in Polymer Chemistry. I finished my first three month report at Warwick. I want to explain to him the problems I am having of synthesizing block-copolymers using ATRP due to the presence of oxygen in the system. Write a paragraph to him to describe this.
\end{quote}

\textbf{Rubric criteria:}
\begin{enumerate}\small\setlength{\itemsep}{0pt}\setlength{\parskip}{0pt}
    \item Avoids focusing on scientific basics or unnecessary detail.
    \item Avoids unnecessary references comparing chemistry issues to the friend's brief coal-mining job.
\end{enumerate}

\textit{Commentary:} Both individuals were offered PhDs in Polymer Chemistry---the friend is a peer who chose a different career, not a layperson. Models consistently patronize: they explain what ATRP is, define ``block-copolymer,'' and draw strained analogies to mining. The rubric rewards treating the friend as the expert he is. The difficulty is entirely social inference---recognizing the implied expertise from a single biographical detail---rather than domain knowledge.

\subsection{Lateral Reasoning}
\label{appendix:ex-biased-die}

\textbf{Category:} Logical Reasoning / Inference (LR-IF) \hfill \texttt{p90067956} \quad \textbf{Frontier avg: 0.323}

\begin{quote}\small
Will and Bill are captains of 2 cricket teams starting a match later today, and I am an umpire. I have a 6-sided die. No one has any coins, so I need to use it to simulate the coin-toss. My normal procedure in such circumstances is just to assign half the sides as ``heads'' and the other half as ``tails'', then get one of the two captains to call heads or tails before rolling the die. The only problem is, the die is biased and I don't know what the bias is---no one does. It's important that the person who calls heads or tails doesn't get a lucky break because of the biased die, especially since the pitch is a bit green. Whoever wins the toss will gain a significant advantage for his side. Is the easiest way to produce a simulated coin toss that's fair to both teams just to proceed as normal?
\end{quote}

\textbf{Rubric criteria:}
\begin{enumerate}\small\setlength{\itemsep}{0pt}\setlength{\parskip}{0pt}
    \item Recognizes that the coin toss in a cricket game usually confers a starting advantage to one of the two teams.
    \item Determines that all that is necessary to confer a starting advantage randomly is that neither captain knows which of heads or tails is more likely prior to the toss.
    \item Infers that the biased die simulating a biased coin confers advantage randomly to one of Bill and Will's teams.
    \item Concludes that the easiest way to ensure a fair simulated coin toss is to proceed as normal.
\end{enumerate}

\textit{Commentary:} Models overwhelmingly propose elaborate fairness procedures (von~Neumann extraction, multiple rolls, etc.)\ rather than recognizing that since neither captain knows the bias, the existing procedure is already fair.

\subsection{Epistemic Vigilance}
\label{appendix:ex-tigers-nest}

\textbf{Category:} World Knowledge / General Knowledge (WK-GK) \hfill \texttt{p34929422} \quad \textbf{Frontier avg: 0.497}

\begin{quote}\small
I am an editor for a small travel magazine. List any and all typos/mistakes one by one in this text, from the end of our writer's Bhutan piece; he got special permission from the government to do a private evening tour of Tiger's Nest, which we are very excited about.

Perched precariously on a sheer granite cliff 5,000 feet above the Paro Valley, Paro Taktsang---better known as the Tiger's Nest---is the undisputed crown jewel of the Last Shangri-La. This 17th-century marvel, built around the cave where Guru Rinpoche is said to have meditated for three years, defies both gravity and the passage of time. To witness its whitewashed walls and gilded roofs shimmering against the jagged Himalayan backdrop is to understand why Bhutan guards its heritage with such fierce, quiet devotion. The journey to the Kingdom of Bhutan remains one of the world's most exclusive pilgrimages, a transition from the frenetic pace of modern life into a realm governed by Gross National Happiness and high-value, low-volume tourism. Most travelers spend their days winding through emerald valleys, visiting ancient dzongs, and adjusting to the thin, crisp air of the high altitudes. It is a land of vibrant prayer flags and silent monasteries, where the `Sustainable Development Fee' ensures that the trails remain pristine and the culture remains remarkably unburdened by the outside world. On a final, golden afternoon, we were granted rare permission to begin our ascent as the day's final pilgrims were descending. Reaching the wooden cafeteria that marks the halfway point, we paused to catch our breath as the valley fell into shadow. From this vantage point, looking directly across the abyss at the monastery's silhouette, the sun began its slow descent. It did not merely set; it seemed to drop directly behind the towering mountain that cradles the Tiger's Nest, casting a halo of fire around the sanctuary before plunging the cliffside into a deep, sacred violet.
\end{quote}

\textbf{Rubric criteria:}
\begin{enumerate}\small\setlength{\itemsep}{0pt}\setlength{\parskip}{0pt}
    \item Notes that Tiger's Nest is approximately 3,000~ft, not 5,000~ft, above the Paro Valley.
    \item Notes that Tiger's Nest is northeast of the midpoint cafeteria, or otherwise argues for why the sunset over the monastery could not be viewed from the cafeteria given their relative locations.
    \item Notes that the sun setting directly behind the monastery is impossible from the described vantage point.
\end{enumerate}

\textit{Commentary:} The passage is fluent and evocative---the kind of polished prose that discourages close scrutiny. Models tend to accept the prose at face value or flag only superficial style issues, failing to cross-reference the spatial claims against real-world geography.

\subsection{Spatial Optimization}
\label{appendix:ex-building-chase}

\textbf{Category:} Quantitative Reasoning / Word Problems (QR-WP) \hfill \texttt{p54121096} \quad \textbf{Frontier avg: 0.521}

\begin{quote}\small
Hecarim is chasing Ashe in a school. The school consists of two parallel buildings, each 10 stories tall. The elementary school building is 60\,m long, and the middle school building is 200\,m long. At both ends of each building, there are staircases connecting the floors. The two buildings are also connected by two corridors: one connects the midpoints of the 5th floor of each building, and the other connects the midpoints of the 8th floor of each building. Each corridor is 20\,m long.

Ashe can run at up to 2\,m/s, while Hecarim can run at up to 3\,m/s. For both of them, the time to climb up one floor is 20\,s, and the time to go down one floor is 5\,s. When moving on a staircase, a player may only travel in whole-floor steps: once they commit to going up (or down) one floor, they must complete that full-floor move, cannot stop midway, and cannot reverse direction between floors. Initially, Hecarim is at an end staircase in the elementary school building on the 9th floor, and Ashe is at the midpoint of the 8th-floor corridor connecting the two buildings.

Assuming they have perfect knowledge of each other's locations at all times and both play optimally (meaning Hecarim tries to catch Ashe as quickly as possible, while Ashe tries to delay capture as long as possible), how long will it take for Hecarim to catch her?
\end{quote}

\textbf{Rubric criteria:}
\begin{enumerate}\small\setlength{\itemsep}{0pt}\setlength{\parskip}{0pt}
    \item Recognizes that on hallways/corridors Hecarim closes the distance at 1\,m/s ($3 - 2 = 1$), while on staircases (same direction) the time gap does not change; therefore Ashe's goal is to maximize time on staircases.
    \item Deduces that Hecarim's optimal first move is to descend from floor~9 to floor~8.
    \item Computes that if Ashe runs toward an end staircase of the middle school building, she is caught on floor~8 at $t = 55$\,s.
    \item Computes that Ashe can reach the far elementary staircase on floor~8 before Hecarim does.
    \item Infers that after Ashe enters that staircase, her only relevant continuations are going all the way down to floor~1 or all the way up to floor~10.
    \item Computes that going down yields capture at $t = 70$\,s.
    \item Computes that going up yields capture at $t = 75$\,s.
    \item Concludes that optimal play yields a capture time of 75 seconds.
\end{enumerate}

\textit{Commentary:} The problem requires building a spatial model of the environment, reasoning about relative velocities across different movement modes, and then solving a minimax pursuit problem with discrete staircase constraints. Each step individually is tractable; the integration is what defeats models.

\subsection{Constraint Satisfaction}
\label{appendix:ex-infinite-monkey}

\textbf{Category:} Constraints / Boundaries (CT-BD) \hfill \texttt{p90665004} \quad \textbf{Frontier avg: 0.312}

\begin{quote}\small
Explain the concept of the infinite monkey theorem in exactly 100 words at five levels: child, teen, college student, graduate student, and expert.
\end{quote}

\textbf{Rubric criteria:}
\begin{enumerate}\small\setlength{\itemsep}{0pt}\setlength{\parskip}{0pt}
    \item Explains the theorem to a child in exactly 100 words.
    \item Explains the theorem to a teenager in exactly 100 words.
    \item Explains the theorem to a college student in exactly 100 words.
    \item Explains the theorem to a graduate student in exactly 100 words.
    \item Explains the theorem to an expert in exactly 100 words.
\end{enumerate}

\textit{Commentary:} The problem requires simultaneously satisfying two orthogonal demands: adapting content sophistication across five distinct audience levels \emph{and} hitting an exact word count for each. Models typically satisfy one constraint at the expense of the other---producing well-calibrated explanations that miss the word count, or hitting exactly 100 words while failing to modulate complexity.

\subsection{State Tracking Through an Information Cascade}
\label{appendix:ex-kennish-family}

\textbf{Category:} Logical Reasoning / Inference (LR-IF) \hfill \texttt{p05435814} \quad \textbf{Frontier avg: 0.461}

\begin{quote}\small
In the Kennish household, there is frantic packing for their flight to Puerto Rico for a two week long visit at their extended family Angelo and Regina Sorrento's house. John and Katherine are a bit flustered after picking up their kids from school, which happened 20 minutes later than usual because Toby and Bay's field hockey game went into overtime. But they get even more flustered after receiving a text from Angelo in their group text: ``I know we talked about this a couple of weeks ago, but Regina and I would really love it if you would bring Kathy's mom's pearl earrings. We really do want to give them to Abby as her first pair of earrings on her birthday tomorrow.''

John exclaimed ``I can't believe they didn't confirm this sooner! It's going to take at least 30 minutes to get those earrings from the storage unit and be back here.'' His wife replied, saying ``Are you kidding? At this hour it will take at least 45 minutes\ldots\ I know mom really would want Abby to have these, but it's getting down to the wire here. Are any of the kids done packing? Maybe they can do a quick round trip before the Uber XL gets here.'' John loudly and sarcastically responded ``Yeah, I am sure the kids are already ready to go---heck they are probably already on the plane! And I bet Daphne's flying them!'' Rolling her eyes, she said ``Very funny John. Can you go ask the kids if anyone is almost done and can quickly get to the storage unit? Tell them the earrings are in that purple bag towards the back of the unit---you know the one. There's no service near it, so make sure to tell them before they go.'' John nodded his head and walked over to the living room where Toby was playing a video game and Bay had just walked in from her room with her yellow suitcase.

``Are either of you guys done packing yet?'' John asked. ``No dad, but we hear Daphne is going to be flying the plane so I think we'll be fine'' quipped Bay while gesturing to the large yellow rectangle next to her and Toby's open suitcase with various clothes scattered about. ``Very funny,'' sighed John. ``Since you're clearly ready, could you run to the storage unit and grab your grandma's earrings? Mom said they're in that pink bag in the back of the unit.'' Toby quickly chimed in ``Um dad isn't it in a blue bag? I think you're thinking of grandma's bracelets.'' John paused for a moment and then said ``Yeah son, you're right! I must have gotten them confused.'' Bay replied ``Okay gotcha, the blue bag! I will leave in just a sec but I need to pack one more thing into my suitcase''. John and Toby chatted about Toby's video game as they heard Bay's heavy suitcase dragging up the stairs in the distance. ``Geez, is she bringing bricks with her?'' Toby snidely remarked. ``Actually she is bringing a lot of textbooks with her---she said she wants to be very prepared for her finals! Though it's so heavy she can only drag it up or down the stairs---I guess we'll be helping her in the airport'' John chuckled. ``Great\ldots'' Toby muttered as he continued his game. Toby sighed that he was sore after a tough game, wishing Daphne had been there to help out.

Bay rummaged upstairs for her missing hat, and then checked out Daphne's room for it. The family dynamic has been much different with her ``semester abroad'', and she was happy to see her soon. Bay longingly looked around a little longer, found her hat, then quietly emerged from upstairs, grabbed the keys from John, and drove out to the storage unit. Bay was about a minute away from the storage unit when Katherine decided she would text her to see if she's okay and to remind her that the earrings are in the purple bag. Bay parked at the storage unit, checked her phone notifications, and headed to their family storage unit 4545. Opening the storage unit door, Bay strolled to the back and saw several colored bags from her grandmother. She immediately saw a tiara poking out of a tiny green bag, a dazzling dress hanging in between that green bag and a purple bag, a small blue bag, and a pink bag with some of her mom's hair rollers in it. Katherine was fully in charge of the storage unit organization and hated whenever people messed around with her systems---Bay rolled her eyes with the thought ``She always knows where everything is''. Bay checked out her notes app that showed the contents of the back shelf in the unit: grandma's tiara, mom's hair rollers, grandma's bracelet, grandma's pearl earrings, and grandma's wedding dress. Then she stopped for a moment, recalled what she was most recently told about the desired bag, and grabbed it. She was a bit stressed after seeing the mess of a storage unit her family had, so she turned off phone notifications before driving back home and listening to a calming playlist she loves.

While Bay was gone, the rest of the family scrambled to get their bags packed. It took them about 40 minutes to finish, and John was getting nervous because the Uber XL just pulled up and Bay wasn't back yet. Katherine had packed her dark gray suitcase and put it by the front door and Toby assured his parents that the kids' bags were by the front door. John tossed his medium sized green duffle bag by the front door, counted five pieces of luggage, and then told them to load up the bags into the trunk while he waited for Bay. Toby carried out his purple bag, Bay's yellow luggage, and his dad's duffle bag. Katherine carried out a dark gray bag and was about to bring out the other bag but realized it was empty and just sitting there from a prior vacation. She shook her head and muttered ``Can't believe Daphne left that there\ldots'' and then briskly walked to the Uber XL to join Toby.

A couple minutes later, Bay pulled up to the driveway, hopped out with her bag, and saw her dad waiting outside the front door. Bay asked how much time they had before they needed to leave, and John replied that they needed to leave immediately. He reassured her that all the bags were packed in the car and that the family was waiting in the car. ``Thank you for bringing the earrings, your aunt and uncle will love it---and Abby will look adorable in them!'' Bay smiled at him, locked her car, and then they loaded up into the car. On the ride to the airport, the kids talked about how to present the gift to Abby. The Sorrentos were going to pick them up and they were pretty sure Abby would be with them. The kids bickered a little bit, then decided that whoever gets from the airplane to give Angelo a hug first would get to present the gift. After bickering, John and Katherine asked them for a bit of silence so they could relax after their stressful day.

Toby initiated a text to the sibling group chat and continued talking about the plan to present the gift to Abby, bragging that his track and field hockey experience would make him the clear winner. Daphne replied that she's pretty sure she will hug Angelo first and sent a winky face. Bay reacted to her message with a laughing emoji and said that she'll have a hard time winning if she's piloting the plane like dad said. Toby and Daphne both sent ``haha'' in the chat. Bay then said that she's confident she'll win because she'll convince mom to hold Toby in a long, awkward hug to give her a head start. After this, they arrived at the airport and Toby said ``time to fly everyone.''

Finally, the Kennishes arrived in Puerto Rico, stood outside, and watched as the Sorrentos pulled up. Following Bay's instructions, Katherine attempted to pull Toby in for a tight squeeze as Bay sprinted towards Angelo. She didn't hold on to Toby very long before he broke away and started sprinting as well. Angelo was startled by the kids running at him, but then braced himself for the impacts. Daphne grabbed onto him first and stuck out her tongue at her siblings, with Toby narrowly beating Bay to a big bear hug. Angelo chuckled as his nieces and nephews squeezed him tight.

At the end of the scenario, who presents what to Abby? Format the answer as: [Name] presents [item] to Abby. Refer to the item name with Bay's note app's terminology. Explain your reasoning.
\end{quote}

\textbf{Rubric criteria:}
\begin{enumerate}\small\setlength{\itemsep}{0pt}\setlength{\parskip}{0pt}
    \item Recognizes that Daphne did not fly with John, Katherine, Toby, and Bay to Puerto Rico.
    \item Acknowledges that Daphne is not an eligible sibling in the contest to hug Angelo first.
    \item States that Toby won the sibling contest.
    \item States that Toby presents an item to Abby.
    \item States that Abby receives grandma's bracelet.
    \item States that grandma's pearl earrings are in the purple bag.
    \item States that grandma's bracelet is in the blue bag.
    \item States that Bay grabs the blue bag from the storage unit.
    \item Avoids stating that Bay saw Katherine's text reminding her the earrings are in the purple bag.
\end{enumerate}

\textit{Commentary:} Information about which bag to retrieve is corrupted as it passes through a telephone-game chain. The reader must recognize that Daphne is on a semester abroad (implied, never stated outright) and therefore not on the flight, making her ineligible despite winning the hug race. The problem tests sustained state tracking across a long, naturalistic narrative where each link in the chain is individually reasonable but the cumulative drift is decisive.

\subsection{Procedural Fidelity Under Compound Conditionals}
\label{appendix:ex-old-english}

\textbf{Category:} Procedural / Execution (PR-EX) \hfill \texttt{p33940796} \quad \textbf{Frontier avg: 0.522}

\begin{quote}\small
``Sir Aldric rode into the mist shrouded village at dawn, his polished armor glinting in the pale light. Word had reached King Marcellus that the nearby harvest festival was in peril, the river banks were swelling, threatening to flood the fields that sustained the villagers. Queen Isolde, wise and compassionate, insisted that they send aid at once, for the people's welfare was the greatest responsibility of the crown.''

Rewrite the above story, making sure every second word is rewritten using Old English spelling (a reasonable translation or synonym; use any form used between the 10th and 15th century), but only if the word prior starts with a vowel. Words with apostrophes in the middle can be considered to be one word (ie: people's).
\end{quote}

\textbf{Rubric criteria:}
\begin{enumerate}\small\setlength{\itemsep}{0pt}\setlength{\parskip}{0pt}
    \item Identifies that the 10th word ``dawn'' starts with a consonant, therefore makes no change to the 11th word ``his''.
    \item Identifies that the 20th word ``had'' starts with a consonant, therefore makes no change to the 21st word ``reached''.
    \item Identifies that the 30th word ``in'' starts with a vowel, therefore replaces the 31st word ``peril'' with an Old English equivalent such as ``f\ae r''.
    \item Identifies that the 40th word ``the'' starts with a consonant, therefore makes no change to the 41st word ``fields''.
    \item Identifies that the 50th word ``compassionate'' starts with a consonant, therefore makes no change to the 51st word ``insisted''.
    \item Identifies that the 60th word ``people's'' starts with a consonant, therefore makes no change to the 61st word ``welfare''.
    \item Preserves all other text unchanged from the original passage.
\end{enumerate}

\textit{Commentary:} The solver must chain three operations---accurate word counting, vowel/consonant classification of the preceding word, and Old English translation---where an error in any step cascades. The rubric samples specific positions across the passage (words 10--11, 20--21, 30--31, etc.) to detect systematic drift in counting or inconsistent rule application. Models frequently miscount words, apply the rule to every second word regardless of the vowel condition, or rewrite words that should be preserved.

\section{Evaluation Pipeline Details}
\label{appendix:eval-details}

This appendix provides implementation details for the evaluation methodology summarized in Section~\ref{section:methodology}.

\subsection{Implementation Stack}
GIM is implemented as an \textbf{Inspect AI}~\citep{inspect2024} task plugin, with all Python dependencies pinned via \texttt{uv}~\citep{astral2024uv} for reproducible environments. A single CLI command launches an evaluation against any provider supported by Inspect AI; all per-provider sampling, retry, and rate-limit behavior is delegated to Inspect AI's harness.

\subsection{Dataset Format}
The GIM dataset is stored in HuggingFace \texttt{datasets} format. A \texttt{DatasetDict} provides named splits: \texttt{public} (615 problems) and \texttt{private} (205 problems held out for contamination control; Section~\ref{subsection:leakage-prevention}). Each record contains the prompt text, a unique prompt ID, comma-separated taxonomy labels, an optional golden answer, an optional list of rubric strings, solution reasoning, citations, and relative paths to media attachments (images or PDF documents).

\subsection{Modality Filtering}
A modality filtering layer allows evaluators to select specific sample subsets---text-only, image-bearing, document-bearing, or all modalities. Models that do not support a given modality are not penalized by problems they are architecturally unable to process.

\subsection{Multimodal Content Handling}
For samples with attachments, the loader constructs multimodal \texttt{ChatMessageUser} objects containing interleaved \texttt{ContentImage}, \texttt{ContentDocument}, and \texttt{ContentText} parts. All images are resized to a maximum of 3\,MB to stay within API limits. Attachment paths are resolved against a configurable base---either a local directory or a remote URI (e.g., Google Cloud Storage)---enabling evaluation against cloud-hosted media without local copies.

\subsection{Fault Tolerance}
A generation wrapper intercepts API failures (timeouts, rate limits, context-length overflows, server errors) and converts them into empty completions rather than aborting the evaluation. Failed generations are recorded as missing data; under the IRT scoring described in Section~\ref{section:results}, missing samples drop out of the ability estimate without biasing it. Each API call is retried up to three times with exponential backoff.

Beyond the in-run retries above, the researchers manually launched a second-pass run at a later time that re-attempted only the prompts that returned errors in the first pass. Results are merged, with second-pass samples replacing first-pass failures where the retry succeeded.

\subsection{Multi-Epoch Evaluation}
Each sample is presented to the model $k$ times ($k=5$ for reported results), and the resulting per-sample scores feed the IRT calibration described in Section~\ref{section:results}. Building on the principle underlying self-consistency \citep{wang2023selfconsistency}, multi-epoch sampling reduces variance from non-deterministic infrastructure and addresses a practical constraint: several frontier systems, including OpenAI's o1 and o3 families \citep{openai2024o1systemcard} and Google's Gemini models with extended thinking, do not allow setting \texttt{temperature} when reasoning is active.

\subsection{Chain-of-Thought and Temperature}
GIM presents every problem as a single-turn user message. We do not explicitly enable any additional tools (e.g., web search, code execution), relying on each provider's default tool configuration so that each model is measured under its out-of-the-box deployed setup. GIM does not inject explicit chain-of-thought prompts into the problem statement. Models with built-in thinking capabilities are always evaluated with reasoning enabled. For models that support configurable thinking budgets, we evaluate at multiple levels (e.g., high vs.\ low) and report results for each.

Because reasoning-enabled models universally prohibit explicit temperature control, we do not set a generation temperature for any evaluated model. We accept each provider's default sampling behavior and rely on multi-epoch averaging ($k=5$) to stabilize scores. The LLM judge also operates in reasoning mode; consistency in scoring is enforced through structured output constraints.

\subsection{Hybrid Scoring Details}
GIM uses two complementary scoring strategies, following \citet{kim2024prometheus, ye2024flask, liu2023geval}. Routing is deterministic at sample load time: if the record carries one or more rubric strings, the sample is rubric-graded; otherwise it falls back to exact-answer grading.

\begin{enumerate}
    \item \textbf{Rubric-graded scoring.} Problems with structured rubrics are decomposed into independently assessable criteria \citep{min2023factscore}. Each rubric item is scored by the LLM judge as a $(s_i, c_i)$ pair, where $s_i \in [0,1]$ is the per-criterion score and $c_i \in [0,1]$ is the judge's self-reported confidence. The per-sample score is the confidence-weighted mean
    \begin{equation*}
    \frac{1}{n}\sum_{i=1}^{n} s_i \cdot c_i,
    \end{equation*}
    rewarding partially correct reasoning with confidence-weighted partial credit~\citep{kadavath2022language}. Rubric authoring guidelines (Appendix~\ref{appendix:rubric-guidelines}) ensure that criteria are atomic, mutually exclusive, and collectively exhaustive so that the per-criterion mean is meaningful as an overall sample score.

    \item \textbf{Exact-answer scoring.} Problems with a definitive golden answer are scored by comparing the model's output against the target, accounting for representational equivalence (e.g., $0.5 = \tfrac{1}{2}$), format variations, and null equivalences (e.g., \texttt{N/A}~$\equiv$~empty~$\equiv$~\texttt{none}).
\end{enumerate}

\subsection{LLM-as-Judge Implementation}
Both scoring strategies use an LLM judge operating under structured output constraints \citep{zheng2023judging}. The judge receives a Pydantic-defined response schema specifying the exact JSON structure it must return (\texttt{ExactAnswerJudgment} or \texttt{RubricJudgment}), enforcing type safety and eliminating parsing failures. Structured output constraints also mitigate known biases in LLM judges, such as verbosity and positional bias \citep{wang2023llmfairevaluators}. Every response is graded by all five judges (Section~\ref{subsection:hybrid-scoring})---Gemini~3~Flash, Gemini~3.5~Flash, GPT-5.4~Mini, Claude~4.5~Haiku, and Gemma~4~31B---each receiving the same structured-output call. Each judge call is retried up to three times with exponential backoff (maximum 30-second wait).

\subsection{Rubric Design Principles}
Each rubric decomposes the ideal response into atomic, independently assessable criteria (authoring guidelines in Appendix~\ref{appendix:rubric-guidelines}). Rubric items are designed to be: (1) \textbf{Clear}---unambiguous in what constitutes meeting the criterion; (2) \textbf{Atomic}---each item tests exactly one aspect; (3) \textbf{Measurable}---evaluable as met or unmet with partial credit; (4) \textbf{Independent}---scores do not depend on other items. Rubrics include both positive and negative criteria.

\subsection{Diagnostic Metrics}
Beyond the primary reward score, the pipeline emits per-sample diagnostic metadata:
\begin{itemize}
    \item \textbf{Per-modality scores:} reward broken down by text, image, document, and mixed modalities.
    \item \textbf{Per-label scores:} reward exploded across the multi-label taxonomy for per-category performance profiles.
    \item \textbf{Generation diagnostics:} success/failure rates by error type (timeout, rate limit, context length overflow).
    \item \textbf{Judge diagnostics:} per-rubric-item scores, confidences, and explanations.
\end{itemize}

\section{LLM Judge Prompts}
\label{appendix:judge-prompts}

This appendix reproduces the exact prompts used by the LLM judge in GIM's evaluation pipeline. Template variables (enclosed in braces) are populated at scoring time with the relevant content for each sample.

\subsection{Exact-Answer Judge Prompt}
\label{appendix:exact-answer-prompt}

The following prompt is used when a sample does not have rubrics and is scored by comparing the model's output against a golden answer.

\begin{small}
\begin{verbatim}
Task Description
I have three inputs for you:
    1) Question: The first input is the question.
    2) Gold target: The second input is the expected answer.
    3) Predicted answer: The third input is the answer to verify,
       which may contain reasoning and should end with a specific
       format (e.g., "Final answer: the final answer is",
       "Answer:", or "Answer is").

Task Requirements
Your task is to:
    1) Check if a real answer is generated.
    2) If it does, extract the final answer and compare it with
       the golden target.
    3) Consider answers as the same if they are represented in
       different formats (e.g., 0.5 and 1/2) or have an absolute
       difference of less than 0.01 (e.g., sqrt(2) and 1.41).
    4) Handle non-numeric answers, such as booleans or lists of
       strings, where order matters.
    5) Account for choice questions where the expected answer is
       a letter (A, B, C, D) and the real answer is a string
       mentioned in the question.
    6) Treat N/A as equivalent to null, none, or empty in program
       analysis outputs.
    7) Identify answers that involve context, such as "the answer
       is something shown above".

Here is a new example. Grade the predicted answer as one of:
CORRECT, INCORRECT.
```
Question: {question}
Gold target: {answer}
Predicted answer: {predicted_answer}
```

The gold target is the ground truth -- do not question its
correctness. Even if the predicted answer appears more reasonable
or accurate than the gold target, grade it as INCORRECT if it
does not match the gold target.
First provide the explanation, then grade the predicted answer.

# Confidence

Also provide a confidence value between 0.0 and 1.0 indicating
how reliable your score is. 1.0 means you are fully certain;
0.0 means the score is a guess.
\end{verbatim}
\end{small}

The judge must return a structured JSON response containing three fields: \texttt{explanation} (string), \texttt{grade} (\texttt{CORRECT} or \texttt{INCORRECT}), and \texttt{confidence} (float in $[0, 1]$).

\subsection{Rubric-Graded Judge Prompt}
\label{appendix:rubric-prompt}

The following prompt is used when a sample has rubrics. It is invoked once per rubric item, and the per-item scores are aggregated into a confidence-weighted mean (Section~\ref{subsection:hybrid-scoring}).

\begin{small}
\begin{verbatim}
You are a helpful grader, and will be given a conversation
including a prompt, expected golden answer, a model response
and a rubric item.
Your job is to look at the conversation and the rubric string,
and determine whether the response meets the criteria specified
in the rubric string.
Please note that the rubric may ask the information from either
prompt, golden response or model response, and itself may
contain some ground truth.

# Golden response
{answer}

# Model Response
{model_response}

# Rubric string
{rubric_string}

When you finish your evaluation, you'll need to provide a final
score as a float between 0.0 and 1.0:

# Scoring

- 1.0 means the rubric criterion is fully met.
- 0.0 means the criterion is not met at all.
- For partial credit, use a proportional value
  (e.g., 0.6 for 6 out of 10).
- If you don't have enough information to make a judgment,
  return 0.0.

# Confidence

Also provide a confidence value between 0.0 and 1.0 indicating
how reliable your score is. 1.0 means you are fully certain;
0.0 means the score is a guess.

# Important Note on Criteria Examples

When a criterion includes phrases like "such as,"
"for example," or "including," the response doesn't need to
cover every example listed to meet the criterion. For instance,
if a criterion states, "States that oral iron supplements can
lead to unpleasant gastrointestinal side effects such as nausea,
vomiting, and constipation," and the response only mentions
"oral iron supplements can lead to unpleasant gastrointestinal
side effects such as cramps," it still satisfies the criterion.
\end{verbatim}
\end{small}

The judge must return a structured JSON response containing three fields: \texttt{explanation} (string), \texttt{score} (float in $[0, 1]$), and \texttt{confidence} (float in $[0, 1]$).

\section{Multi-Judge Robustness}
\label{appendix:judge-multi}

GIM's released scores come from a \emph{joint judge-aware} IRT calibration over five LLM judges (\S\ref{subsection:irt-model}): a single shared item bank $\{a_j, b_j\}$, one ability $\theta_i$ per model configuration, and a centered fixed effect $\gamma_k$ for each judge. This appendix asks whether that modeling choice is a good one. The motivation for $\gamma_k$ is \emph{decoupling the score from the judge}: by publishing one item bank together with five judge offsets, a researcher who wants to place a new model on the leaderboard can re-score it with \emph{any single one} of the calibrated judges and still recover a comparable $\theta$ --- they are not tied to the one judge we happened to run, and they do not have to run all five. We quantify how well this works and compare it to the two alternatives it replaces: a single fixed judge, and a panel/vote over all five.

\subsection{The Judge-Aware Model and the Five Judges}
\label{appendix:judge-multi-model}

Each of five judges --- \textbf{Gemini~3~Flash}, \textbf{Gemini~3.5~Flash}, \textbf{GPT-5.4~Mini}, \textbf{Claude~4.5~Haiku}, and the open-weight \textbf{Gemma~4~31B} --- scored every response on all 820 prompts and 53 model$\times$thinking configurations across five epochs. After averaging repeated epochs within each (model, prompt, judge) cell and dropping empty completions, this yields 203{,}800 observed cells. The model
\begin{equation*}
y_{ijk} = a_j(\theta_i - b_j) + \gamma_k + \varepsilon_{ijk}, \qquad \textstyle\sum_k \gamma_k = 0
\end{equation*}
is fit jointly; $\gamma_k$ is the judge's leniency in logit-score units (positive $=$ more lenient). The fitted effects are $+0.560$ (Gemini~3~Flash), $+0.485$ (Gemma~4~31B), $-0.208$ (GPT-5.4~Mini), $-0.296$ (Gemini~3.5~Flash), and $-0.541$ (Claude~4.5~Haiku) --- an inter-judge spread of $1.10$ logits. Re-fitting this model from scratch reproduces the released bank: the judge effects are identical, item discrimination and difficulty correlate at $r = 1.000$ with the released $\{a_j, b_j\}$, and abilities at $r = 0.9998$.

\subsection{Single-Judge Ability Recovery}
\label{appendix:judge-multi-recovery}

\begin{figure}[t]
    \centering
    \includegraphics[width=0.95\textwidth]{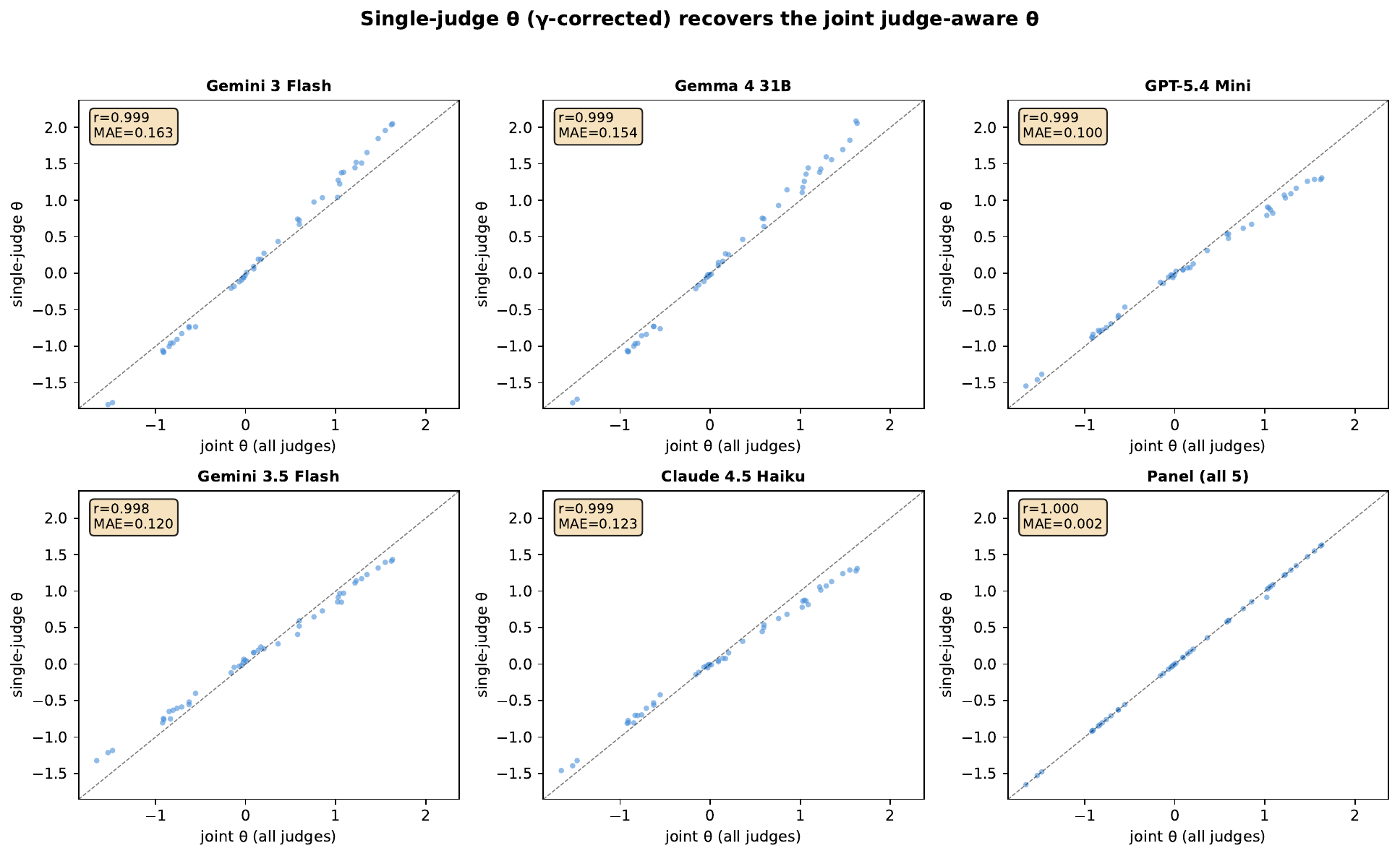}
    \caption{Single-judge ability (estimated from one judge's scores against the released $\{a_j, b_j\}$ with the $\gamma_k$ correction) vs.\ the joint judge-aware $\theta$, per judge and for the five-judge panel mean. Each point is one of 53 configurations; the dashed line is $y = x$.}
    \label{fig:judge-recovery}
\end{figure}

The central question: using the released $\{a_j, b_j, \gamma_k\}$, if a researcher scores a model with \emph{one} judge $k$, do they recover the joint ability? Each configuration's ability is re-estimated from a single judge's scores with the closed-form weighted least-squares scorer,
\begin{equation*}
\hat\theta_i^{(k)} = \frac{\sum_j a_j\,\big[(y_{ijk} - \gamma_k) + a_j b_j\big]}{\sum_j a_j^2},
\end{equation*}
subtracting that judge's offset $\gamma_k$, and compared to the joint $\theta_i$ (Figure~\ref{fig:judge-recovery}, Table~\ref{tab:judge-recovery}). Every single judge recovers the joint ability tightly: Pearson $r \geq 0.998$, Spearman $\rho \geq 0.997$, mean absolute error $0.10$--$0.16$ logits, and a maximum rank displacement of \emph{at most three} places across all 53 configurations. A researcher can therefore place a new model on the published scale with a \textbf{single} judge run. Averaging all five judges (the panel) is essentially exact ($r = 0.9999$, MAE $0.002$, zero rank change), but as \S\ref{appendix:judge-multi-alternatives} discusses it costs five judge runs for a ranking that one judge already reproduces.

\begin{table}[t]
\centering
\caption{Single-judge $\theta$ recovery against the joint judge-aware scale, using the released $\{a_j, b_j, \gamma_k\}$. \emph{Rank shift} is the largest change in leaderboard position over the 53 configurations. The last two columns drop the $\gamma_k$ correction (raw single judge); the resulting bias equals the judge's leniency.}
\label{tab:judge-recovery}
\small
\begin{tabular}{lrrrrrrr}
\toprule
Judge & $\gamma_k$ & $r$ & $\rho$ & MAE & bias & rank shift & bias (no $\gamma$) \\
\midrule
Gemini 3 Flash   & $+0.560$ & 0.999 & 0.998 & 0.163 & $+0.043$ & 2 & $+0.356$ \\
Gemma 4 31B      & $+0.485$ & 0.999 & 0.997 & 0.154 & $+0.045$ & 2 & $+0.317$ \\
GPT-5.4 Mini     & $-0.208$ & 0.999 & 0.997 & 0.100 & $-0.064$ & 3 & $-0.180$ \\
Gemini 3.5 Flash & $-0.296$ & 0.998 & 0.998 & 0.120 & $+0.015$ & 3 & $-0.151$ \\
Claude 4.5 Haiku & $-0.541$ & 0.999 & 0.997 & 0.123 & $-0.051$ & 3 & $-0.354$ \\
\midrule
Panel (all 5)    & $0.000$  & 1.000 & 1.000 & 0.002 & $-0.002$ & 0 & $-0.002$ \\
\bottomrule
\end{tabular}
\end{table}

\subsection{Necessity of the Judge Fixed Effect}
\label{appendix:judge-multi-why-gamma}

\begin{figure}[t]
    \centering
    \includegraphics[width=0.95\textwidth]{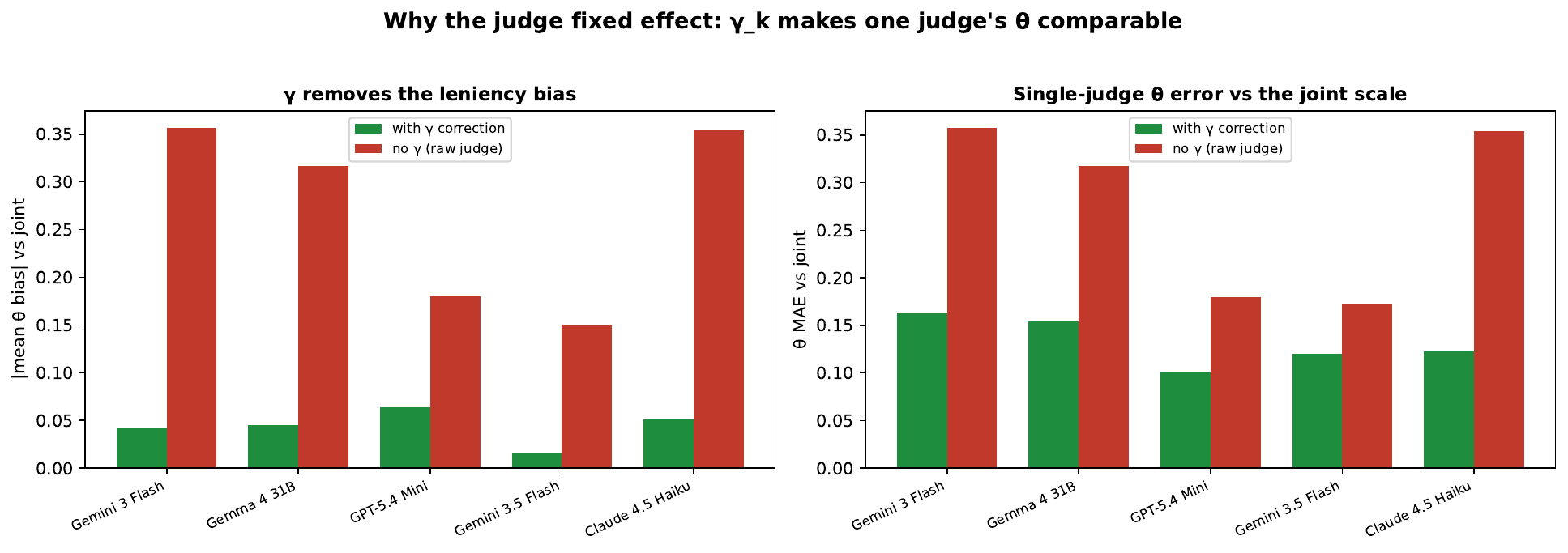}
    \caption{The role of the judge fixed effect. \textbf{Left:} absolute $\theta$ bias vs.\ the joint scale when a model is scored by each single judge. Without the offset (red) the bias equals that judge's leniency --- the error a judge-agnostic (single-judge) bank would suffer the moment the judge is swapped; with the $\gamma_k$ correction (green) it nearly vanishes. \textbf{Right:} the corresponding $\theta$ MAE.}
    \label{fig:judge-gamma}
\end{figure}

The judges differ in leniency by up to $1.10$ logits --- about a third of the full ability range --- and that gap is what forces the judge term. Consider the simplest design, a bank calibrated on a \emph{single} judge: that judge's leniency is silently baked into the item parameters $\{a_j, b_j\}$. A researcher who later re-scores a model with a \emph{different} judge has no way to separate ``my model is stronger'' from ``my judge is more lenient'' --- the whole leniency gap is added straight to $\theta$ (Figure~\ref{fig:judge-gamma}, red: up to $0.36$ logits, MAE $0.15$--$0.36$). Such a bank is usable \emph{only} with the exact judge it was built on; swapping in any other judge silently re-ranks the board. A single-judge leaderboard is therefore not judge-portable --- you cannot change the judge.

The fixed effect removes this lock-in by estimating each judge's offset $\gamma_k$ explicitly and centering it ($\sum_k \gamma_k = 0$). The leniency becomes a known constant the scorer subtracts (Figure~\ref{fig:judge-gamma}, green: bias $\leq 0.06$), so the five calibrated judges become interchangeable --- which is precisely the single-judge recovery of \S\ref{appendix:judge-multi-recovery}: pick any of the five and a model lands on the same scale.

This interchangeability has a hard boundary: it covers \emph{only} the five judges that entered the joint fit. A judge with no estimated $\gamma_k$ cannot be placed on the released scale at all --- its offset is unknown, so its raw leniency would resurface as bias. Admitting a new judge means first calibrating its $\gamma_k$ (re-scoring the item bank with it and re-fitting the centered offsets). The released artifact thus buys a free \emph{choice} of judge within the calibrated panel, not portability to an arbitrary judge.

\subsection{Comparison to Alternatives}
\label{appendix:judge-multi-alternatives}

The judge-fixed-effect design sits between two simpler options:
\begin{itemize}
    \item \textbf{Single fixed judge.} One bank tied to the one judge we ran. As shown in \S\ref{appendix:judge-multi-why-gamma} it is judge-locked: a different judge re-ranks the board, so everyone must run our exact (proprietary) judge.
    \item \textbf{Panel / majority vote.} Score every model with all five judges and aggregate. This recovers the joint ability essentially perfectly (Table~\ref{tab:judge-recovery}, last row), but to add a single model to the leaderboard a researcher must run \emph{all five} judges; one run is not enough.
    \item \textbf{Judge fixed effect (ours).} One released bank plus five offsets. Any single calibrated judge places a model on the shared scale at $\approx 0.13$ logit MAE and $\leq 3$ rank places --- one judge run, free choice of judge, and rankings essentially unchanged.
\end{itemize}

\subsection{Inter-Judge Agreement}
\label{appendix:judge-multi-agreement}

\begin{figure}[t]
    \centering
    \includegraphics[width=\textwidth]{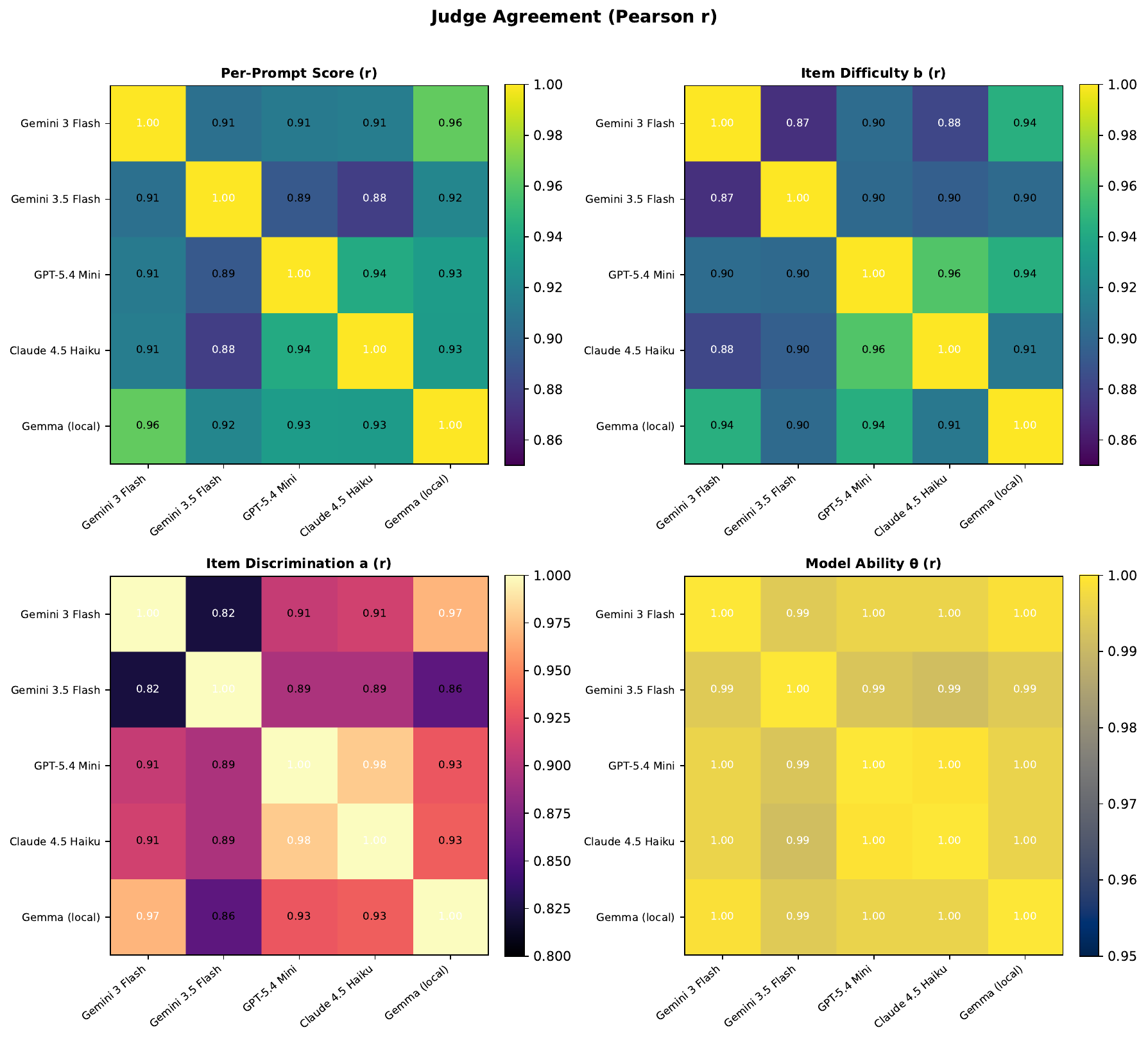}

    \vspace{0.6em}
    \includegraphics[width=0.72\textwidth]{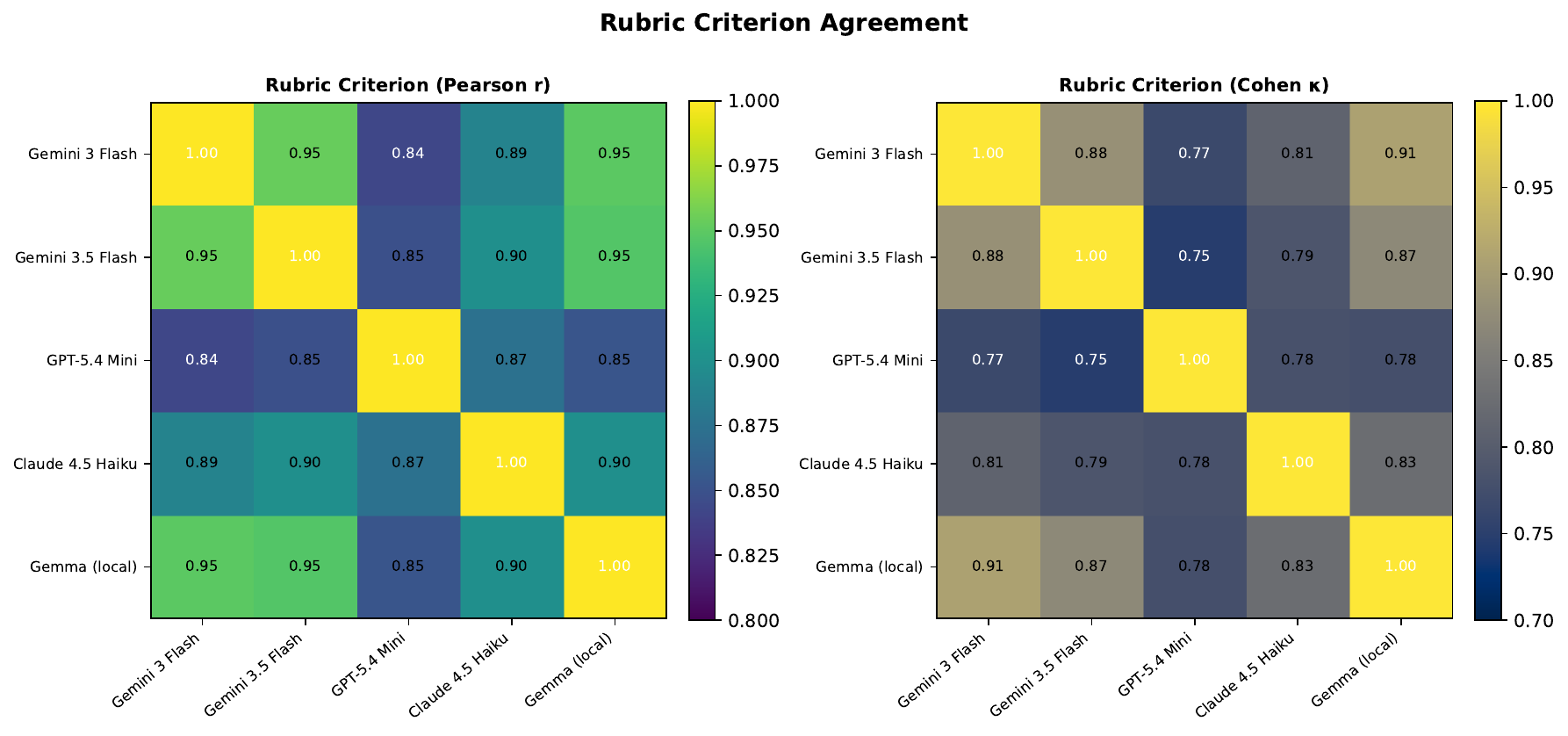}
    \caption{Pairwise agreement across the five judges. \textbf{Top:} Pearson $r$ for per-prompt scores, IRT difficulty $b$, discrimination $a$, and ability $\theta$ (from independent per-judge calibrations). \textbf{Bottom:} rubric-criterion agreement (Pearson $r$ and Cohen $\kappa$). Rankings are nearly judge-invariant; the finest signal (rubric criteria) spreads most, driven by GPT-5.4~Mini.}
    \label{fig:judge-agreement}
\end{figure}

A shared item bank with a simple additive judge term is only reasonable if the judges agree on the \emph{shape} of difficulty and on relative ability, differing mainly by an overall offset. They do. Averaged over the ten judge pairs (Figure~\ref{fig:judge-agreement}), per-prompt scores correlate at $r = 0.919$, item difficulty at $r = 0.911$, discrimination at $r = 0.910$, and rubric criteria at $r = 0.896$ (Cohen's $\kappa = 0.816$); model ability $\theta$ from independent per-judge calibrations correlates at Pearson $r = 0.996$ and Spearman $\rho = 0.996$. Table~\ref{tab:judge-leaderboard} shows that the per-judge ability rankings --- i.e.\ what a \emph{single-judge} model would produce --- move only modestly: the top eight configurations are stable (spread $\leq 3$) and the largest movement across all five judges is five places.

\begin{figure}[h]
    \centering
    \includegraphics[width=\textwidth]{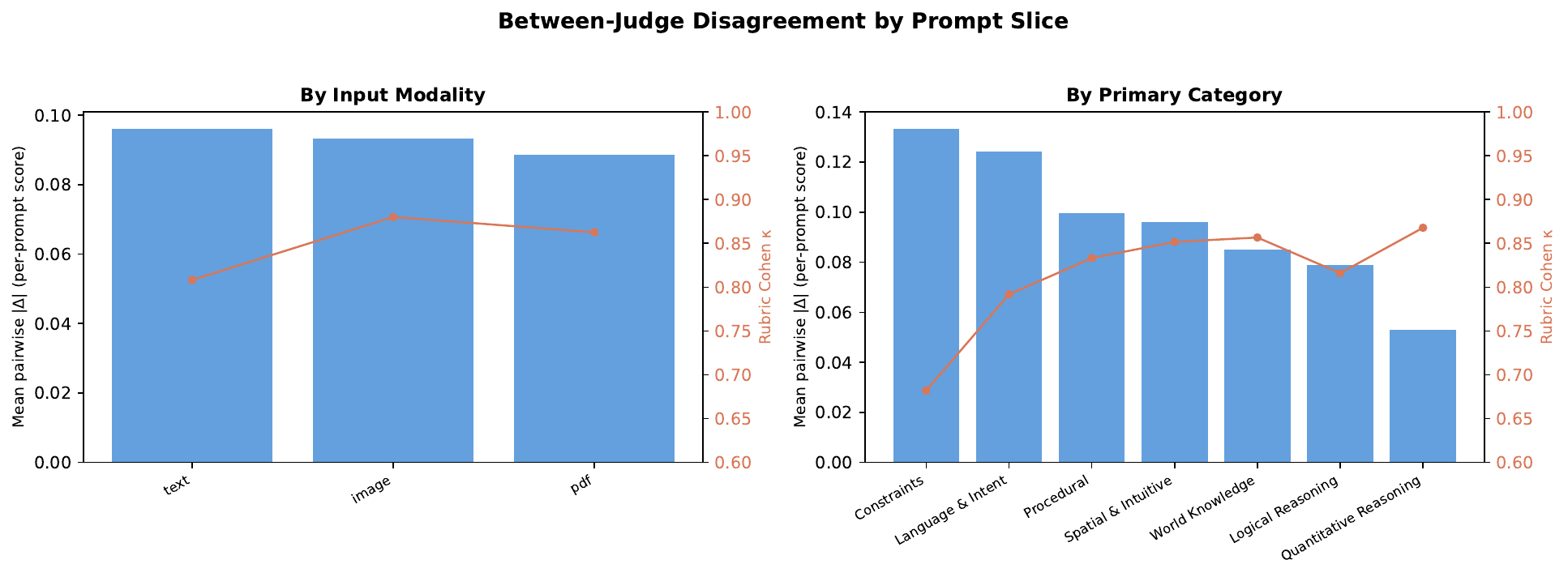}

    \vspace{0.6em}
    \includegraphics[width=0.72\textwidth]{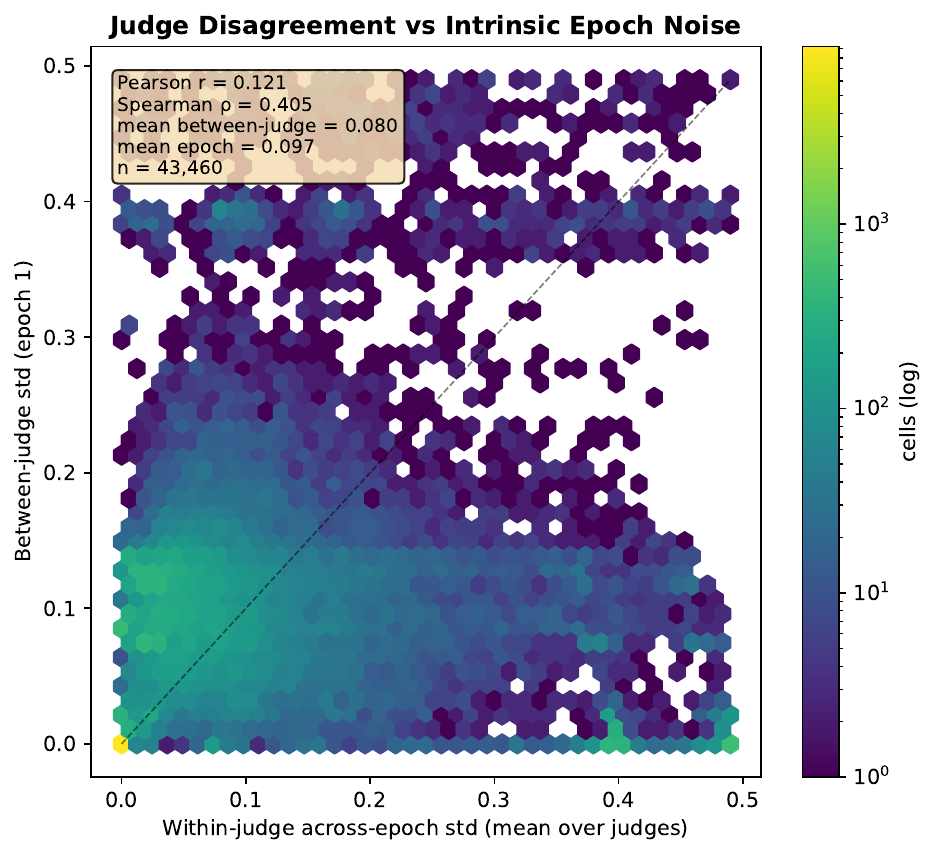}
    \caption{\textbf{Top:} between-judge disagreement by prompt slice --- mean pairwise $|\Delta|$ of per-prompt scores (bars) and rubric Cohen $\kappa$ (line); judges diverge most on subjective categories. \textbf{Bottom:} between-judge spread vs.\ within-judge across-epoch spread per cell; disagreement is smaller than resampling noise and concentrates on intrinsically unstable cells.}
    \label{fig:judge-epoch}
\end{figure}

Disagreement, where it occurs, is structured rather than arbitrary (Figure~\ref{fig:judge-epoch}, top): judges agree least on the subjective \textbf{Constraints} ($\kappa = 0.682$) and \textbf{Language~\&~Intent} categories and most on verifiable \textbf{Quantitative~Reasoning} ($\kappa = 0.868$). And it is small relative to the benchmark's own sampling noise: the between-judge spread of a (model, prompt) cell ($0.080$) is smaller than its within-judge spread across epochs ($0.097$), and the two are positively correlated (Spearman $\rho = 0.405$) --- judges diverge most exactly where re-sampling the model is itself unstable (Figure~\ref{fig:judge-epoch}, bottom).

\subsection{The Released Leaderboard}
\label{appendix:judge-multi-leaderboard}

Table~\ref{tab:judge-leaderboard} is the released joint judge-aware leaderboard --- ability $\theta$ on the shared judge-neutral scale onto which any one judge's single-run score lands (\S\ref{appendix:judge-multi-recovery}) --- shown alongside how each judge's \emph{independent} calibration ranks the same configurations. The human--LLM \emph{Top Human + AI} baseline leads and \emph{Average Human + AI} sits mid-pack. Both are scored across all five judges with the judge fixed-effect model (Appendix~\ref{appendix:centaur-details}), exactly like the autonomous models; for the per-judge columns each baseline is placed on a given judge's independent scale by scoring its participant responses against that judge's item bank (which reproduces the fitted abilities to within $0.001$). Every judge ranks \emph{Top Human + AI} in its top three and \emph{Average Human + AI} between 21st and 26th, matching their joint-model positions.

{\small
\begin{longtable}{rlrrrrrrr}
\caption{Released joint judge-aware leaderboard with single-judge agreement. \emph{Rank} is the overall leaderboard position by $\theta$. The five judge columns rank all 48 configurations under each judge's \emph{own} independent calibration (G3F = Gemini~3~Flash, G3.5F = Gemini~3.5~Flash, GPTm = GPT-5.4~Mini, C4.5H = Claude~4.5~Haiku, Gma = Gemma~4~31B); the two human baselines are placed on each judge's scale by scoring their participant responses against that judge's item bank. A judge's rank is \textbf{bold} when it deviates from the configuration's $\theta$-rank by $\geq 3$, and $\Delta$ is the rank spread (max $-$ min). Six non-reported configurations are omitted here but are included in the agreement statistics of \S\ref{appendix:judge-multi-agreement}.}
\label{tab:judge-leaderboard} \\
\toprule
Rank & Configuration & $\theta$ & G3F & G3.5F & GPTm & C4.5H & Gma & $\Delta$ \\
\midrule
\endfirsthead
\multicolumn{9}{l}{\emph{\tablename~\thetable\ (continued)}} \\
\toprule
Rank & Configuration & $\theta$ & G3F & G3.5F & GPTm & C4.5H & Gma & $\Delta$ \\
\midrule
\endhead
1 & Top Human + AI & $+1.728$ & 3 & 1 & 1 & 2 & 3 & 2 \\
2 & GPT 5.4 Pro / X-High & $+1.631$ & 1 & 2 & 2 & 1 & 2 & 1 \\
3 & GPT 5.4 Pro / High & $+1.619$ & 2 & 3 & 4 & 3 & 1 & 3 \\
4 & GPT 5.4 Pro / Medium & $+1.550$ & 4 & 4 & 3 & 4 & 4 & 1 \\
5 & Gemini 3.1 Pro / High & $+1.472$ & 5 & 5 & 5 & 5 & 5 & 0 \\
6 & Gemini 3.1 Pro / Medium & $+1.347$ & 6 & 6 & 6 & 6 & 7 & 1 \\
7 & Muse Spark / X-High & $+1.288$ & 7 & 8 & 7 & 7 & 6 & 2 \\
8 & GPT 5.4 / X-High & $+1.227$ & 8 & 7 & 9 & 9 & 9 & 2 \\
9 & GPT 5.4 / High & $+1.214$ & 9 & 9 & 8 & 8 & 10 & 2 \\
10 & Claude 4.7 Opus / High & $+1.088$ & 10 & 10 & \textbf{13} & \textbf{13} & 8 & 5 \\
11 & Claude 4.6 Opus / High & $+1.064$ & 11 & 13 & 12 & 10 & 11 & 3 \\
12 & GPT 5.4 / Medium & $+1.045$ & 13 & 11 & 10 & 11 & 12 & 3 \\
13 & Gemini 3.1 Pro / Low & $+1.028$ & 12 & 12 & 11 & 12 & 13 & 2 \\
14 & Muse Spark fp8 / High & $+1.021$ & 14 & 14 & 14 & 14 & 15 & 1 \\
15 & Claude 4.6 Sonnet / High & $+0.852$ & 15 & 15 & 15 & 15 & 14 & 1 \\
16 & Claude 4.6 Opus / Medium & $+0.759$ & 16 & 16 & 16 & 16 & 16 & 0 \\
17 & GPT 5.4 / Low & $+0.596$ & 19 & 17 & 17 & 17 & 19 & 2 \\
18 & Claude 4.6 Sonnet / Medium & $+0.595$ & 18 & 18 & 19 & 18 & 18 & 1 \\
19 & GPT 5 / High & $+0.578$ & 17 & 19 & 18 & 19 & 17 & 2 \\
20 & GPT 5 / Medium & $+0.360$ & 20 & 20 & 20 & 20 & 20 & 0 \\
21 & Claude 4.6 Sonnet / Low & $+0.204$ & 21 & 22 & 22 & 21 & 22 & 1 \\
22 & Gemma 4 31B bf16 / Medium & $+0.170$ & 23 & 21 & 23 & 24 & 21 & 3 \\
23 & Claude 4.6 Opus / Low & $+0.140$ & 22 & 24 & 24 & 23 & 23 & 2 \\
24 & Average Human + AI & $+0.133$ & 26 & 23 & \textbf{21} & 22 & 26 & 5 \\
25 & Gemma 4 31B fp4 / Medium & $+0.090$ & 25 & 25 & 25 & 26 & 24 & 2 \\
26 & Gemma 4 31B fp8 / Medium & $+0.090$ & 24 & 26 & 26 & 25 & 25 & 2 \\
27 & Gemini 2.5 Pro / High & $+0.012$ & 27 & 29 & 27 & 29 & 28 & 2 \\
28 & Gemini 2.5 Pro / Dynamic & $-0.006$ & 29 & 28 & 29 & 28 & 29 & 1 \\
29 & GPT 5.4 Mini / High & $-0.021$ & 30 & 27 & \textbf{32} & 27 & 31 & 5 \\
30 & Gemini 3.1 FL / High & $-0.027$ & 28 & 30 & 31 & 32 & \textbf{27} & 5 \\
31 & GPT 5 / Low & $-0.043$ & 31 & 31 & \textbf{28} & 30 & 30 & 3 \\
32 & Gemini 2.5 Pro / Medium & $-0.071$ & 32 & 33 & 30 & 31 & 32 & 3 \\
33 & GPT 5.4 Mini / Medium & $-0.127$ & 33 & 32 & 34 & 33 & 33 & 2 \\
34 & Gemini 2.5 Pro / Low & $-0.161$ & 34 & 34 & 33 & 34 & 34 & 1 \\
35 & GPT OSS 120B / Medium & $-0.555$ & 35 & 35 & 35 & 35 & 37 & 2 \\
36 & Gemini 3.1 FL / Medium & $-0.627$ & 37 & 37 & 36 & 37 & 35 & 2 \\
37 & Claude 4 Opus / High & $-0.628$ & 36 & 36 & 37 & 36 & 36 & 1 \\
38 & Claude 4.5 Haiku / High & $-0.709$ & 38 & 38 & 38 & 38 & 38 & 0 \\
39 & GPT 5.4 Mini / Low & $-0.761$ & 39 & 39 & 39 & 40 & 39 & 1 \\
40 & Claude 4.5 Haiku / Medium & $-0.806$ & 41 & 40 & 40 & 41 & 41 & 1 \\
41 & Claude 4 Opus / Medium & $-0.835$ & 40 & 43 & 42 & 39 & 40 & 4 \\
42 & Gemma 4 31B Std / OFF & $-0.849$ & 43 & 41 & 41 & 44 & 42 & 3 \\
43 & Gemini 3.1 FL / Low & $-0.909$ & 45 & 44 & 43 & 45 & 44 & 2 \\
44 & Claude 4.5 Haiku / Low & $-0.912$ & 44 & 42 & 45 & 42 & 45 & 3 \\
45 & Claude 4 Opus / Low & $-0.922$ & \textbf{42} & 45 & 44 & 43 & 43 & 3 \\
46 & Llama 4 Maverick / OFF & $-1.478$ & 46 & 46 & 46 & 46 & 46 & 0 \\
47 & GPT-4o / OFF & $-1.528$ & 47 & 47 & 47 & 47 & 47 & 0 \\
48 & Llama 3.3 70B / OFF & $-1.653$ & 48 & 48 & 48 & 48 & 48 & 0 \\
\bottomrule
\end{longtable}
}

\subsection{Takeaways}
\label{appendix:judge-multi-takeaways}

Adding a judge fixed effect to the IRT model achieves its design goal. The five judges agree closely on difficulty shape and on relative ability (Pearson $r \geq 0.91$; rank Spearman $\rho = 0.996$), so they differ essentially by a single per-judge offset --- precisely what $\gamma_k$ models. Releasing the shared item bank with those offsets lets any one of the five calibrated judges place a new model on the common $\theta$ scale from a \emph{single} run, recovering the full-panel ability at $\approx 0.13$ logit MAE and at most three rank places (Table~\ref{tab:judge-recovery}). The judge term is necessary: without it a single judge's ability is biased by its leniency, up to a $1.10$-logit gap between judges. A single fixed judge would lock users to one proprietary model and impose that gap; a five-judge panel is essentially exact but costs five runs to add one model. The judge-fixed-effect design keeps the panel's judge-neutral scale while requiring only one judge of the user's choice.

\section{Centaur Study Protocol}
\label{appendix:centaur-details}

This appendix details the human--LLM (``centaur'') evaluation protocol summarized in Section~\ref{subsection:centaur-evaluation}. The protocol is specified at two levels: the recruitment, batching, and quality requirements imposed on the staffing partner that ran the study, and the participant-facing instructions distributed to each participant at the start of their session.

\subsection{Staffing, Vetting, Compensation, and Consent}
\label{appendix:centaur-staffing}

The study was conducted under contract with \textbf{Labelbox} (\url{https://labelbox.com}), a third-party data-labeling vendor that handled all participant-facing logistics. Labelbox was responsible for (i) recruiting participants matching the profile specifications in Appendix~\ref{appendix:centaur-profiles}, (ii) vetting participant credentials and submitting each profile to the GIM evaluation team for approval before any work was assigned, (iii) compensating participants for their time, and (iv) collecting informed consent from each participant authorizing the use of their responses as part of the centaur study reported in this paper. The GIM evaluation team specified the recruitment criteria, batch structure, and execution constraints described in the subsections below, and Labelbox enforced these requirements and managed all direct contact with participants. Scoring of completed responses was performed independently by the GIM team using the same hybrid LLM-as-judge pipeline as for autonomous models (Appendix~\ref{appendix:centaur-scoring}).

\subsection{Participant Profiles and Recruitment}
\label{appendix:centaur-profiles}

Participants were drawn from three pre-specified profiles, two with a Science, Technology, Engineering, and Mathematics (STEM) background and one without:
\begin{itemize}
    \item \textbf{STEM Undergraduate} --- currently enrolled as an undergraduate student in a STEM field, with a foundational understanding of general STEM principles.
    \item \textbf{STEM PhD} --- currently enrolled in or graduated from a PhD program in a STEM field, with deep, specialized knowledge in a specific STEM discipline.
    \item \textbf{Non-STEM Undergraduate} --- currently enrolled as an undergraduate student in a non-STEM field, contributing general knowledge and critical-thinking skills applicable to a variety of subjects.
\end{itemize}
Profiles were submitted to the GIM evaluation team for review and approval before any participant was permitted to begin answering prompts.

The 820 prompts were divided into 41 batches of 20 prompts each ($820 / 20 = 41$). We recruited \textbf{246 unique participants}---two from each of the three profiles per batch---yielding six independent runs per prompt (two from each profile). Every participant worked on exactly one batch of 20 prompts; Labelbox handled the batch-to-participant assignment and presented the prompts to each participant directly through their platform.

\subsection{Execution Constraints}
\label{appendix:centaur-execution}

\begin{itemize}
    \item \textbf{Time budget.} Each participant had a strict total of \textbf{5 hours} to complete their assigned batch of 20 prompts. The 5 hours covered all phases of the work: research, LLM interaction, writing the final answer, and submitting the completed batch.
    \item \textbf{LLM access.} All participants were guaranteed consistent and reliable access to \textbf{Gemini 3.1 Pro}. Participants were free to additionally use any other LLMs, Google Search, or reference materials available to them.
    \item \textbf{Resource reporting.} Participants were asked to record the categories of resources they used in broad strokes (e.g., ``Google Search'', ``Books'', ``ArXiv'', ``Scientific papers'', ``Other LLMs''), aggregated at the participant level rather than per-question. Detailed lists of individual websites or papers were not required.
    \item \textbf{Verification expectation.} Participants were asked to verify model output and were encouraged to use the full time budget to research, validate, and edit answers before submission. Copying and pasting an LLM answer was an acceptable way to complete a question---particularly when time was short---but participants were asked to treat it as a fallback rather than the default.
\end{itemize}

\subsection{Participant-Facing Instructions}
\label{appendix:centaur-instructions}

The instructions below were distributed to each participant at the start of their 5-hour session.

\textbf{Task overview:} You will answer 20 questions using LLM assistance. Your task is to answer all questions as accurately as possible.

\textbf{Time limit:} You will have 5 hours total to complete your 20 questions. This includes researching, using LLMs, writing your final answers, and submitting your work.

\textbf{Tools you may use:} You may use Gemini 3.1 Pro, other LLMs, Google Search, and any other reference materials available to you.

\textbf{Reporting the resources you used:} If you use materials other than Gemini 3.1 Pro, please list them in broad categories. You do not need to provide a detailed log. Examples include ``Google Search'', ``Books'', ``ArXiv'', ``Scientific papers'', and ``Other LLMs''. You do not need to list every website you visited, every paper you read, or which resource was used for each individual question.

\textbf{How to complete the task:} Work through all 20 questions in your batch. Use the allowed tools and references as needed. Enter your final answers in the designated area. Record the resources you used in broad terms, and submit your completed work within the 5-hour time limit.

\textbf{Quality expectations:} Please answer as accurately as possible. The study is designed to evaluate the quality of your final answers when working with LLM assistance. We ask that you verify and, where appropriate, edit any model-generated content before submitting it as your answer. Copying and pasting an LLM answer is an acceptable way to complete a question---especially if you are running short on time---but please treat it as a fallback rather than your default approach.

\subsection{Post-Hoc Filtering, Scoring, and Aggregation}
\label{appendix:centaur-scoring}

Two patterns of off-task responses were observed during scoring. A small number of participants misunderstood the assignment and provided feedback on the LLM or on the question itself rather than answering it. A second group submitted only minimal content (e.g., single-word answers) or did not record answers for the prompts in their batch. For the purposes of scoring the centaur configurations, we excluded these participants from the analysis, leaving \textbf{195 retained participants} whose responses were used for scoring. All recruited participants were compensated by the staffing partner for their time in accordance with the contract.

The retained answers were scored by the same hybrid LLM-as-judge pipeline used for autonomous-model evaluation (Section~\ref{subsection:hybrid-scoring}): rubric-graded prompts were decomposed into independently judged criteria with confidence-weighted partial credit, and exact-answer prompts were graded against the reference.

We aggregated the scored responses into two centaur test-configurations:
\begin{itemize}
    \item \textbf{Average Human + AI.} Combines the responses of all 195 retained participants.
    \item \textbf{Top Human + AI.} Restricted to the 15 participants with the highest average raw score.
\end{itemize}
Each group's responses were graded by all five calibrated judges and scored against the item bank $\{a_j, b_j\}$ with the judge fixed-effect model (Section~\ref{subsection:irt-model}): the closed-form weighted-least-squares scorer subtracts each judge's leniency offset $\gamma_k$ and pools the five judges' graded scores, yielding a single $\theta$ for each centaur configuration on the same judge-neutral scale as the autonomous-model leaderboard. The participant set is unchanged from the single-judge analysis (the same 195 retained participants and the same top-15 subset); only the grading is extended from one judge to the full panel.

\section{Compute Footprint}
\label{appendix:compute-footprint}

We report compute usage as a transparency artifact. Inference tokens are computed precisely on the GPT and Gemini configurations---where input, reasoning, and final-output tokens are all exposed by the provider API. Judge tokens are estimated by measuring per-call token rates on a sister evaluation run that used a more verbose judge as a proxy and adjusting for the lower per-call cost expected from the calibrated judge panel; all five judges share the same inputs (rubric text plus the model's response) per call, but their output lengths and reasoning-token use differ by model. All reported numbers exclude tokens spent on retried or failed API calls and are therefore lower bounds.

Evaluating one (model, thinking-level) configuration on the full 820-prompt benchmark at 5 epochs costs on average $\sim$56M inference tokens plus judge-scoring tokens. The legacy one-judge estimate is $\sim$20M judge tokens, or $\sim$76M total tokens per configuration; five-judge panel scoring reuses model responses but adds judge calls. Extrapolating to the 47 reporting configurations, the full leaderboard consumed roughly $10^{21}$ inference FLOPs. At a representative frontier active-parameter count of $\sim$$10^{11}$ and the standard $2N$ FLOPs-per-token approximation \citep{kaplan2020scaling, hoffmann2022chinchilla}, judge-token cost scales with the selected calibrated judge panel.

These figures separate model-response generation from judge scoring because GIM supports one or more calibrated judges from the five-judge panel. Additional token budget was spent on trial-and-error setup---identifying correct per-provider inference settings, debugging the harness, discarded exploratory runs, and an initial judging pass with a more expensive judge configuration---and is not included here.

\section{Detailed Results and Bank Diagnostics}
\label{appendix:results-detail}

This appendix collects the per-figure detail and discussion that supports the headline numbers in Sections~\ref{section:results} and~\ref{section:benchmark-characteristics}. The first part expands the IRT scoring affordances, the per-category and per-label heatmaps, and the finer thinking-gain breakdowns. The second part collects the bank-side diagnostics summarized in Section~\ref{section:benchmark-characteristics}.

\subsection{Why an IRT Layer? Five Affordances over a Raw Mean}
\label{appendix:irt-affordances}

As Section~\ref{subsection:irt} notes, our $\theta$ is near-monotone in the rubric-weighted raw mean ($r \approx 0.986$, Section~\ref{subsection:saturation}); the value of the IRT layer is therefore not a different ranking but five properties that a raw mean cannot deliver:
\begin{enumerate}
    \item \textbf{Cross-slice comparable scores with information-weighted standard errors.} A raw mean is also closed-form, but means on different subsets of the bank are not comparable to each other (a $0.7$ on an easy slice is not the same ability as a $0.7$ on a hard slice), and its $s/\sqrt{n}$ SE treats every item as equally informative. The IRT scorer puts every slice on the same $\theta$ scale by absorbing item difficulty $b_j$, and its analytic SE weights each item by its discrimination $a_j$ via the Fisher information, so high-$a_j$ items dominate and noisy items contribute little. We exploit this throughout the paper for category-conditional scores (Section~\ref{subsection:categories}), per-epoch reliability (Section~\ref{subsection:variance}), and the public/private contamination diagnostic (Section~\ref{subsection:contamination-diagnostic}).
    \item \textbf{Missing-data robustness.} Different (model, thinking-level) configurations cover slightly different subsets of the bank (e.g.\ image-disabled models skip image prompts; partially-run configurations skip late prompts). The IRT scorer down-weights or omits unobserved cells coherently, whereas a raw mean would silently change its support.
    \item \textbf{Item-level diagnostics.} The fitted item difficulty $b_j$ and discrimination $a_j$ become first-class objects, used in Section~\ref{section:benchmark-characteristics} to characterize the bank, in Section~\ref{subsection:categories} to compare cognitive dimensions, and as an ongoing tool for retiring saturated items.
    \item \textbf{A calibrated logit scale.} The transform stretches differences at the frontier (where high-$a_j$ items concentrate) and compresses differences in the middle, which is precisely the property we want from an ability scale on a benchmark designed to remain unsaturated.
    \item \textbf{Robustness to inference-failure noise.} Higher-end models at higher thinking budgets fail more often outright (longer chains of thought $\to$ more timeouts and truncated responses), and these zero/missing cells can drag a naive raw mean below a less-thinking sibling whose surviving answers are uniformly better. For example, GPT 5.4 fails more often at X-High than at High; this is enough to nudge its naive raw mean \emph{down} from $0.620$ at High to $0.616$ at X-High, even though $\theta$ correctly orders X-High above High ($1.23$ vs.\ $1.21$) and manual inspection confirms the X-High responses are at least as good wherever both succeed. The same IRT scoring separates the Gemma 4 quantization family on $\theta$ with margins that remain visible despite very similar raw means. We characterize the per-configuration failure rates that motivate this in Section~\ref{subsection:inference-failures}.
\end{enumerate}

\subsection{Per-Category and Per-Label Ability Breakdowns}
\label{appendix:category-breakdown}

Section~\ref{subsection:categories} summarized the per-category and per-label heatmaps as showing broad row-wise stability rather than sharp specialization. This appendix reproduces the underlying figures and discussion. Before reading the heatmaps, one interpretive caveat is worth flagging. Neither the seven primary categories nor the free-text labels are mutually exclusive and collectively exhaustive (MECE) in the dimensions of reasoning they tax. Most GIM prompts recruit several categories at once, and the assigned primary code merely identifies the most pronounced dimension rather than the only one: a ``quantitative reasoning'' prompt typically also requires logical reasoning, planning, and constraint-tracking; a ``spatial \& intuitive'' prompt often layers in world knowledge and language understanding. The free-text labels share this property even more strongly. They are non-exclusive by design, applied freehand by human annotators during dataset construction, and a single prompt routinely carries several labels (e.g.\ \emph{planning} + \emph{spatial} + \emph{multi-step}) without any of them being a clean isolation of the underlying skill. As a consequence, the per-category and per-label $\theta$ slices below should be read as differential lenses on the same underlying ability rather than as scores on cleanly separable sub-skills.

\subsubsection{Model ability by primary cognitive category}
\label{appendix:ability-by-category}
\begin{figure}[htbp]
\centering
\includegraphics[width=\textwidth,height=0.9\textheight,keepaspectratio]{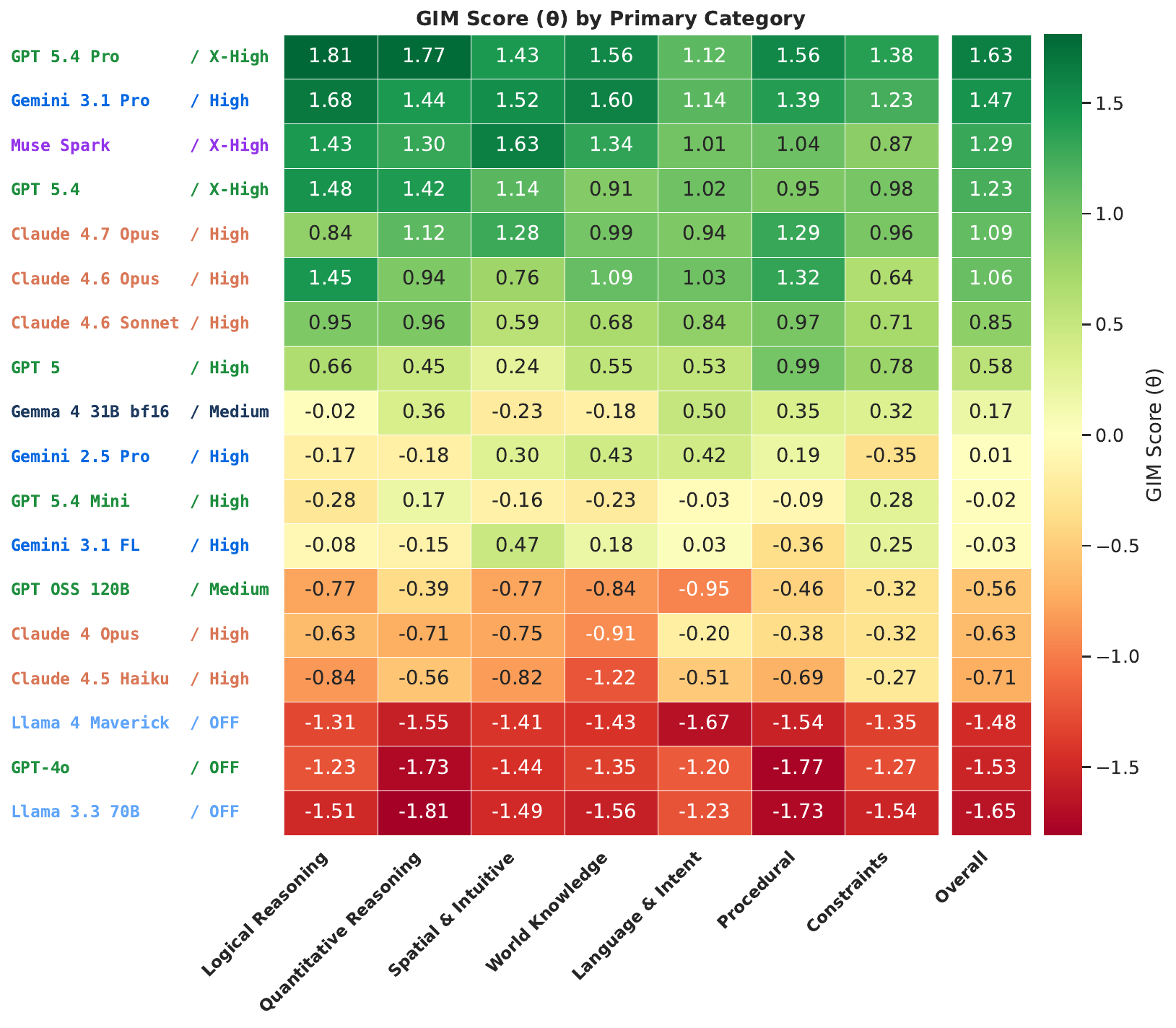}
\caption{Per-model GIM score $\theta$ by primary cognitive category (right block) alongside overall $\theta$ (left column). One row per model, deduped to the best variant within each base name and ordered by the canonical leaderboard. Color encodes $\theta$ on a divergent scale centered at zero.}
\label{fig:ability-by-category}
\end{figure}

Aggregating per-model scoring into the seven editorial primary categories gives a coarser, partition-style rollup. Figure~\ref{fig:ability-by-category} shows, for every model (deduped to the strongest variant within each base name), $\theta$ on each of the seven categories alongside its overall $\theta$. The dominant impression is broad stability rather than sharp per-category specialization: rows that sit high on Overall sit high across most columns, and individual cells rarely depart far from a model's Overall $\theta$. Constraints (CT) is the most visible across-the-board exception, sitting easier for essentially every model, but we are reluctant to read this as a clean ``constraint-following is solved'' signal. Because labeling assigns a single primary code to each multi-faceted prompt, harder prompts that mix constraint-following with quantitative or logical demands tend to be coded under their dominant non-CT axis; a CT slice enriched in the easier prompts is at least as plausible an explanation as a substantive ability gap on constraints per se. On the open frontier (QR/WK/SI/LR) some residual family-level variation is visible---e.g.\ GPT 5.4 Pro is comparatively uniform across the four columns, while Muse Spark and Claude 4.7 Opus tilt one way or the other on individual cells---but these contrasts are small relative to the overall-row spread, and we read them as directional indications rather than as cleanly separable strengths.

\subsubsection{Model ability by free-text label}
\label{appendix:ability-by-label}
\begin{figure}[htbp]
\centering
\includegraphics[width=\textwidth,keepaspectratio]{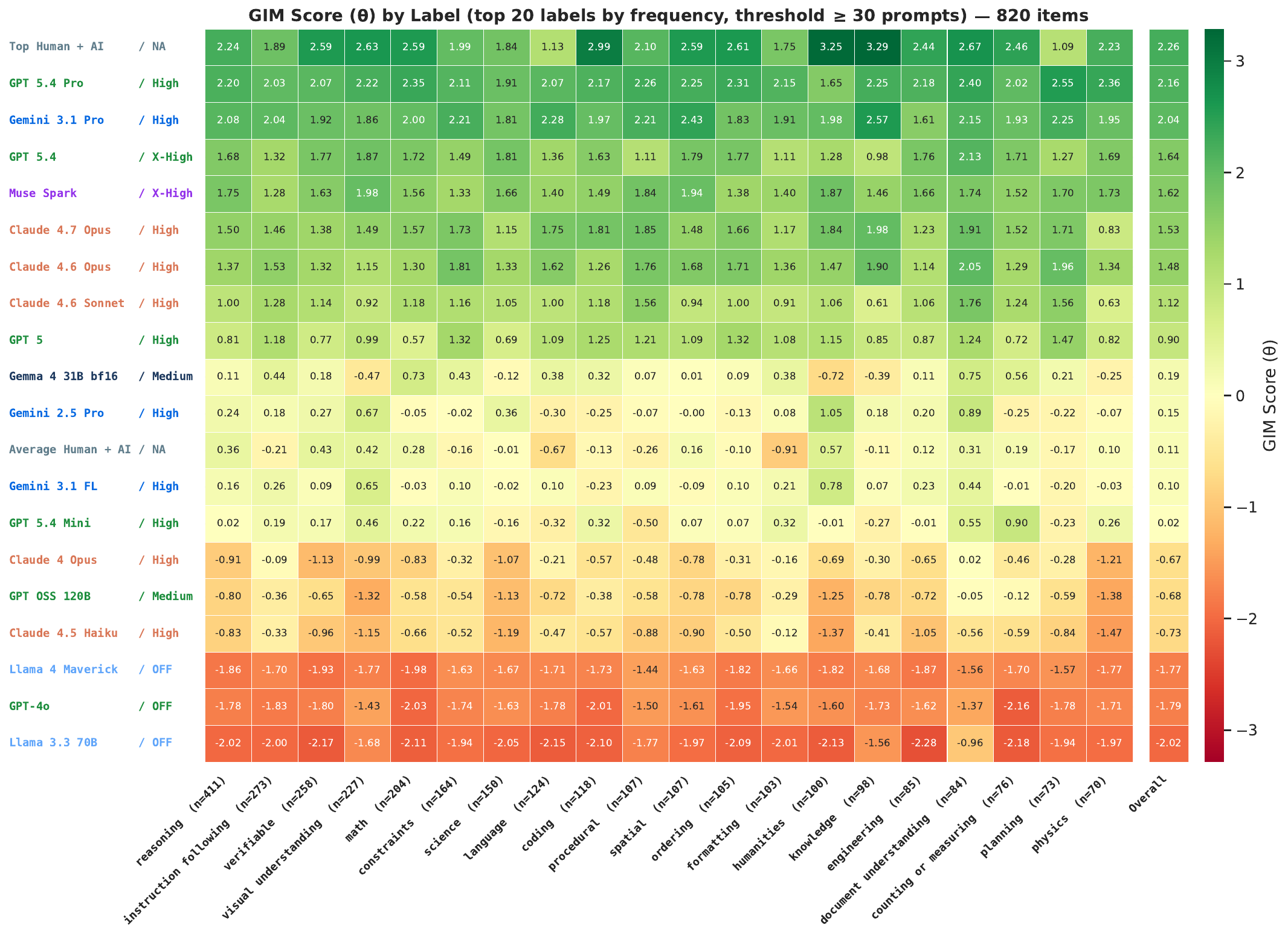}
\caption{Per-model GIM score $\theta$ by free-text label (right block) alongside overall $\theta$ (left column). One row per model, deduped to the best variant within each base name and ordered by the canonical leaderboard; columns are the most frequent free-text labels, ordered by frequency. Color encodes $\theta$ on a divergent scale centered at zero. Labels are non-exclusive, so the columns do not partition the bank.}
\label{fig:ability-by-label}
\end{figure}

Beyond the editorial primary categories, each GIM prompt carries a set of finer-grained free-text labels (e.g.\ \emph{spatial}, \emph{planning}, \emph{temporal reasoning}, \emph{web search}, \emph{chemistry}). Scoring each (model, label) cell with the same closed-form WLS estimator gives a finer-grained view of where each model tends to be relatively stronger or weaker. Figure~\ref{fig:ability-by-label} shows the resulting heatmap, restricted to the most frequent labels, with one row per deduped model and one column per label, ordered by canonical leaderboard rank and label frequency respectively. Within the open-frontier reasoning labels, peaks roughly track use-case-relevant axes: Muse Spark sits high on the \emph{spatial} and \emph{intuitive} labels; GPT 5.4 leads on \emph{puzzles}, \emph{temporal reasoning}, and \emph{lateral thinking}; and several knowledge-flavored labels (\emph{web search}, \emph{geography}, \emph{biology}, \emph{chemistry}) compress the inter-model spread substantially relative to the reasoning-flavored labels. Because labels are non-exclusive and overlap heavily, these label-level peaks should be read as directional rather than as clean isolations of any single skill.

\subsubsection{Top-model category radar}
\label{appendix:radar-top-models}
\begin{figure}[htbp]
\centering
\includegraphics[width=0.7\textwidth,keepaspectratio]{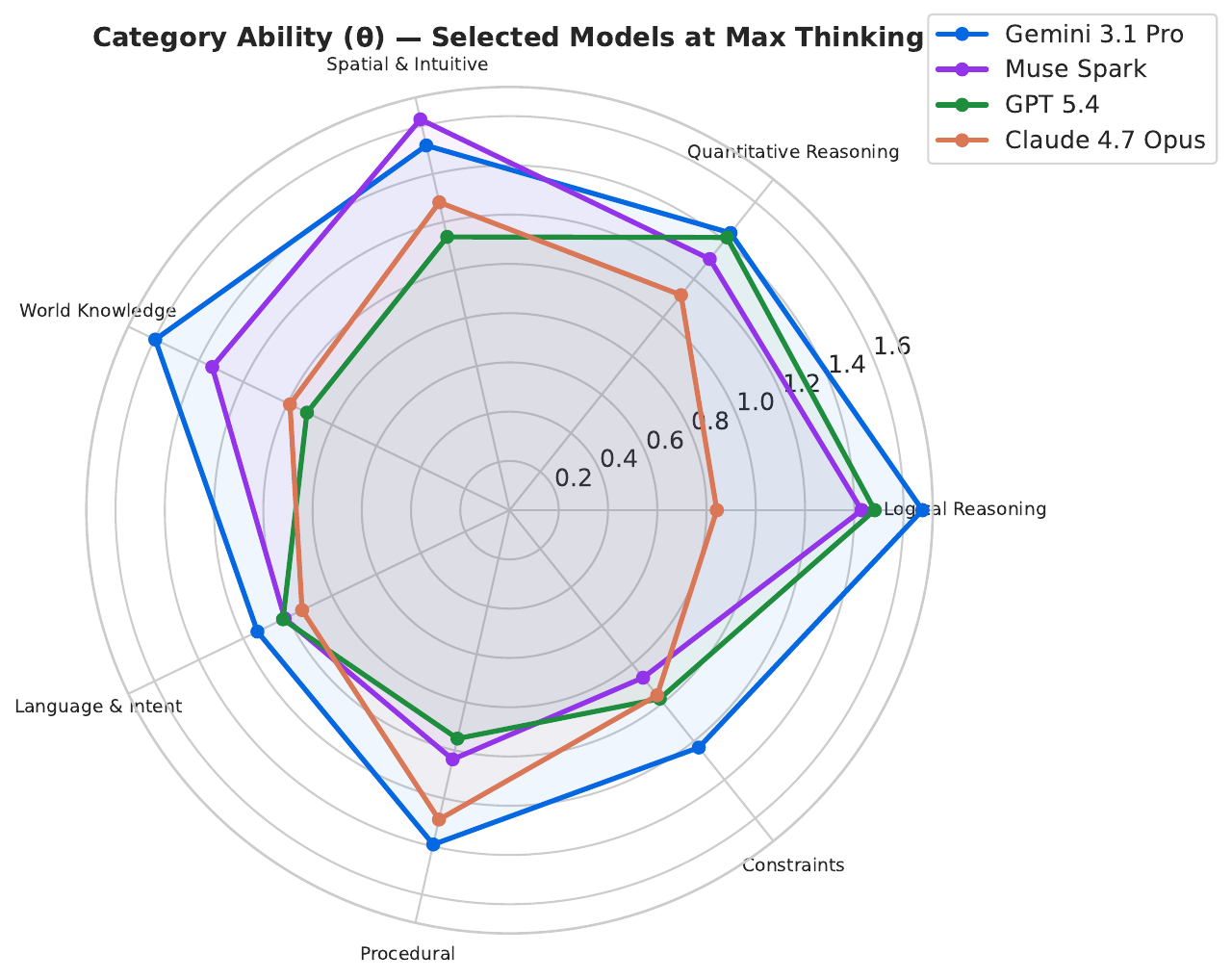}
\caption{Category $\theta$ for the four top configurations at their highest available thinking level. Models with similar overall $\theta$ trace visibly different shapes across the seven categories.}
\label{fig:radar-top-models}
\end{figure}

Figure~\ref{fig:radar-top-models} overlays the four leading configurations---GPT 5.4 (X-High), Gemini 3.1 Pro (High), Muse Spark (X-High), and Claude 4.7 Opus (High)---on a category radar. Despite very similar overall $\theta$ in the $1.6$--$2.1$ band, the four polygons trace somewhat different shapes: Muse Spark sits highest on Spatial \& Intuitive (image-driven), Claude 4.7 Opus is comparatively lower on the same axis, GPT 5.4 leads on Quantitative Reasoning, and Gemini 3.1 Pro is the most uniform of the four. The leaderboard ranking compresses these shape differences into a single number; given the category overlap noted above, the radar is best read as a directional aid for matching a model to a use case rather than as a clean per-skill profile.

\subsection{Item Difficulty Breakdowns}
\label{appendix:difficulty-breakdown}

Turning the same $b_j$ machinery on the items themselves, Section~\ref{subsection:categories} reported that reasoning-heavy slices cluster on the hard end and instruction-following-flavored slices on the easy end, in agreement at both label and category resolutions. The supporting figures are reproduced here.

\subsubsection{Item difficulty by free-text label}
\label{appendix:difficulty-by-label}
\begin{figure}[htbp]
\centering
\includegraphics[width=\textwidth,keepaspectratio]{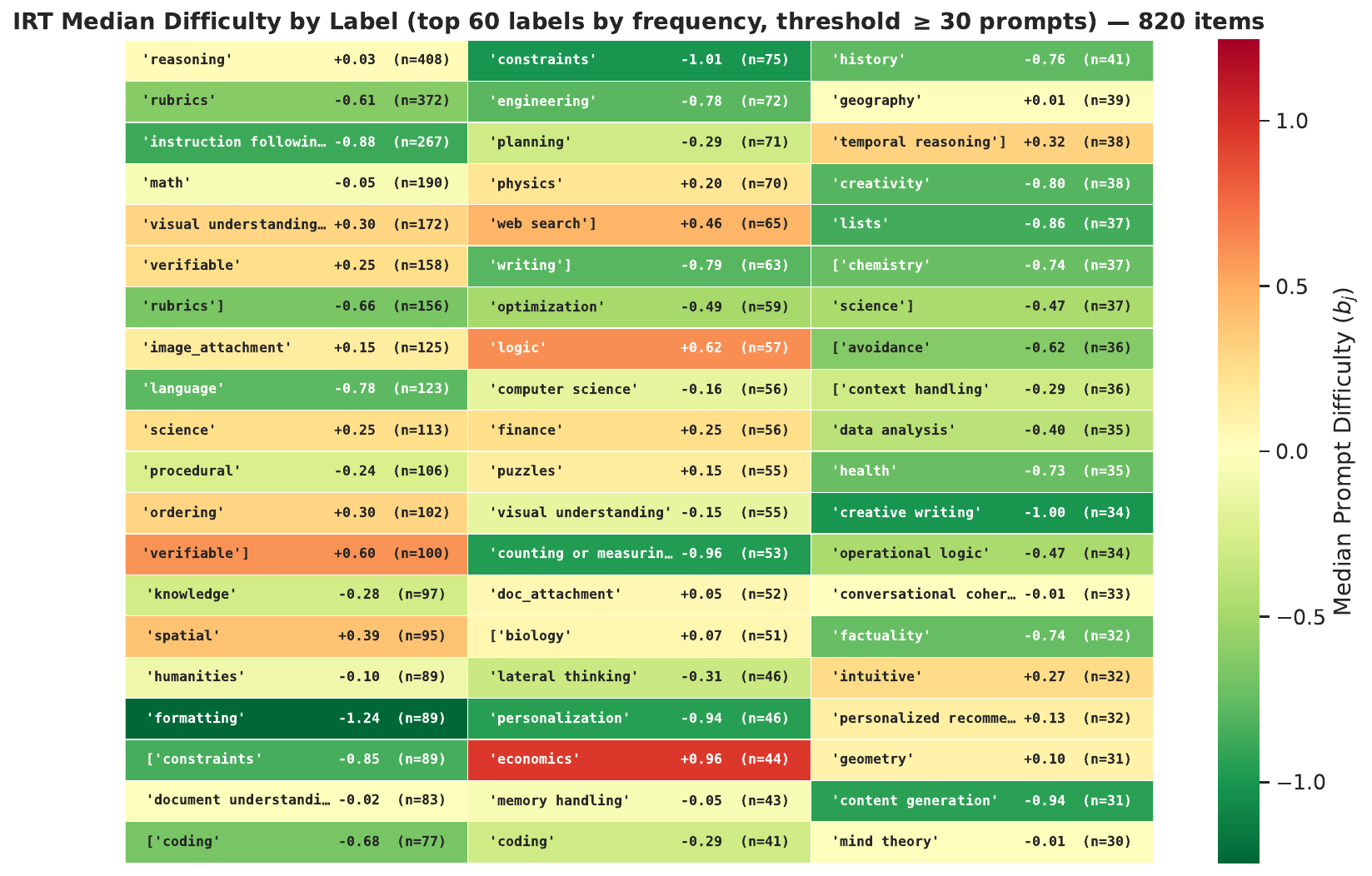}
\caption{Median fitted item difficulty $b_j$ by free-text label (only labels appearing on $\geq 30$ prompts). Labels are arranged hardest (top-left) to easiest (bottom-right). Reasoning-flavored labels dominate the hard end; instruction-following-flavored labels dominate the easy end.}
\label{fig:difficulty-by-label}
\end{figure}

Figure~\ref{fig:difficulty-by-label} ranks free-text labels with at least 30 prompts by median item difficulty $b_j$. The ordering separates cleanly by flavor: reasoning-heavy labels (planning, ordering, spatial, document understanding, multi-step) cluster on the hard end, while instruction-following-flavored labels (formatting, language, constraint adherence) cluster on the easy end. Because labels are non-exclusive and heavily overlapping---a single hard prompt can contribute to several reasoning-flavored cells at once---the absolute median values per label are correlated rather than independent, so the ranking is best read as a coarse hardness ordering of label flavors rather than as precise per-label difficulty estimates.

\subsubsection{Item difficulty by primary cognitive category}
\label{appendix:difficulty-by-category}
\begin{figure}[htbp]
\centering
\includegraphics[width=0.85\textwidth,keepaspectratio]{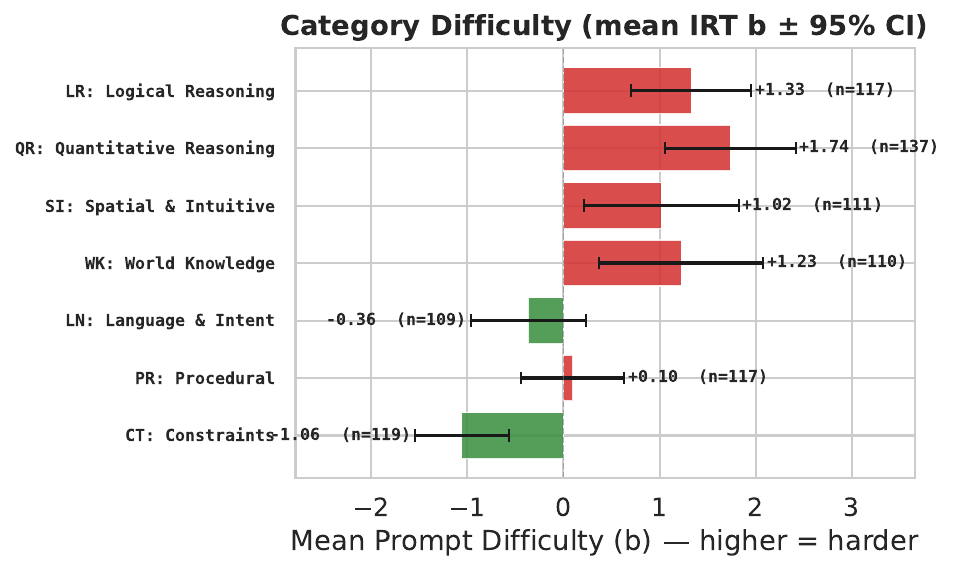}
\caption{Mean fitted item difficulty $b_j$ by primary cognitive category, with 95\% CI on the mean. Higher $b$ = harder. Quantitative reasoning sits at the hard end of the per-category ordering and constraint-following at the easy end; see text for the single-primary-code labeling caveat that limits a stronger reading.}
\label{fig:difficulty-by-category}
\end{figure}

Aggregating the same $b_j$ values into the seven editorial primary categories (Figure~\ref{fig:difficulty-by-category}) reproduces the label-level story along an entirely separate axis: Quantitative Reasoning (QR) items have the highest mean difficulty $b$, with World Knowledge (WK), Spatial \& Intuitive (SI), and Logical Reasoning (LR) close behind; Procedural (PR) and Language \& Intent (LN) cluster around zero; and Constraints (CT) is the clearest easy-side outlier. The same single-primary-code labeling caveat applies as in the model-side view above: harder multi-faceted prompts tend to be tagged under their dominant non-constraint axis even when constraint-tracking is materially involved, which would push CT toward the easier end and QR toward the harder end as a labeling artifact rather than as a substantive separation. That the label-level and category-level views agree, despite being constructed from completely different sources of structure in the dataset, is a useful cross-validation of the taxonomy.

\subsection{Detailed Test-Time Compute Breakdowns}
\label{appendix:thinking-detail}

Section~\ref{subsection:thinking} reported per-model and per-category thinking gains; this appendix gives the finer label-level and modality-level slices.

\subsubsection{By free-text label}
\label{appendix:thinking-by-label}
\begin{figure}[htbp]
\centering
\includegraphics[width=\textwidth,keepaspectratio]{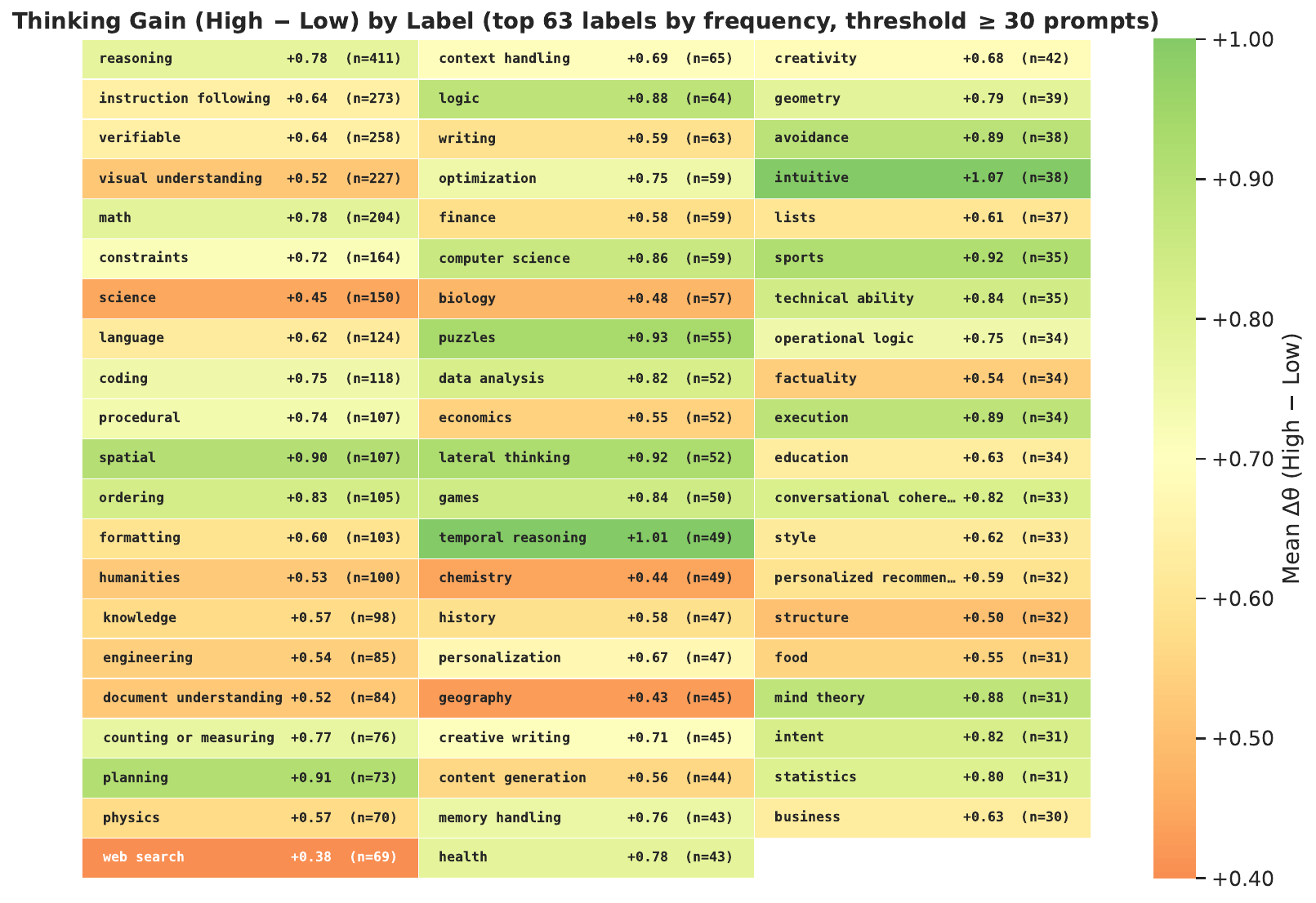}
\caption{Mean per-(model, label) $\Delta\theta$ (High~$-$~Low thinking, averaged across models) per free-text label. Labels are arranged most-frequent (top-left) to least-frequent (bottom-right). Reasoning-flavored labels gain the most from added thinking; knowledge- and retrieval-flavored labels gain the least. All qualifying labels show positive aggregate gain.}
\label{fig:thinking-gain-by-labels}
\end{figure}

By free-text label (Figure~\ref{fig:thinking-gain-by-labels}), we measure $\Delta\theta$ on per-(model, label) IRT slice scores from Low to High thinking, then average across models per label. Reasoning-flavored labels---\emph{intuitive} ($+0.80$), \emph{temporal reasoning} ($+0.81$), \emph{puzzles} ($+0.67$), \emph{lateral thinking} ($+0.72$), \emph{planning} ($+0.69$), \emph{spatial} ($+0.68$)---show the largest aggregate gains, while knowledge- and retrieval-flavored labels---\emph{web search} ($+0.37$), \emph{geography} ($+0.33$), \emph{chemistry} ($+0.35$), \emph{science} ($+0.35$), \emph{biology} ($+0.34$)---show the smallest. Every qualifying label has a positive Low-to-High aggregate gain, so on the GIM bank additional thinking is never aggregate-harmful at that resolution; its marginal value varies by roughly $2\times$ between the most reasoning-heavy and most retrieval-heavy slices.

\subsubsection{By input modality}
\label{appendix:thinking-by-modality}
\begin{figure}[htbp]
\centering
\includegraphics[width=0.7\linewidth,keepaspectratio]{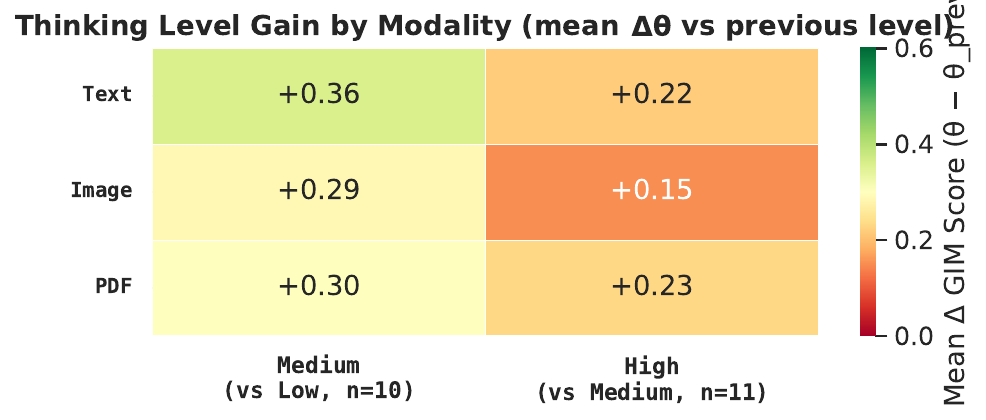}
\caption{Mean per-model marginal gain $\Delta\theta$ when stepping up one thinking level, broken down by input modality. Text inputs gain somewhat more than image and PDF inputs at both the Medium and High steps.}
\label{fig:thinking-gain-by-modality}
\end{figure}

By input modality (Figure~\ref{fig:thinking-gain-by-modality}), text inputs benefit somewhat more from added thinking than image or PDF inputs at the Medium step ($+0.36$ text vs.\ $+0.29$ image, $+0.30$ PDF). At the High step, PDF and text inputs are similar ($+0.23$ and $+0.22$) and image is lower ($+0.15$). The X-High step is small across modalities. The text advantage at the first step is consistent with the per-category and per-label views: text-heavy prompts skew toward Logical, Quantitative, and Procedural categories and toward the reasoning-flavored labels, which are exactly the dimensions that continue to reward thinking past the Medium step.

\subsection{Bank Diagnostics}
\label{appendix:bank-diagnostics}

This appendix reproduces the supporting figures and discussion for the diagnostics summarized in Section~\ref{section:benchmark-characteristics}.

\subsection{Item Bank Discrimination Structure}
\label{appendix:item-bank-detail}

\begin{figure}[htbp]
\centering
\includegraphics[width=\textwidth,keepaspectratio]{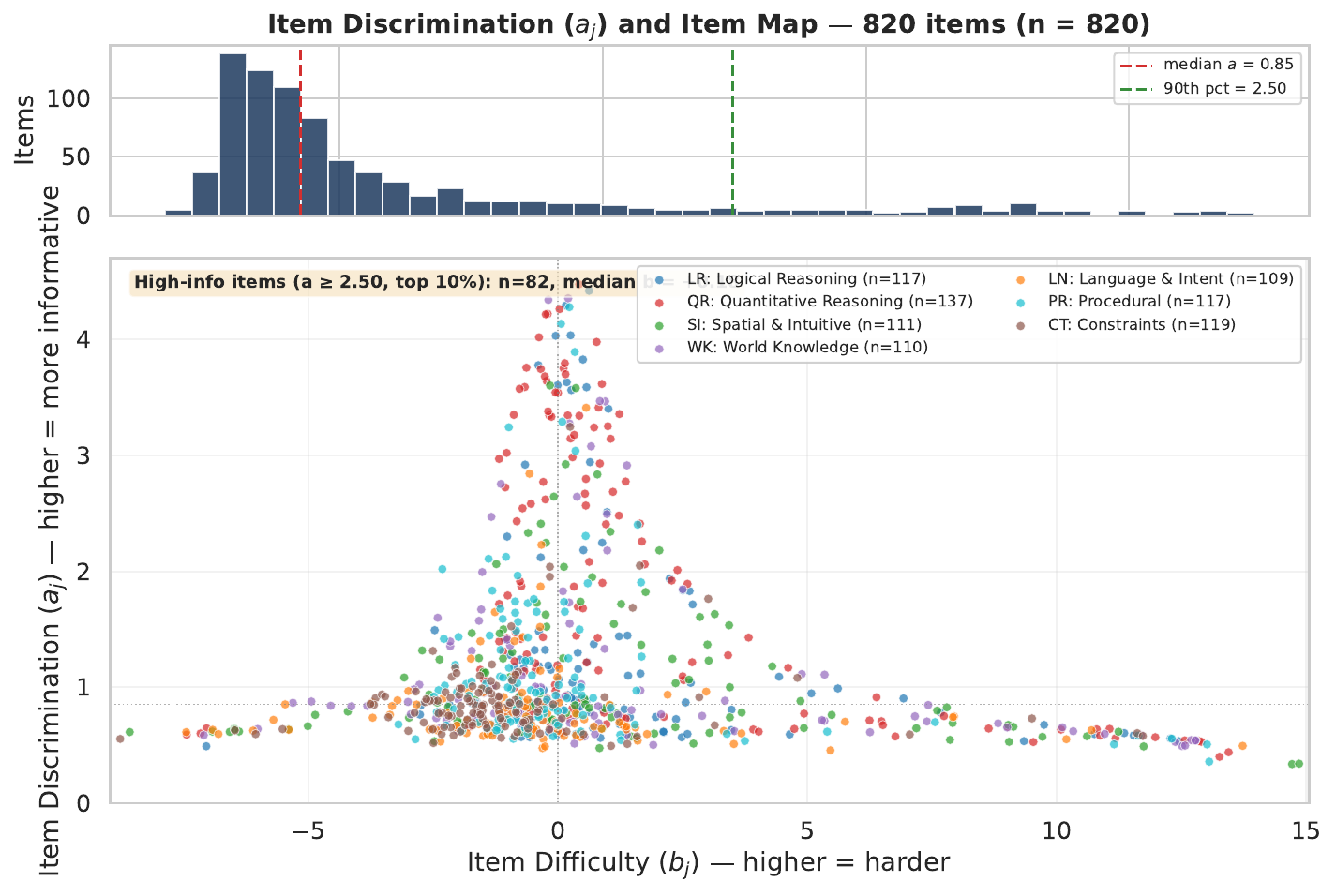}
\caption{Item discrimination structure of the calibrated bank. Top: marginal histogram of $a_j$, with median (red) and 90th-percentile (green) reference lines. Bottom: $(b_j, a_j)$ scatter colored by primary category. High-discrimination items concentrate near $b \approx 0$; items at extreme $b$ have low fitted $a$ both mechanically (bounded responses) and substantively (uniform success or failure across the model set).}
\label{fig:discrimination-difficulty}
\end{figure}

Figure~\ref{fig:discrimination-difficulty} shows the marginal histogram of item discrimination $a_j$ across the 820 items (top) and the full $(b_j, a_j)$ scatter colored by primary cognitive category (bottom). Two readings stand out. First, the $a_j$ marginal is right-skewed, with median $a$ near unity and a small tail of high-discrimination items (top $10\%$ at $a \geq 2.40$, $n=82$). The bank's information is not uniform across items---a minority of items carry disproportionate weight, which is exactly what an IRT-derived ability scale exploits. Second, the $(b, a)$ scatter has a characteristic ``tent'' shape: discrimination peaks near $b \approx 0$ and falls off symmetrically, with both the very hardest items ($b > 5$) and the very easiest ($b < -5$) settling at low $a$. This shape is partly mechanical---under a continuous-response 2PL with a bounded squeeze, items at extreme $b$ produce near-zero across-model variance and therefore admit no high-$a$ identification---but it is also informative: the bank's high-information items concentrate in $b \in [-3, +3]$, the difficulty range where current models are spread on both sides of the threshold. Items at $b > 5$ are an \emph{above-frontier reserve}: presently uniformly failed by every configuration, and so contributing little to today's leaderboard, but available to begin discriminating as future models cross into that range.

\subsection{Saturation Coverage and Raw-Mean Coherence}
\label{appendix:saturation-detail}

\begin{figure}[htbp]
\centering
\includegraphics[width=\textwidth,height=0.9\textheight,keepaspectratio]{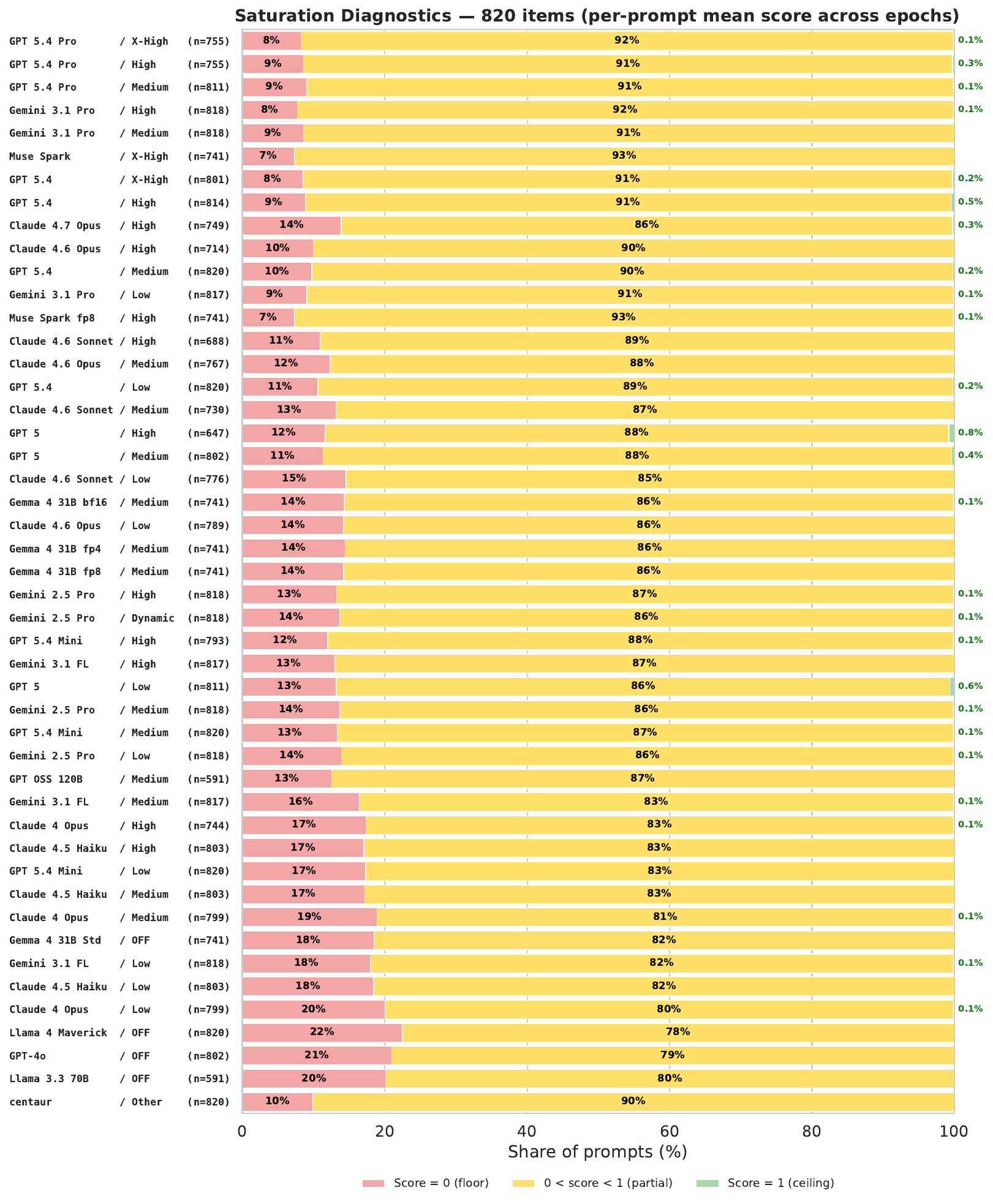}
\caption{Saturation diagnostics: per-(model, thinking-level) perfect-score rate (left, ceiling) and zero-score rate (right, floor), defined as the fraction of prompts whose epoch-mean score is exactly $1.0$ or $0.0$ respectively. Configurations are ordered by canonical leaderboard rank. Top configurations stay well short of a runaway ceiling; bottom configurations zero out a substantial slice, confirming the bench has neither saturated for the frontier nor lost signal at the floor.}
\label{fig:saturation-rates}
\end{figure}

Figure~\ref{fig:saturation-rates} reports, for every (model, thinking-level) configuration in canonical leaderboard order, the fraction of prompts whose epoch-mean score is exactly $1.0$ (a perfect-score rate, the ceiling diagnostic) and the fraction whose epoch-mean is exactly $0.0$ (a zero-score rate, the floor diagnostic). Even the strongest configurations sit well short of a runaway perfect-score rate---no model maxes out on more than a small fraction of the bank---confirming from the model side that GIM has not yet been solved by any frontier configuration. Symmetrically, the lowest-tier anchors zero out a substantial slice of the bank, exactly as required for the bank to retain signal at the bottom of the ability range.

\begin{figure}[htbp]
\centering
\includegraphics[width=0.85\textwidth,keepaspectratio]{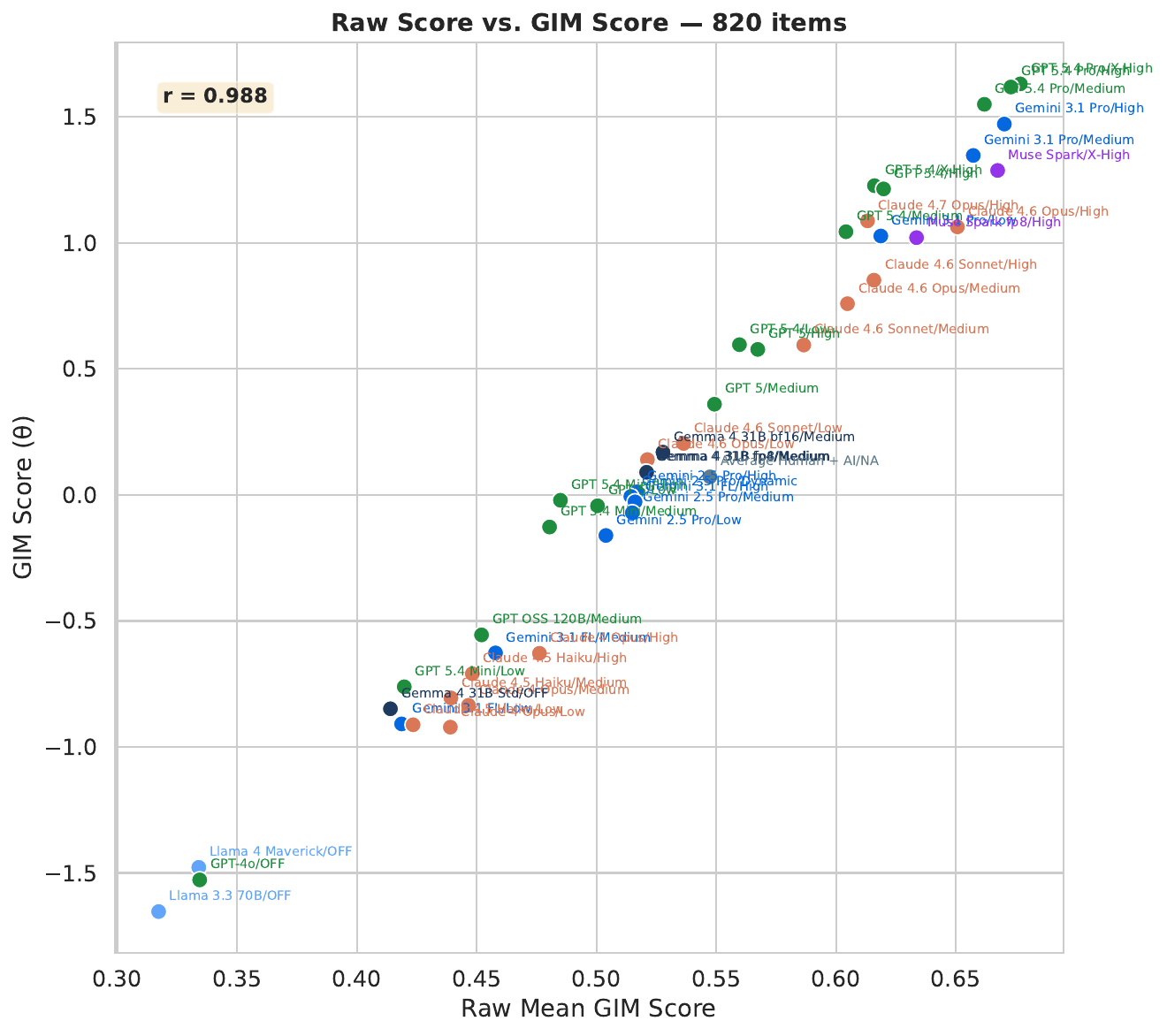}
\caption{Raw mean GIM score vs.\ IRT $\theta$ across all 47 reporting configurations. Points are colored by model family. Pearson $r \approx 0.986$: $\theta$ is a near-monotone reparameterization of the raw mean that stretches the upper tail of the leaderboard.}
\label{fig:raw-vs-theta}
\end{figure}

Figure~\ref{fig:raw-vs-theta} shows the raw-mean vs.\ $\theta$ scatter across all 47 reporting configurations. The Pearson correlation is $r \approx 0.986$. We read this as a sanity check rather than as evidence that IRT is redundant. A high rank correlation is exactly what we should see when models are evaluated on a (largely) common subset of items---the IRT layer is not in the business of inventing a different ranking, but of providing the affordances enumerated in Appendix~\ref{appendix:irt-affordances} (closed-form per-slice scoring, missing-data robustness, item-level diagnostics, a calibrated logit scale, and rectification of inference-failure noise). What the figure does show is the calibrated nonlinearity of the scale: $\theta$ stretches differences at the top of the leaderboard---where high-discrimination items are concentrated---and compresses them in the middle.

\subsection{Token Cost vs.\ Item Difficulty}
\label{appendix:tokens-detail}

\begin{figure}[htbp]
\centering
\includegraphics[width=\textwidth,keepaspectratio]{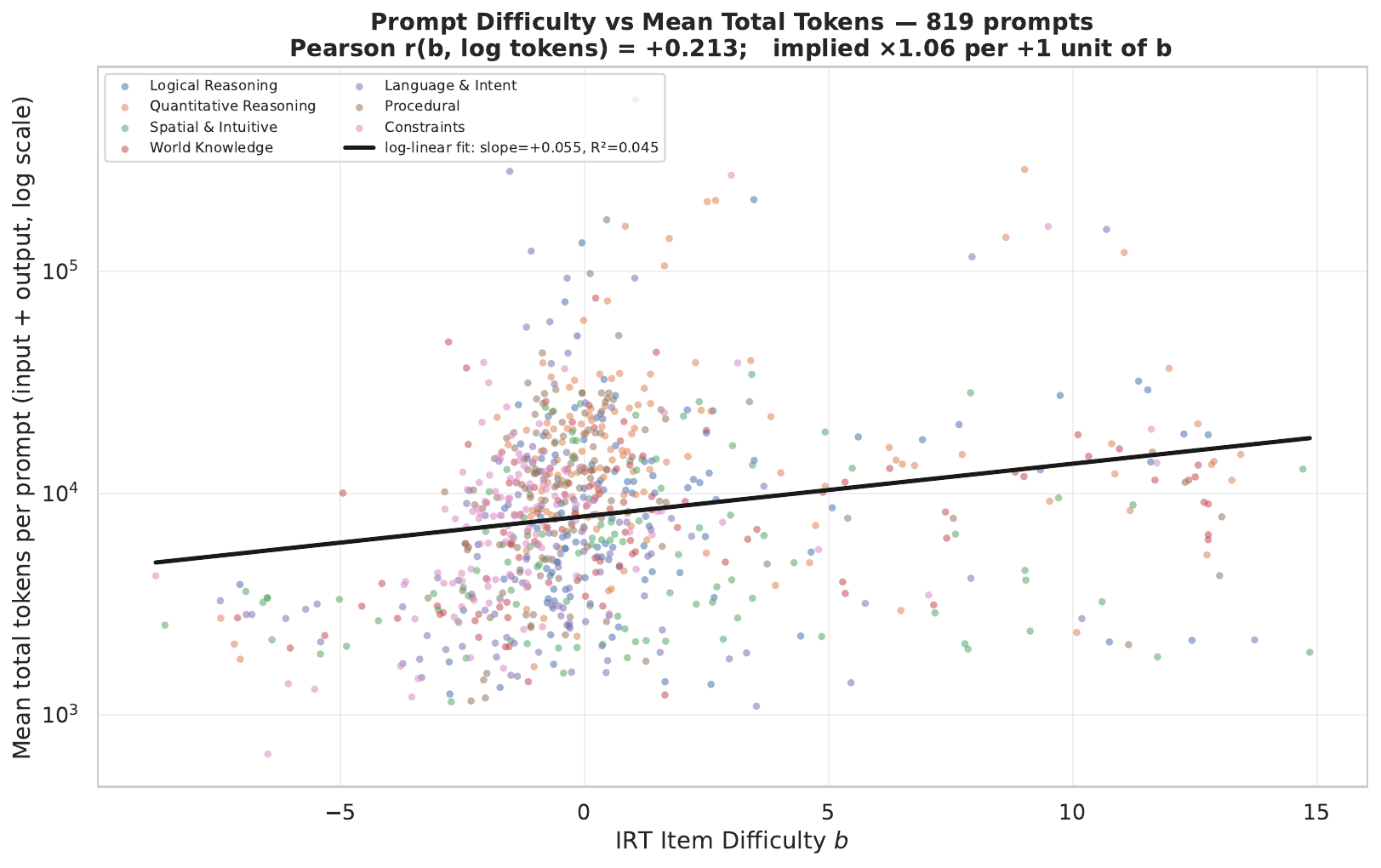}
\caption{Per-prompt mean total tokens (input + output, averaged across all retained model$\,\times\,$thinking-level attempts) vs.\ IRT item difficulty $b_j$, on a log y-axis. Each point is one prompt, colored by primary cognitive category; the black line is a log-linear fit. Harder prompts elicit exponentially more tokens on average across models.}
\label{fig:difficulty-vs-tokens}
\end{figure}

Figure~\ref{fig:difficulty-vs-tokens} plots, for each prompt, the mean total tokens (input + output, averaged across all (model, thinking-level) configurations that attempted it) against the calibrated $b_j$, with each point colored by primary cognitive category and a log-linear fit overlaid. Token spend grows monotonically and approximately exponentially in $b_j$: harder prompts elicit longer reasoning traces from essentially every model, regardless of family or thinking tier. Two implications follow. First, the bank's difficulty axis is not just a statistical fit---it tracks a directly observable behavioral cost, in tokens, that all models pay on the harder items. Second, this bank-level cost-vs-difficulty curve is the item-side complement to the model-side cost-of-thinking analysis in Section~\ref{subsection:thinking}: GIM rewards models that allocate more inference compute to the prompts that need it, and the bank's hardest items are precisely the ones that absorb the most thinking on average.

\subsection{Calibration Trust: Robustness, LOMO, and Per-Epoch Variance}
\label{appendix:calibration-trust-detail}
\label{subsection:irt-robustness}

This subsection collects the supporting evidence for the calibration trust summary in Section~\ref{subsection:calibration-trust}. All checks operate on the same observed response tensor $\{p_{ijk}\}$ used for the judge-aware headline calibration.

\subsubsection{Logit squeeze transform}
\label{appendix:logit-squeeze}
The continuous-response logit transform we apply to the rubric-weighted score $p_{ijk}$, referenced from Section~\ref{subsection:irt-model}, is
\begin{equation}
p'_{ijk} = 0.998\,p_{ijk} + 0.001, \qquad y_{ijk} \;=\; \ln\!\left(\frac{p'_{ijk}}{1-p'_{ijk}}\right),
\end{equation}
following \citet{smithson2006better}. The squeeze maps the closed interval $[0,1]$ into $[0.001, 0.999]$ so that perfect and zero scores receive finite logits; the constants are chosen to bound the maximum logit at $\pm \ln(999) \approx \pm 6.9$, comfortably outside the empirical range of $|y_{ijk}|$ but small enough that interior values are essentially unaffected. Refitting the bank with the squeeze magnitude varied over $[10^{-4}, 10^{-2}]$ leaves abilities, item parameters, judge effects, and the leaderboard ordering essentially unchanged.

\subsubsection{Optimization and penalty terms}
\label{appendix:irt-optimization}
The penalty terms in the calibration loss (Section~\ref{subsection:irt-model}) resolve the well-known scale and location indeterminacies of the 2PL: the likelihood is invariant to the joint shift $\theta_i \to \theta_i + c$, $b_j \to b_j + c$ and to the joint rescaling $\theta_i, b_j \to k\theta_i, k b_j$, $a_j \to a_j / k$. We use $\lambda = 0.5$ on the log-discrimination ridge and $\mu = 0.01$ on the centering term; the judge effects are constrained to sum to zero so their offsets are identifiable separately from the global ability/difficulty location. Abilities, item difficulties, judge effects, and the resulting leaderboard are insensitive to $\lambda \in [0.1, 2]$ and $\mu \in [10^{-3}, 10^{-1}]$. We optimize with Adam (lr $=10^{-2}$) until convergence (here, 9{,}207 steps); the final fit achieves $R^2 = 0.642$ and residual standard deviation $\hat\sigma = 2.24$ in logit space.

\subsubsection{Leave-one-model-out (LOMO) calibration check}
\label{appendix:lomo}
\begin{figure}[htbp]
\centering
\includegraphics[width=\textwidth,keepaspectratio]{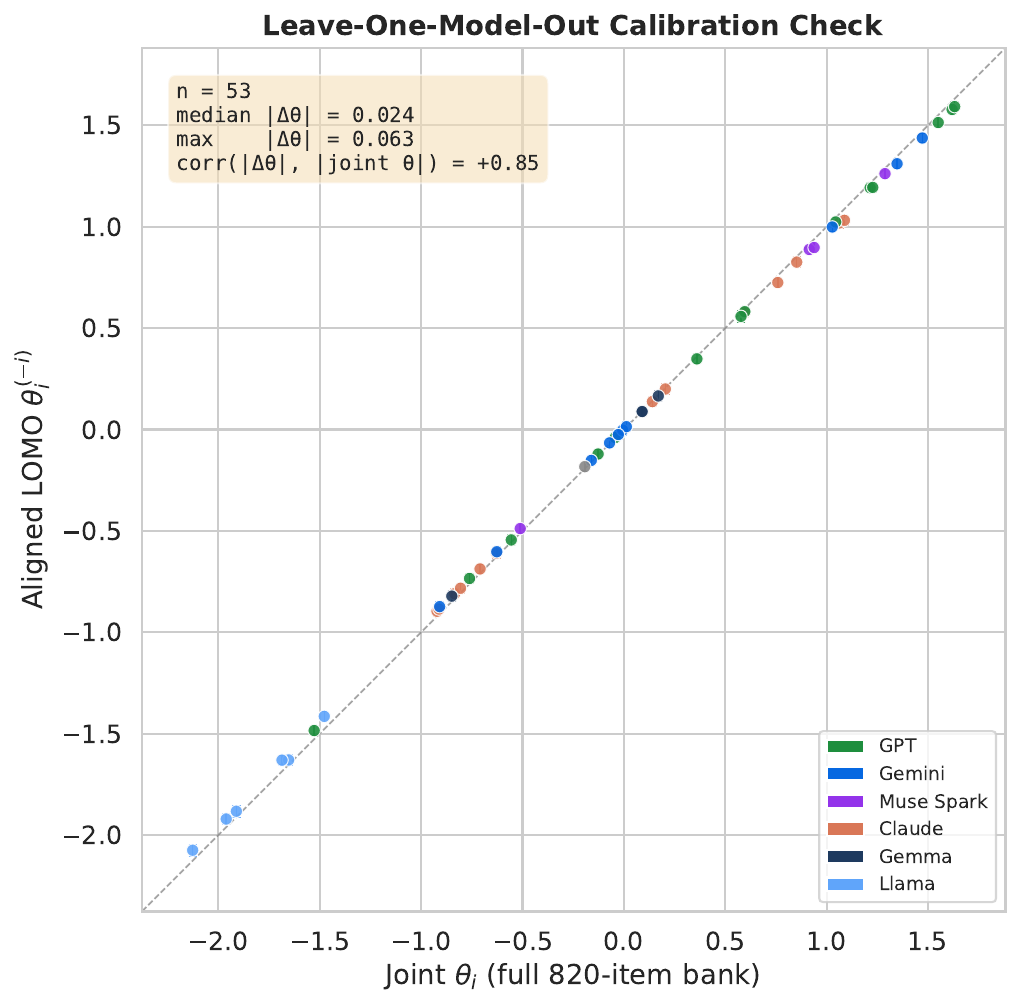}
\caption{Leave-one-model-out (LOMO) calibration check. For each of the 47 reporting configurations, we refit the bank with that configuration removed and re-score it against the held-out item parameters $\{a_j^{(-i)}, b_j^{(-i)}, \gamma_k^{(-i)}\}$. The held-out $\theta_i^{(-i)}$ recovers the joint $\theta_i$ to within $0.063$ logits across all configurations (median $0.024$, mean $0.024$), close to the typical 95\% CI half-width of $\approx 0.05$ logits.}
\label{fig:lomo-calibration}
\end{figure}

The LOMO ability $\theta_i^{(-i)}$ recovers the joint estimate $\theta_i$ to within $0.063$ logits across all 47 reporting configurations; the median absolute deviation is $0.024$ logits, the mean is $0.024$, and the 95th percentile is $0.053$---close to the typical 95\% CI half-width of $\approx 0.05$ logits. Figure~\ref{fig:lomo-calibration} plots the result. The largest residuals occur at the high- and low-ability extremes, where the bank has fewer ability-matched neighbours to anchor the held-out estimate, exactly as a finite-bank calibration should behave. No refit produced a deviation in the ``uncomfortably model-dependent'' range we would consider a calibration failure ($\geq 0.3$ logits).

\subsubsection{Per-epoch sampling reliability}
\label{subsection:variance}
\begin{figure}[htbp]
\centering
\includegraphics[width=\textwidth,keepaspectratio]{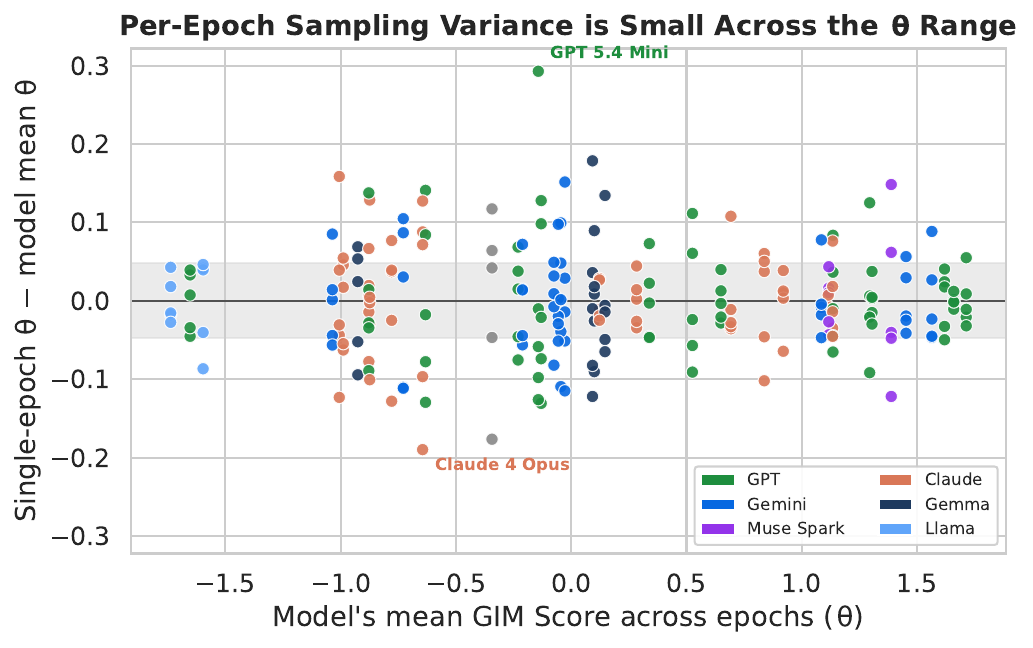}
\caption{Per-epoch sampling variance of $\theta$. Each point is one (model, epoch) re-scoring; the y-axis is that epoch's $\theta$ minus the model's across-epoch mean. The gray band is $\pm 1.96 \cdot \widetilde{\mathrm{SE}}$ from the joint scoring. Single-epoch noise is small and roughly constant across the ability range.}
\label{fig:epoch-variance}
\end{figure}

Each (model, thinking-level) configuration is evaluated for 5 independent epochs on the 820-item bank. Beyond robustness checks on the calibration itself, this lets us separate two things that are typically conflated when reporting a single benchmark number: how strong a model is on average ($\theta$ from the joint scoring) and how reliable a single benchmark run is. To probe the latter we re-score each model independently on each epoch's responses, against the same frozen item bank, and compare the per-epoch $\theta_{\text{epoch}}$ to the model's across-epoch mean. Figure~\ref{fig:epoch-variance} summarizes the result. Single-epoch sampling noise is small in absolute terms---the bulk of the cloud sits inside the gray band of $\pm 1.96 \cdot \widetilde{\mathrm{SE}}$, where $\widetilde{\mathrm{SE}}$ is the median analytical SE from the joint scoring---and the noise is roughly constant across the ability range. The leaderboard ordering would therefore not meaningfully change if we had run a single epoch instead of five, although confidence intervals on individual $\theta$ values would be roughly $\sqrt{5}\times$ wider.

\subsubsection{Inference-failure rate across configurations}
\label{subsection:inference-failures}
Defining a failure as any sample with \texttt{sample\_length}~$=0$ (the model returned nothing usable) and computing the per-(model, thinking-level) failure rate over all (prompt~$\times$~epoch) attempts, the 53 test-configurations span $0.00\%$ to $\sim\!13\%$ with a median of $1.1\%$. Failures concentrate in two structurally distinct regimes that have little to do with model quality per se. At the top end of the reasoning-budget axis, a handful of high-thinking configurations sit in the $8$--$13\%$ band: longer chains of thought compound exposure to API timeouts, server-side errors, and context-length truncation, so the most aggressive thinking budgets pay the highest infrastructural tax. At the small or aggressively quantized end of the model axis, the lowest-capacity open-weight configurations (e.g., a 1B-parameter model with thinking off, fp4 quantizations at Low thinking) cluster in a similar band, where output collapse and decoding instability dominate. The intermediate configurations cluster well below $5\%$, with several registering zero failures across all attempts, including the centaur cohort. The IRT scorer absorbs these failures by treating them as missing rather than zero (Appendix~\ref{appendix:irt-affordances}, bullets 2 and 5).

\subsection{Public vs.\ Private Contamination Diagnostic (Detailed)}
\label{appendix:contamination-detail}

\begin{figure}[htbp]
\centering
\includegraphics[width=0.85\textwidth,keepaspectratio]{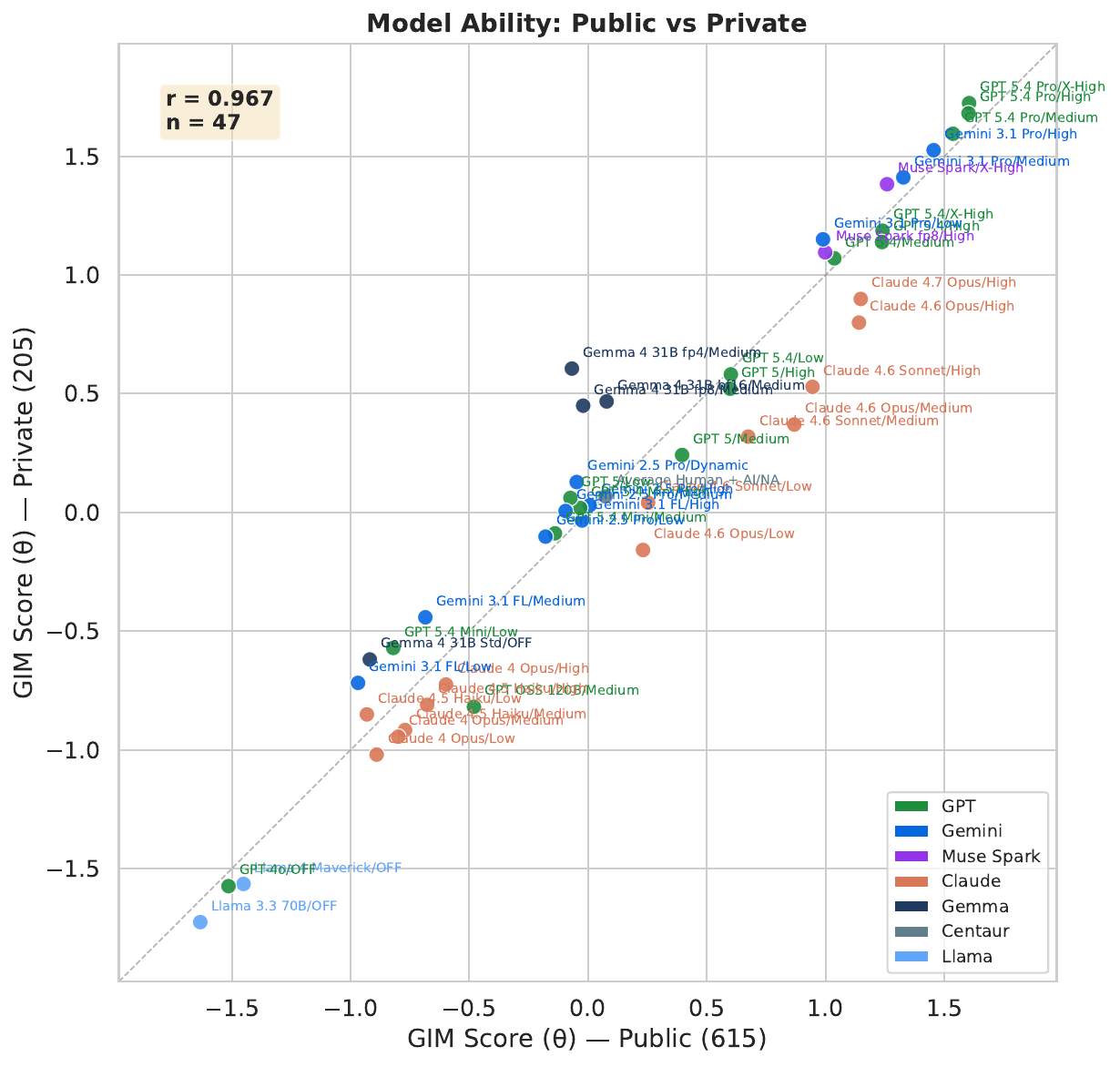}
\caption{Per-model GIM score $\theta$ on the 615-prompt public split (x-axis) vs.\ the 205-prompt private split (y-axis). Error bars are 95\% CIs from each subset's closed-form scoring; the dashed gray line is $y = x$. All 47 reporting configurations sit near the identity line ($r \approx 0.97$).}
\label{fig:public-vs-private}
\end{figure}

Figure~\ref{fig:public-vs-private} shows the (public $\theta$, private $\theta$) scatter for all 47 reporting configurations. The points hug the $y = x$ identity line tightly across the full $\sim 3$-logit range; the Pearson correlation between $\theta_{\text{public}}$ and $\theta_{\text{private}}$ is $r \approx 0.97$. No model is a visible outlier above the diagonal in a direction that would indicate public-only inflation.

\begin{figure}[htbp]
\centering
\includegraphics[width=0.85\textwidth,keepaspectratio]{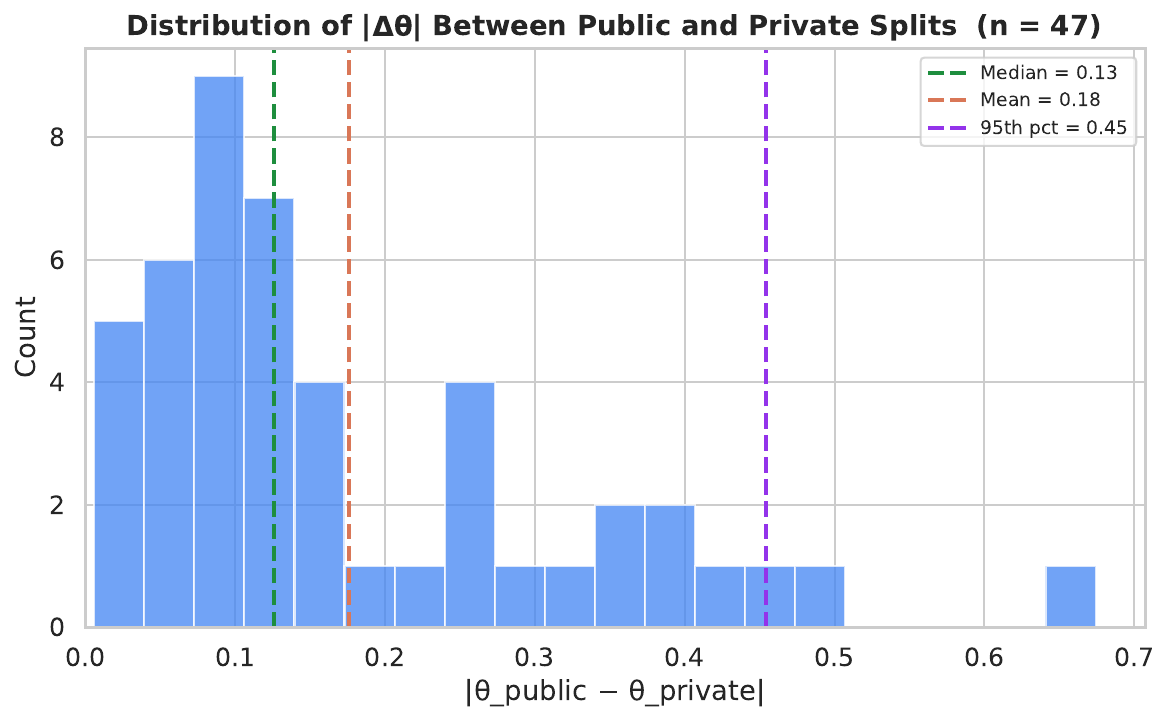}
\caption{Distribution of $|\theta_{\text{public}} - \theta_{\text{private}}|$ across all 47 reporting configurations. Median, mean, and $95^\text{th}$ percentile are overlaid as dashed lines. The bulk of the mass sits well below the typical per-model analytical SE.}
\label{fig:public-vs-private-absdiff}
\end{figure}

To make the size of any public--private gap concrete, Figure~\ref{fig:public-vs-private-absdiff} plots the histogram of $|\theta_{\text{public}} - \theta_{\text{private}}|$ across the same configurations. The distribution is concentrated at small values: the median absolute difference is well below the typical per-model SE from Section~\ref{subsection:leaderboard}, the mean is similarly small, and even the $95^\text{th}$ percentile is on the order of one analytical SE. In other words, the private split, which no model could plausibly have trained on, places models in essentially the same ability ranking as the public split.

\section{AI Usage Disclosure}
\label{appendix:ai-usage}

AI generation of questions or reference answers was explicitly forbidden: every prompt, reference answer, rubric criterion, and free-text descriptive label in GIM was written by a human author. Humans also designed and approved the seven-category taxonomy, selected the public/private split, decided which analyses and charts to produce, and verified every numerical result and every line of wording in this paper. AI was used in supporting roles that humans directed and reviewed---probing candidate prompts against state-of-the-art models as a difficulty filter, brainstorming candidate category structures, assigning primary and secondary categories under sample human review, running model inference and first-pass LLM-as-judge scoring (with humans sampling and reviewing outputs from every run), and writing the codebase, analysis pipelines, and chart-generating code under explicit human specification. The human authors reviewed and approved every AI-assisted output before it was incorporated, and take full responsibility for the content. Table~\ref{table:ai-usage} provides a stage-by-stage breakdown of human and AI contributions during GIM's construction, evaluation, and write-up.

\begin{table}[!htbp]
\centering
\caption{Human vs.\ AI contribution at each stage of GIM's construction, evaluation, and write-up. ``---'' indicates no involvement at that stage. Every AI-assisted step was reviewed and approved by the human authors before being incorporated.}
\label{table:ai-usage}
\small
\begin{tabular}{l p{0.36\textwidth} p{0.36\textwidth}}
\toprule
\textbf{Stage} & \textbf{Human role} & \textbf{AI assistance} \\
\midrule
Question authoring          & Wrote every prompt; AI generation explicitly forbidden.                          & --- \\
Reference answers / rubrics & Authored and verified all reference answers and rubric criteria.                 & --- \\
Difficulty calibration      & Reviewed flagged prompts; revised or discarded those judged too easy.            & Probed candidate prompts against SOTA models to surface those solved easily. \\
Taxonomy design             & Reviewed and approved the seven primary categories and 18 sub-categories.        & Assisted in brainstorming, refining, and discussing candidate category structures. \\
Category assignment         & Sample-reviewed AI-assigned primary/secondary categories for quality.            & Assigned primary and secondary categories to each prompt. \\
Free-form labels            & Manually authored every descriptive label (e.g., ``math'', ``reasoning'', ``verifiable''). & --- \\
Public/private split        & Selected and held out the 205 private problems.                                  & --- \\
Model inference             & Reviewed a sample of model outputs from every run.                               & Generated model responses to every prompt under each (model, thinking-level) configuration. \\
Judge scoring               & Reviewed a sample of judge outputs from every run; validated judge reliability.  & Default scoring of model responses by the LLM-as-judge framework. \\
Codebase                    & Specified what needed to be built (pipelines, IRT calibration, judge harness); reviewed all code before it was merged and run. & Wrote the implementation under human direction. \\
Analysis \& charts          & Decided which analyses to run and which charts to produce; reviewed all outputs.    & Wrote the analysis and plotting code that generated the artifacts. \\
Numerical results           & Verified all numbers reported in the paper.                                      & Produced via the (AI-written, human-reviewed) analysis code. \\
Manuscript drafting         & Authored the final scientific claims and wording; signed off on every section.   & Suggested phrasing, critiqued structure, and surfaced gaps during drafting. \\
\bottomrule
\end{tabular}
\end{table}

\end{document}